\documentclass{article}

    \usepackage[preprint]{score_workshop}

\usepackage[utf8]{inputenc} 
\usepackage[T1]{fontenc}    
\usepackage{hyperref}       
\usepackage{url}            
\usepackage{booktabs}       
\usepackage{amsfonts}       
\usepackage{nicefrac}       
\usepackage{microtype}      
\usepackage{xcolor}         

\usepackage{lmodern}
\usepackage[T1]{fontenc}
\usepackage{relsize}

\usepackage{amsfonts}
\usepackage[utf8]{inputenc}
\usepackage[english]{babel}
\usepackage{amsthm}
\usepackage{algorithm}
\usepackage{algorithmic}
\usepackage{bbold}
\usepackage{dsfont}

\usepackage[nolabel, final]{showlabels}
\usepackage{graphicx}
\usepackage{subfigure}
\usepackage{centernot}
\usepackage{mathtools}
\usepackage{bbm}

\usepackage{wrapfig}
\usepackage{enumerate}

\usepackage{cleveref}
\usepackage{todonotes}

\newtheorem{Theorem}{Theorem}

\newtheorem{Definition}{Definition}
\newtheorem{Proposition}{Proposition}

\newcommand{\R}{\mathbb{R}} 
\renewcommand{\H}{\mathcal{H}} 
\newcommand{\G}{\mathcal{G}} 
\newcommand{\A}{\mathcal{T}} 
\renewcommand{\P}{\mathbb{P}} 
\newcommand{\E}{\mathbb{E}} 
\newcommand{\N}{\mathbb{N}} 
\newcommand{\T}{\mathcal{T}}
\newcommand{\Trans}{\hspace{-0.25ex}\top\hspace{-0.25ex}}

\newcommand{\q}{\widehat{q}}

 	\definecolor{amber}{rgb}{1.0, 0.75, 0.0}

\title{On {RKHS} 
Choices for Assessing Graph Generators via Kernel Stein Statistics}

\author{%
   Moritz Weckbecker 
   \\
   Department of Statistics \\
   University of Oxford \\
  \texttt{moritz.weckbecker@mansfield.ox.ac.uk} \\
   \And
      Wenkai Xu \\
   Department of Statistics \\
   University of Oxford \\
   \texttt{wenkai.xu@stats.ox.ac.uk} \\
   \And
      Gesine Reinert \\
   Department of Statistics \\
   University of Oxford \\
   and The Alan Turing Institute\\
   \texttt{reinert@stats.ox.ac.uk} \\
}

\begin{document}

\maketitle

\begin{abstract}
{{Score-based} kernelised Stein discrepancy (KSD) tests 
have emerged as a powerful tool for the goodness of fit tests{, especially} in high dimensions{; however, the test performance {may}}
depend on the choice of kernels in an underlying reproducing kernel Hilbert space (RKHS).} 
{Here we assess the effect of RKHS choice 
{for} KSD tests {of}} 
random networks {models},
developed for 
exponential random graph models (ERGMs) {in} \cite{xu2021stein} and {for}  synthetic graph generators {in}  \cite{xu2022agrasst}. 
We investigate 
the power {performance} and 
the {computational} runtime 
of the test in different scenarios, in{cluding} both dense and sparse graph regimes. 
Experimental results on kernel performance for model assessment tasks are shown and discussed on 
synthetic 
and real-world network applications.
\end{abstract}

\section{Introduction}

Recent advances in high-dimensional goodness of fit tests ha{ve} been achieved by score-based kernelised Stein discrepancy (KSD) tests, starting with \cite{chwialkowski2016kernel} and \cite{liu2016kernelized}.  {A} {KSD} 
test {relies on an underlying reproducing kernel Hilbert space (RKHS) and hence its}
{performance}
{may} depend on the choice of 
RKHS  for the 
set of test functions. Typically, 
the choice of {RKHS} 
is restricted by {having} 
to {lie} in the {\it Stein class} of the target distribution {(see  \Cref{sec:kss} for details)}. A notable exception is the case that the target distribution is that of a finite random network, as then any finite function is a member of the Stein class. Thus, this situation is well suited to assessing the choice of RKHS.

In this paper we 
assess the choice of RKHS for {the two available}   graph-based kernelised Stein goodness of fit tests, {which take a single network as input}, namely gKSS from \cite{xu2021stein} and AgraSSt from \cite{xu2022agrasst}. The 
RKHS kernels which we explore 
are the constant kernel, the vertex-edge histogram kernel  \citep{kriege2016valid}, the shortest path kernel \citep{borgwardt2005shortest}, random walk kernels \citep{gartner2003graph, sugiyama2015halting}, the Weisfeiler-Lehman kernel  \citep{shervashidze2011weisfeiler}, the graphlet kernel \citep{ahmed2017graphlet} and the connected graphlet kernel  \citep{shervashidze2009efficient}.
%
{As t}he {influence of the} 
RKHS {choice}
{on the power} 
of the goodness of fit test {may depend on the problem setting, {here we investigate a collection of test problems, including  edge-two star (E2S) models,  geometric random graph (GRG) models, Barabasi-Albert (BA) models, and  the black-box random graph generator CELL  \citep{rendsburg2020netgan}}.} 
{T}he kernel choice may also have a significant effect {on} the runtime {which is hence included in the analysis}. 


The paper is structured as follows. {In}  \Cref{sec:kss}, we 
{introduce} the basic form of kernel Stein statistics and its corresponding goodness-of-fit testing procedure.
Then we discuss 
kernel choices 
in \Cref{sec:kernel choice}.  
In \Cref{sec:exp}, we present the experimental results on 
E2S models 
relating to the experiments in \cite{xu2021stein}, the  
GRG models,
{as well as {in} 
CELL, trained
on real-world networks, } followed by concluding discussions.
Additional 
background, {results on the BA models and a GRG on a unit square,}  {{as well as a} computational efficient  algorithm for geometric random walk kernels}
are deferred to the appendix. 
Code for the experiment is 
available at
\url{https://github.com/MoritzWeckbecker/dissertation-kss}. 

\section{Background: kernel Stein statistic for
random graph models}\label{sec:kss}
{{Let}  $\mathcal{G}_{n}^{lab}$ {denote} the set of}
vertex-labeled {simple} graphs 
on {$n$ vertices}.
{For a 
probability distribution $q$, an operator  $\T_q$  on $\mathcal{G}_{n}^{lab}$ }
{satisfying} 
the {\it Stein identity}  $\E_q[\T_q f(x)]=0$ { for all test functions $f:\mathcal{G}_{n}^{lab} \to \R$ in a {\it Stein class}  $\mathcal F$ is called a {\it Stein operator}}. {With 
$q(x^{(s,1)}|{x_{-s}})$ denoting the  conditional probability that vertex pair $s$ has an edge, given the network except the edge indicator of $s$ {(and}
similarly  $q(x^{(s,0)}|{x_{-s}})$ 
the conditional probability that 
vertex pair 
$s$ does not have an edge{)}
so that
$q(x^{(s,\cdot)}|{x_{-s}})$ is a discrete score function,}  the Stein operator {in \cite{xu2021stein}} is 
$\T_q f 
= \frac{1}{N}\sum_{s\in [N]} \T^{(s)}_q f$ where $${\A_q^{(s)} f(x)} = q(x^{(s,1)}|{x_{-s}} )f (x^{(s,1)}) + q(x^{(s,0)}|{x_{-s}} )  f (x^{(s,0)})  - f(x)  
. \footnote{$N = 
n(n-1)/2$ is the total number edges; $[N]:=\{1, \ldots, N\}$ denotes the 
set of vertex pairs; $x^{(s,1)}$ has the $s$-entry replaced of $x$ by $1$; 
$x_{-s}$ is the network $x$ with edge index $s$ removed.}$$
{If a distribution $p$ is close to $q$ then {one would expect that}  $\E_p[\T_q f(x)]\approx 0$; hence $\sup_{f \in {\mathcal F}}  | \E_p[\T_q f(x)]| $ can be used to assess the distributional distance between $q$ and $p$.  
Choosing as ${\mathcal F}$ the unit ball of 
{a} 
RKHS $\H$ allows to compute the supremum exactly.}   
The {\it  kernel Stein statistic}  (KSS) {based on a single network sample $x$} is 
defined as  
\begin{equation}\label{eq:kss}
    \operatorname{KSS}(q;x) 
    = \sup_{\|f\|_{\H}\leq 1} \left|\E_{q(\cdot|x)}[f(X_{t_1}) | X_0 = x] - f(x) \right|  
    =\sup_{\|f\|_{\H}\leq 1} \Big|\frac{1}{N}\sum_{s\in [N]}\left[\A^{(s)}_q f(x)\right] \Big|.
\end{equation}
{For computational efficiency, instead of averaging over all $N$ possible edges, a vertex pair re-sampling version is also provided.} 
Let {the} re-sample size be $B$, {then} draw $\{s_1,\dots,s_B\}$ uniformly from $[N]$. Denote $k$ the kernel associate with $\H$. {Then we estimate} 
$$\widehat{\operatorname{KSS}}^2(q;x) =  \sup_{\|f\|_{\H}\leq 1} \Big|\frac{1}{B}\sum_{b\in [B]}\left[\A^{(s_b)}_q f(x)\right] \Big|^2 =  \frac{1}{B^2} \sum_{b,b'} \underset{{h(s_b,s_{b'})}}{\underbrace{\left\langle  \A^{(s_b)}_q k(x,\cdot), \A^{(s_{b'})}_q k(x,\cdot)  \right\rangle}},$$
 where $h$ is referred to as the Stein kernel\footnote{The Stein kernel {$h{(s,s')}$}  has to be clearly distinguished from the RKHS kernel $k{(x,x')}$.}.
When $q$ {is the distribution of an 
exponential random graph model,
} 
{KSS coincides with gKSS from}
\cite{xu2021stein}. 
When $q$ does not {have an} 
explicit 
form, e.g. for {samples} {from}
$q$ generated by a 
black-box random graph generator,
\cite{xu2022agrasst} approximate the conditional distribution
using samples {generated from $q$ to estimate the conditional score function.

\cite{xu2022agrasst} {also suggest} 
conditioning on a user-defined graph summary statistic} 
$t(x)$ {and replacing}  $q(x^{(s,1)}|x_{-s})$ 
by  $\widehat q(x^{(s,1)}|t(x_{-s}))$ in Eq.\eqref{eq:kss} {; {this} is termed}  
the 
{\it approximate graph Stein Statistic }(AgraSSt) 
in \cite{xu2022agrasst}.  I{n \cite{xu2021stein} and \cite{xu2022agrasst}   theoretical guarantees are also given}. 

{For testing the goodness of fit of the model $q$ {based on} a single network observation $x$, 
{gKSS and AgraSSt use 
the Monte Carlo procedure}, } 
simulating $l$ {independent} networks  $z_1,\dots,z_l \sim q$  and 
compar{ing} the {observed} $\widehat{\operatorname{KSS}}^2(q;x)$ with 
the {set of} $\widehat{\operatorname{KSS}}^2(q;z_i), {i=1 \in [l]}$. 
{W}e reject the null model {if $\widehat{\operatorname{KSS}}(q;x)$ is large.}
{Details are given} in \Cref{alg:kernel_stein_monte_carlo} in \Cref{app:background}.

\section{Graph kernel choices and their effects on KSS}\label{sec:kernel choice}

\paragraph{Graph kernels considered}
While details of {the}  Gaussian vertex-edge histogram (GVEH) kernel, 
the shortest path (SP)  kernel,
the  { $K$-step random walk (KRW) and geometric random walk (GRW)}  kernels,
and the Weisfeiler-Lehman (WL) kernels 
can be found in {A}ppendix B of \citep{xu2021stein}, we recollect them in \Cref{app:graph_kernel}. We additionally consider
graphlet {counts} 
{(}sub-graphs with a small number of vertices{)}, as the
features to 
compare graph structures. 

 \begin{wrapfigure}{r}{80mm}
    \centering
    \includegraphics[width=0.48\textwidth]{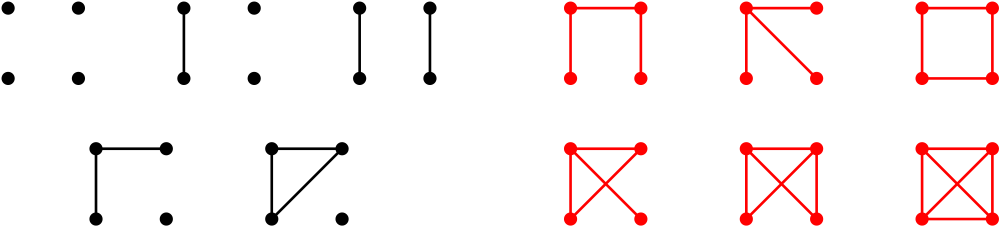}
    \caption{\footnotesize  Graphlets illustration for $l=4$: $N_4 = 11$; the connected graphlets are marked as red.}
    \label{fig:graphlet}
    \vspace{-0.25cm}
\end{wrapfigure}
The idea of {a} graphlet kernel \citep{ahmed2017graphlet} is to {count the  occurrences of} 
all simple undirected graphs of size $l$, up to permutation of the vertices, $
\{\mathfrak g_1,\dots, \mathfrak g_{N_l}\}$ where $N_l$ denotes the number of distinct $l$-graphlets. An example 
{for}  $l=4$ is shown in \Cref{fig:graphlet}.
The graphlet feature $\phi_{glet}(x) \in \N^{N_{l}}$
{has as} $i$-th
entry 
the number of occurrences of $\mathfrak g_i$ in $x$.
{This} feature naturally induces the graphlet kernel $k_{GLET}(x,x') = \langle \phi_{glet}(x), \phi_{glet}(x')\rangle.$
When only the connected graphlets (e.g. \Cref{fig:graphlet} in red) are considered to construct the feature $\phi_{conglet}(x)$, the connected graphlet kernel  {is given by} $k_{CONGLET}(x,x') = \langle \phi_{conglet}(x), \phi_{conglet}(x')\rangle.$

\begin{wraptable}{r}{65mm}
\footnotesize
\vspace{-.4cm}
    \centering
    \begin{tabular}{l|l}
        \toprule
 \textbf{Graph kernels}   & \textbf{Parameters} \\
    \midrule
GVEH & bandwidth $\sigma>0$
\\
KRW 
& maximal walk length $K$ 
\\
GRW & discount weight $\lambda$
\\
WL & level parameter $h$
\\
GLET & size of graphlets $l$ 
\\
        \bottomrule
    \end{tabular}
    \vspace{-0.25cm}
    \caption{\footnotesize Graph kernels and the parameters.}
    \vspace{-0.15cm}
    \label{tab:kernel_param}
\end{wraptable}

For our experiment, we use the implementation provided by
the R package \texttt{graphkernels} \citep{sugiyama2018graphkernels}. In addition, we devise the ``constant'' kernel $k_{Const}(x, x')\equiv 1$
as a benchmark in our experiment. 
{Like} 
{the} SP kernel, the constant kernel has no parameter.

We list the parameters for {the} kernels
in \Cref{tab:kernel_param}.


\section{Experimental results}\label{sec:exp}
{In our  experiments 
the {observed network is assumed to be generated by model $M0$, and the goodness of fit is tested for  model  $M1$; the null
} hypothesis $H_0: M0 = M1$ is rejected in favour of $H_1: M0 \neq M1$ at the $5\%$-level
using a
gKSS or an
AgraSSt 
test  
with $n_{M1}$ graphs simulated under $M1$. In the synthetic examples we  repeat this procedure on  $n_{M0}$ graphs from the null model to obtain the rejection rate. Unless otherwise stated, $n_{M0} = n_{M1} = 500.$ More details can be found in \Cref{app:exp}.}

\begin{figure*}[htp!]
    \centering
    \vspace{-0.18cm}

{\includegraphics[width=0.33\textwidth]{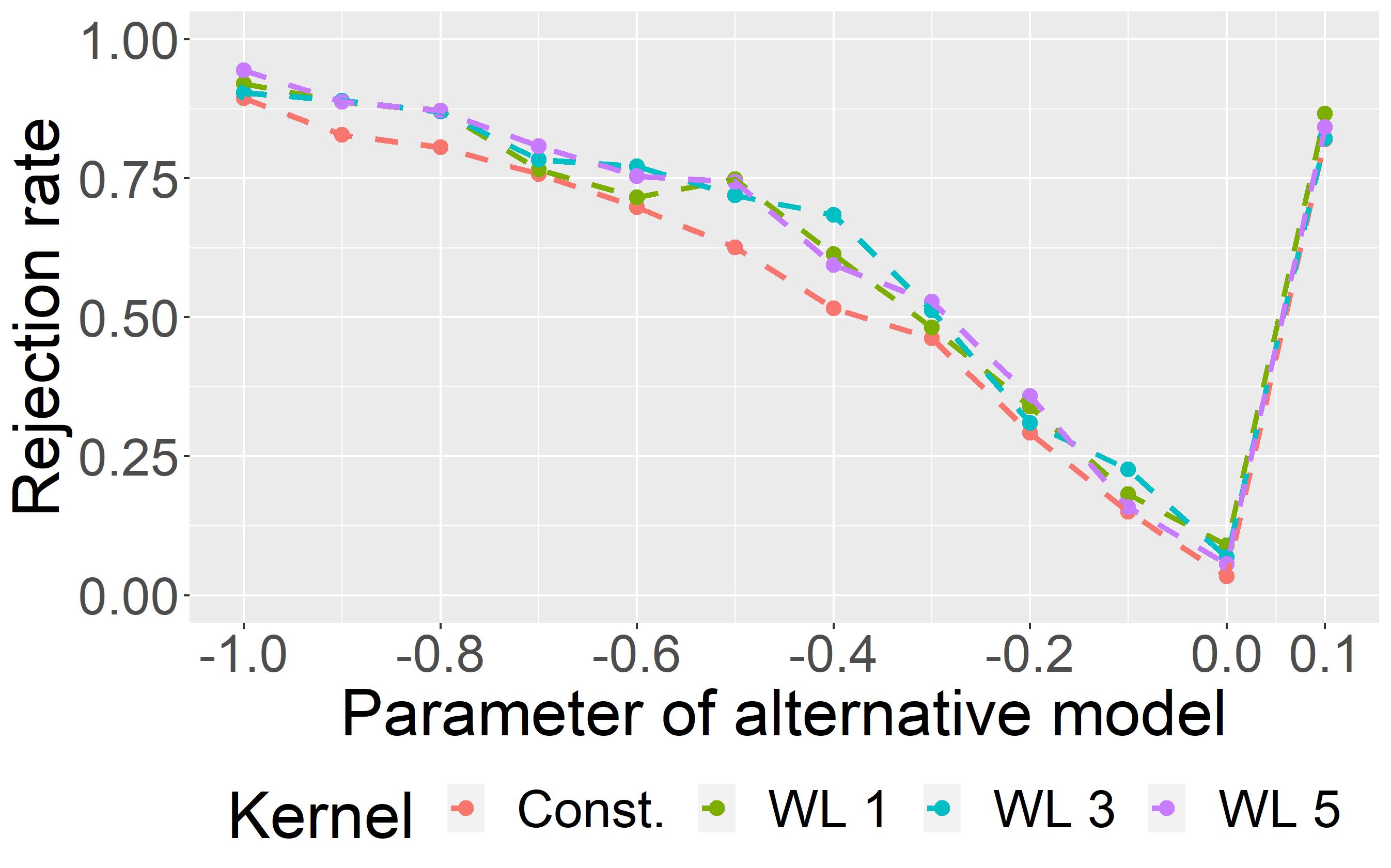}}\includegraphics[width=0.33\textwidth]{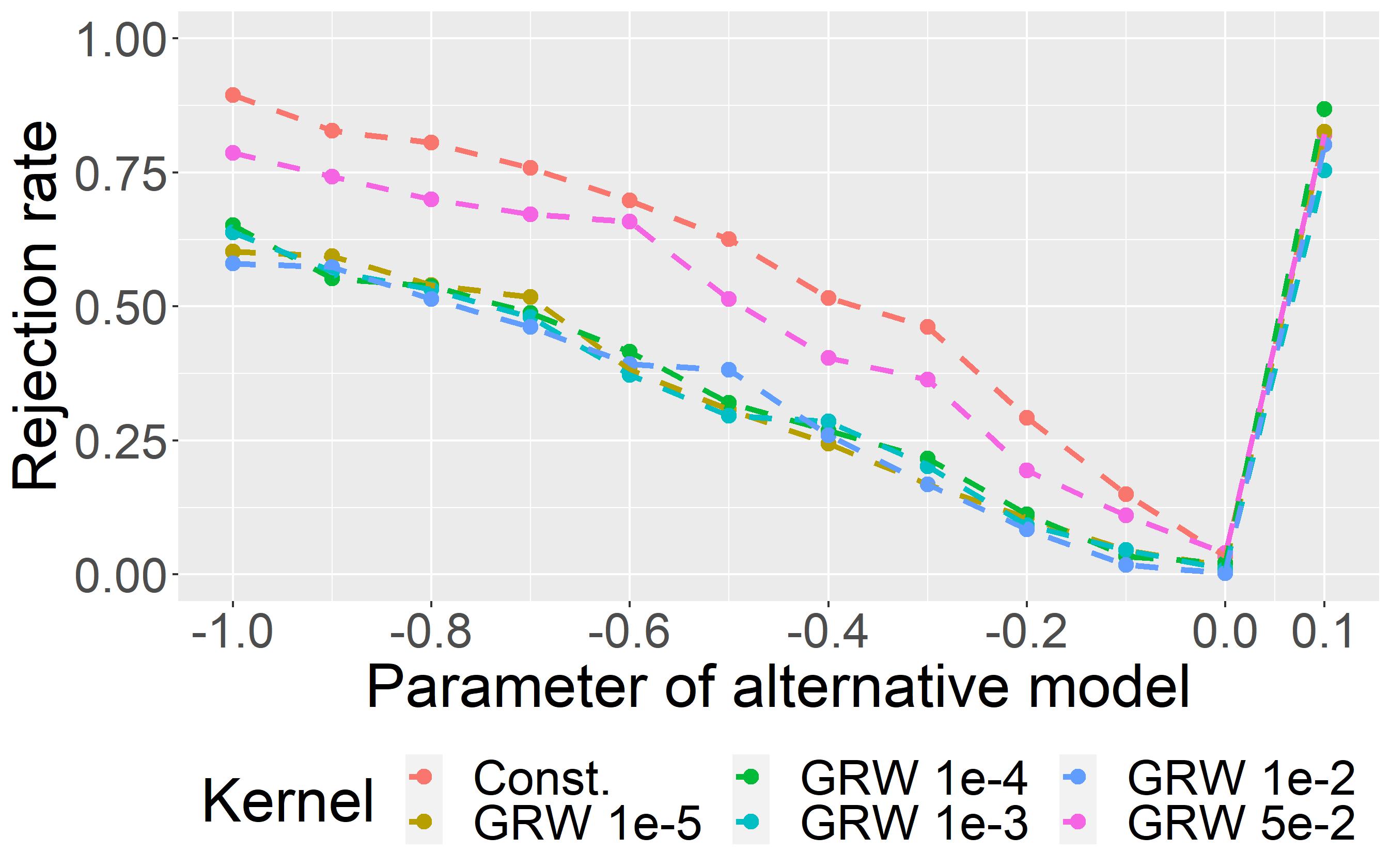}
    {\includegraphics[width=0.33\textwidth]{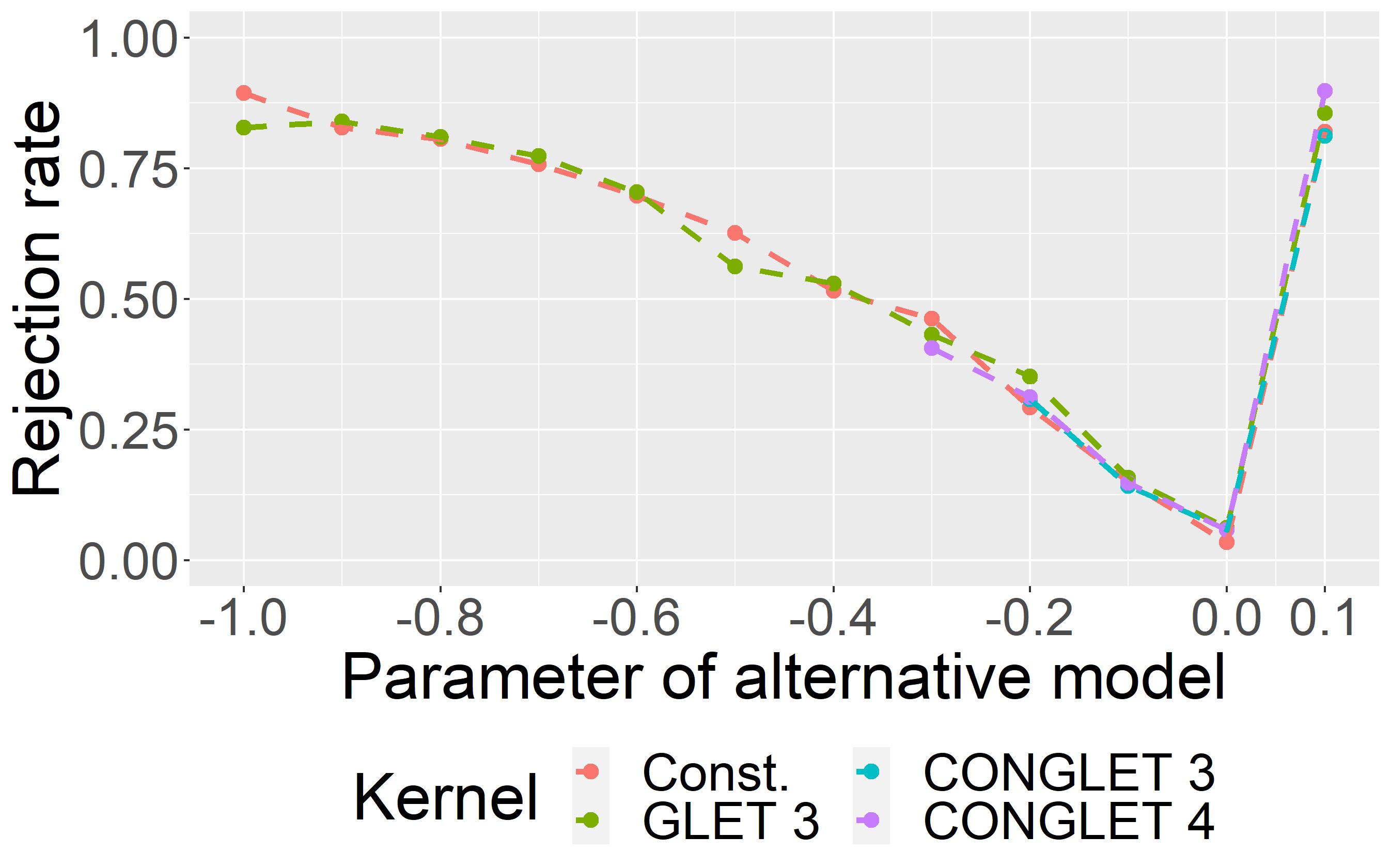}}
    \vspace{-0.5cm}
    \caption{\footnotesize gKSS on the E2S {model}  with $\beta_2$ perturbed; $n=20$ and $B=200$.
    }
    \vspace{-0.25cm}
    \label{fig:e2s}
\end{figure*}

\paragraph{An ERGM example} The Edge-2Star (E2S) model {on $\mathcal{G}_{n}^{lab}$ is an ERGM with} 
density $q(x) \propto \exp(\beta_1 E(x) + \beta_{s} S_2(x))$ where $E(x)$ denotes the edge count and $S_2$ denotes the number of two-stars. {In our experiments,  following the setting of \cite{xu2021stein}  t}he null model is $\beta = (-2,0)$ and the alternative is set by perturbing $\beta_2$.  
{As this is an ERGM, we}
apply {an} gKSS test  \citep{xu2021stein};   \Cref{fig:e2s}  show{s} rejection rates for WL, GRW and graphlet kernels with different parameters.
%
From the result, 
all kernels had well-controlled type 1  error, {and even} 
using 
a constant kernel already had 
good test power. While {a} WL kernel generally outperformed the constant kernel, GRW kernels of different parameters {were outperformed by} 
the constant kernel. {This finding could perhaps be explained by the change in density between null and alternative; {this change}  may already suffice for separating the two, and the constant kernel picks this up. In contrast, GRW kernels focus more on local structure.} 
The graphlet kernels generally have similar test power compared to {the} constant kernel; {they would pick up a  change in density through a change in graphlet counts}.

\begin{figure*}[tp!]
    \centering
        {\includegraphics[width=0.32\textwidth]{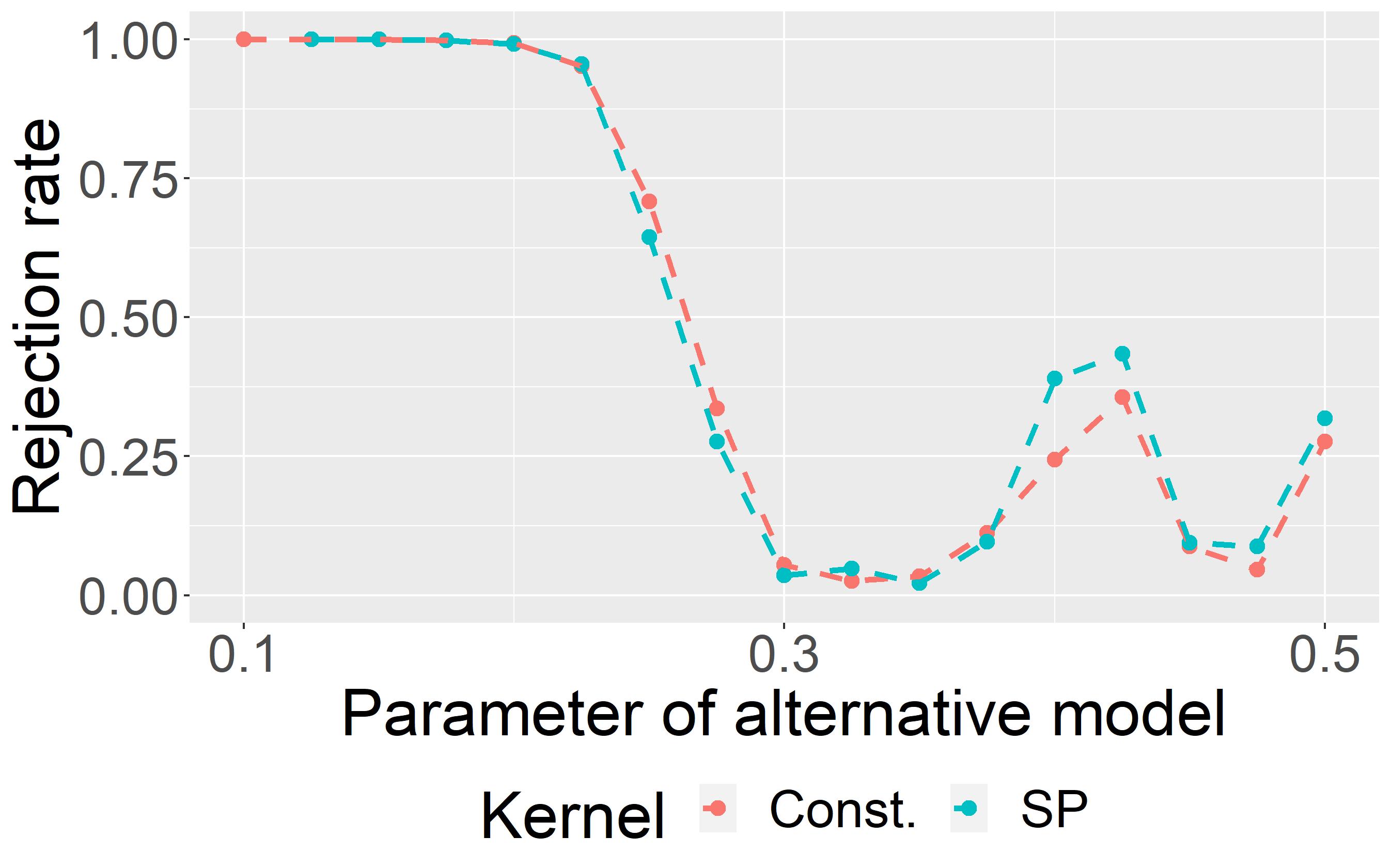}}
\includegraphics[width=0.325\textwidth]{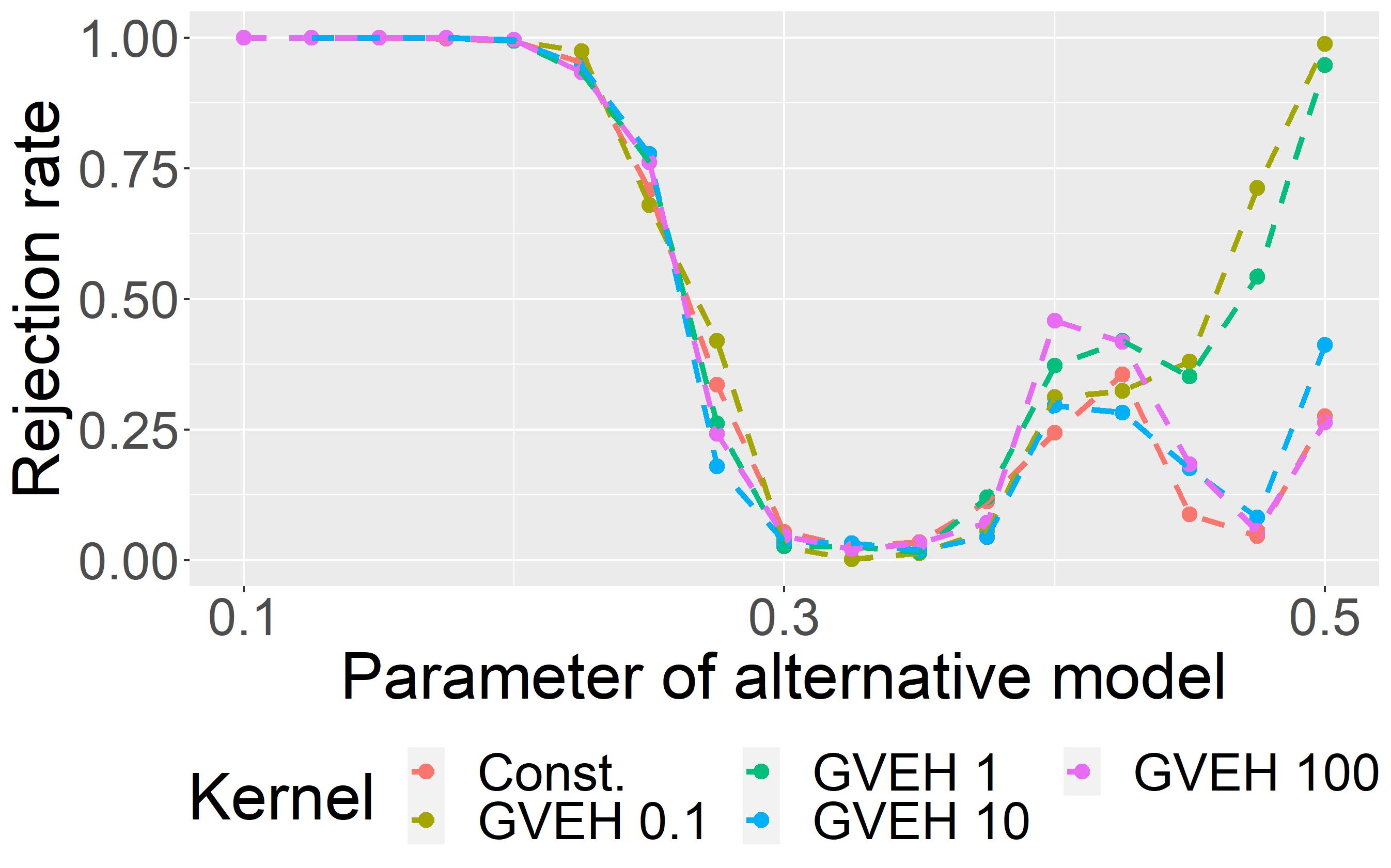}
    {\includegraphics[width=0.32\textwidth]{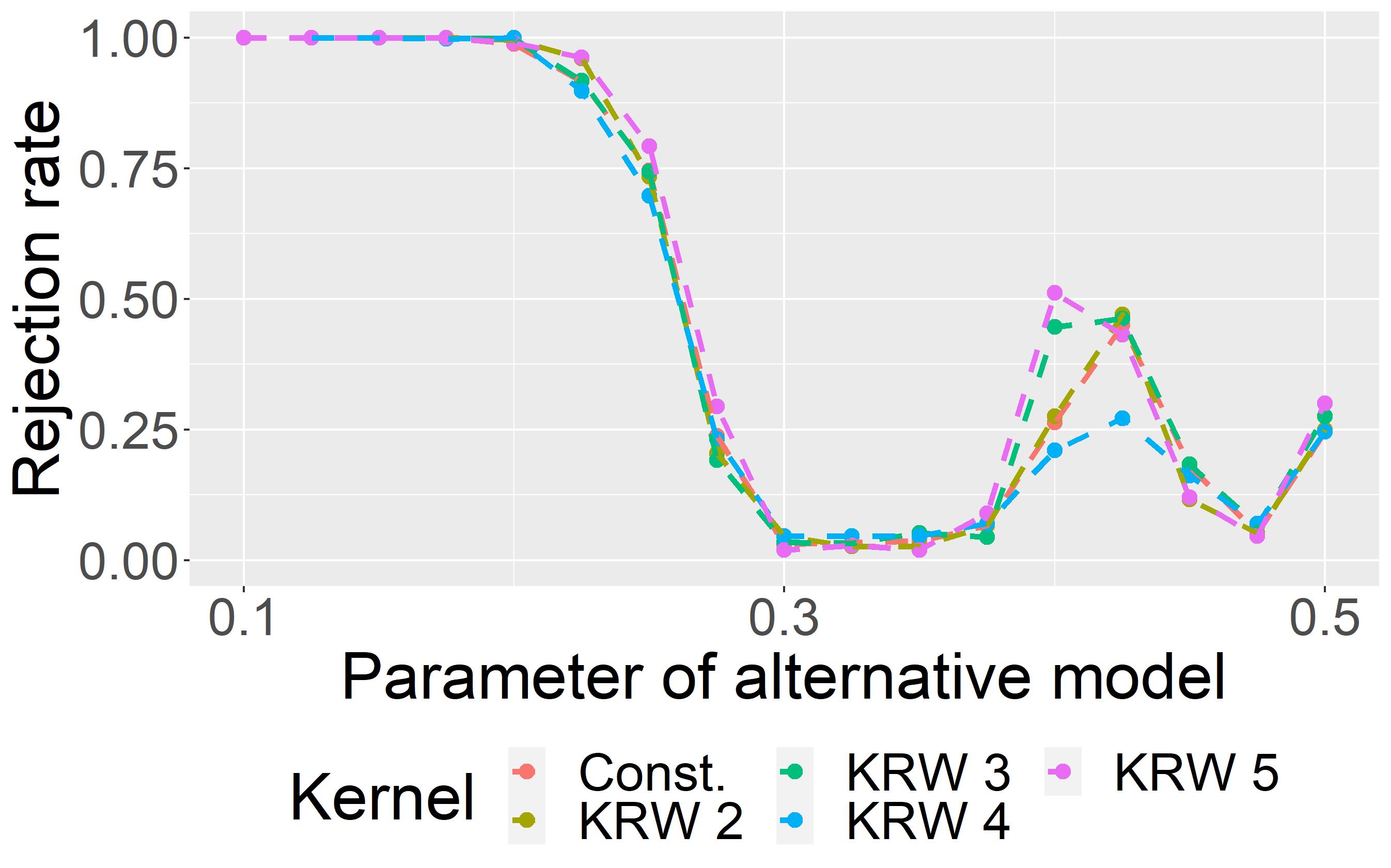}}
    \vspace{-0.23cm}
    \caption{\footnotesize AgraSSt for {the} GRG model with perturbed $r$ alternatives;  $n=20$ and $B=200$;
    $t(x)$ is set to be {the} bivariate (vertex) degree vector.
    }
    \vspace{-0.2cm}
    \label{fig:grg}
\end{figure*}

\paragraph{A geometric random graph example}
{In our} geometric {r}andom graph (GRG) models \citep{penrose2003random}, 
vertices are uniformly placed on a {2-dimensional unit} 
torus and
two vertices are connected by an edge if their distance is smaller than a pre-defined radius
parameter $r$. The null model sets $r=0.3$ where the alternative is set by perturbing $r$.  
{Here we use AgraSSt with some of the summary statistics suggested in \cite{xu2022agrasst}}.   
AgraSSt  
{with {a} WL kernel {as suggested in \citep{xu2022agrasst}}}  achieves well controlled type 1 error in all {presented settings} {in the main text and \Cref{app:exp}}.
When using edge density as summary statistics, 
{\Cref{fig:grg-density} in \Cref{app:exp} shows that} 
{all kernel choices achieve}
good performance, {with 
the graphlet kernels performing best, and the constant kernel and GRW kernels performing better than the WL kernel.} 
{W}hen using {instead} bi-degree as summary statistic for predicting conditional edge probability in AgraSSt (see \Cref{fig:grg}), 
there is a region around $r=0.45$ {in which} 
the test statistics {struggle} 
to distinguish the alternative from the null;
{an explanation can be found in \Cref{app:explain}}. {Here,} GVEH with small bandwidth exihibits the best performance.
%
For GRG {on a square instead of a torus, there is no such ambiguous region;}
%
results are shown in \Cref{app:exp}.

\paragraph{CELL} 
To assess the {effect of} kernel {choice} on {a} \emph{black-box} deep generative model, we {train} 
the   Cross-Entropy Low-rank
Logit (CELL) \citep{rendsburg2020netgan}
{on Zachary's Karate Club network \cite{zachary1977information} using different kernels. 
Model $M1$ is obtained by training CELL on the Karate club network.
Here we take $n_{M1}=100$;  $n_{M0}=1$ as we observe only one network. We repeat {the}  procedure 100 times to obtain average rejection rates. Rejection rates, shown in \Cref{zachary-CELL} in the Appendix, are reasonable for most kernels; GVEH struggles for some of the statistics, and so do KRW and GRW.} 


\paragraph{Runtimes} 
{The runtimes of the algorithms, with their standard implementation from 
the R package \texttt{graphkernels} \citep{sugiyama2018graphkernels},  are shown in \Cref{tab:runtime-sparse} and \Cref{tab:runtime-dense}, for a sparse as well as a dense setting. 
The constant kernel is by far the fastest kernel, followed by WL kernels. The random walk and shortest path kernels take three to 6 orders of magnitude longer to compute than the constant kernel 
{and their runtime is greatly increased by larger edge density}. We note that the runtime depends on the implementation; \Cref{app:runtimeidea} includes an idea for speeding up GRW kernel computations.}  

\section{Conclusion and Discussion} \label{sec:conclusion}
In this work, we explored the effect of kernel 
and parameter choices on KSS for model assessment. Beyond observing {some} case specific phenomen{a} where some choices can outperform others, we also conclude that both gKSS and AgraSSt are fairly robust in producing decent test power under the alternative, using a large class of kernels.
  Overall the constant kernel, {which does not encode network information beyond density,}  performs surprisingly well {as soon as there is a distinguishing signal in the edge density}. Given its clear runtime advantage if density is assumed to be a strong signal then this could 
 be a reasonable kernel to use.
 Also it is reassuring that the WF kernels in AgraSSt perform well. 

If runtime is not an issue, then one may like to combine tests based on suitably selected  kernels as in \cite{schrab2022ksd}, and reject the null hypothesis if any of the single kernel 
tests {distinguishes the alternative from the null hypothesis by rejecting it}.

In future work, it would be interesting to carry out a similar study for other KSD tests. In general then the additional issue of Stein class may feature, as not every kernel choice may yield a RKHS which is a Stein class for the operator. 

{Finally, Stein operators for distributions are not unique 
 \citep{ley2017stein}. Exploring the interplay between Stein operator choice and RKHS is another future research direction.}

\begin{ack}
M.W.'s research is partly funded by
the Studienstiftung des deutschen Volkes. 
W.X. and G.R. acknowledge support from  EPSRC grant EP/T018445/1. G.R. also acknowledges  support from EPSRC grants EP/R018472/1, EP/V056883/1, and EP/W037211/1. 
\end{ack}

\bibliographystyle{apalike}
\bibliography{main}

\begin{thebibliography}{}

\bibitem[Ahmed et~al., 2017]{ahmed2017graphlet}
Ahmed, N.~K., Neville, J., Rossi, R.~A., Duffield, N.~G., and Willke, T.~L.
  (2017).
\newblock Graphlet decomposition: Framework, algorithms, and applications.
\newblock {\em Knowledge and Information Systems}, 50(3):689--722.

\bibitem[Bhamidi et~al., 2011]{bhamidi2011mixing}
Bhamidi, S., Bresler, G., and Sly, A. (2011).
\newblock Mixing time of exponential random graphs.
\newblock {\em The Annals of Applied Probability}, 21(6):2146--2170.

\bibitem[Borgwardt and Kriegel, 2005]{borgwardt2005shortest}
Borgwardt, K.~M. and Kriegel, H.-P. (2005).
\newblock Shortest-path kernels on graphs.
\newblock In {\em Fifth IEEE International Conference on Data Mining
  (ICDM'05)}, pages 8--pp. IEEE.

\bibitem[Chatterjee and Diaconis, 2013]{chatterjee2013estimating}
Chatterjee, S. and Diaconis, P. (2013).
\newblock Estimating and understanding exponential random graph models.
\newblock {\em The Annals of Statistics}, 41(5):2428--2461.

\bibitem[Chwialkowski et~al., 2016]{chwialkowski2016kernel}
Chwialkowski, K., Strathmann, H., and Gretton, A. (2016).
\newblock A kernel test of goodness of fit.
\newblock In {\em JMLR: Workshop and Conference Proceedings}.

\bibitem[Clauset et~al., 2004]{clauset2004cluster}
Clauset, A., Newman, M. E.~J., and Moore, C. (2004).
\newblock Finding community structure in very large networks.
\newblock {\em Physical Review E}, 70(6).

\bibitem[Cormen et~al., 2022]{cormen2022introduction}
Cormen, T.~H., Leiserson, C.~E., Rivest, R.~L., and Stein, C. (2022).
\newblock {\em Introduction to {A}lgorithms}.
\newblock MIT press.

\bibitem[Frank and Strauss, 1986]{frank1986markov}
Frank, O. and Strauss, D. (1986).
\newblock Markov graphs.
\newblock {\em Journal of the American Statistical Association},
  81(395):832--842.

\bibitem[G{\"a}rtner et~al., 2003]{gartner2003graph}
G{\"a}rtner, T., Flach, P., and Wrobel, S. (2003).
\newblock On graph kernels: Hardness results and efficient alternatives.
\newblock In {\em Learning Theory and Kernel Machines}, pages 129--143.
  Springer.

\bibitem[Girvan and Newman, 2002]{girvan2002community}
Girvan, M. and Newman, M. E.~J. (2002).
\newblock Community structure in social and biological networks.
\newblock {\em Proceedings of the National Academy of Sciences},
  99(12):7821--7826.

\bibitem[Gorham et~al., 2020]{gorham2020stochastic}
Gorham, J., Raj, A., and Mackey, L. (2020).
\newblock Stochastic stein discrepancies.
\newblock In {\em Advances in Neural Information Processing Systems},
  volume~33, pages 17931--17942.

\bibitem[Holland and Leinhardt, 1981]{holland1981exponential}
Holland, P.~W. and Leinhardt, S. (1981).
\newblock An exponential family of probability distributions for directed
  graphs.
\newblock {\em Journal of the American Statistical Association},
  76(373):33--50.

\bibitem[Hunter et~al., 2008a]{hunter2008goodness}
Hunter, D.~R., Goodreau, S.~M., and Handcock, M.~S. (2008a).
\newblock Goodness of fit of social network models.
\newblock {\em Journal of the American Statistical Association},
  103(481):248--258.

\bibitem[Hunter et~al., 2008b]{morris2008ergm}
Hunter, D.~R., Handcock, M.~S., Butts, C.~T., Goodreau, S.~M., and Morris, M.
  (2008b).
\newblock ergm: A package to fit, simulate and diagnose exponential-family
  models for networks.
\newblock {\em Journal of Statistical Software}, 24(3):nihpa54860.

\bibitem[Kriege et~al., 2016]{kriege2016valid}
Kriege, N.~M., Giscard, P.-L., and Wilson, R. (2016).
\newblock On valid optimal assignment kernels and applications to graph
  classification.
\newblock In {\em Advances in Neural Information Processing Systems}, pages
  1623--1631.

\bibitem[Kriege et~al., 2020]{kriege2020survey}
Kriege, N.~M., Johansson, F.~D., and Morris, C. (2020).
\newblock A survey on graph kernels.
\newblock {\em Applied Network Science}, 5(1):1--42.

\bibitem[Ley et~al., 2017]{ley2017stein}
Ley, C., Reinert, G., and Swan, Y. (2017).
\newblock Stein’s method for comparison of univariate distributions.
\newblock {\em Probability Surveys}, 14:1--52.

\bibitem[Liu et~al., 2016]{liu2016kernelized}
Liu, Q., Lee, J., and Jordan, M. (2016).
\newblock A kernelized {S}tein discrepancy for goodness-of-fit tests.
\newblock In {\em International Conference on Machine Learning}, pages
  276--284.

\bibitem[Penrose, 2003]{penrose2003random}
Penrose, M. (2003).
\newblock {\em Random geometric graphs}.
\newblock Oxford University Press.

\bibitem[Reinert and Ross, 2019]{reinert2019approximating}
Reinert, G. and Ross, N. (2019).
\newblock Approximating stationary distributions of fast mixing {G}lauber
  dynamics, with applications to exponential random graphs.
\newblock {\em The Annals of Applied Probability}, 29(5):3201--3229.

\bibitem[Rendsburg et~al., 2020]{rendsburg2020netgan}
Rendsburg, L., Heidrich, H., and Von~Luxburg, U. (2020).
\newblock Net{GAN} without {GAN}: From random walks to low-rank approximations.
\newblock In {\em International Conference on Machine Learning}, pages
  8073--8082. PMLR.

\bibitem[Schrab et~al., 2022]{schrab2022ksd}
Schrab, A., Guedj, B., and Gretton, A. (2022).
\newblock {KSD} aggregated goodness-of-fit test.
\newblock {\em arXiv preprint arXiv:2202.00824}.

\bibitem[Sherman and Morrison, 1950]{sherman1950adjustment}
Sherman, J. and Morrison, W.~J. (1950).
\newblock Adjustment of an inverse matrix corresponding to a change in one
  element of a given matrix.
\newblock {\em The Annals of Mathematical Statistics}, 21(1):124--127.

\bibitem[Shervashidze et~al., 2011]{shervashidze2011weisfeiler}
Shervashidze, N., Schweitzer, P., Leeuwen, E. J.~v., Mehlhorn, K., and
  Borgwardt, K.~M. (2011).
\newblock Weisfeiler-{L}ehman graph kernels.
\newblock {\em Journal of Machine Learning Research}, 12(Sep):2539--2561.

\bibitem[Shervashidze et~al., 2009]{shervashidze2009efficient}
Shervashidze, N., Vishwanathan, S., Petri, T., Mehlhorn, K., and Borgwardt, K.
  (2009).
\newblock Efficient graphlet kernels for large graph comparison.
\newblock In {\em Artificial Intelligence and Statistics}, pages 488--495.
  PMLR.

\bibitem[Sugiyama and Borgwardt, 2015]{sugiyama2015halting}
Sugiyama, M. and Borgwardt, K. (2015).
\newblock Halting in random walk kernels.
\newblock In {\em Advances in Neural Information Processing Systems}, pages
  1639--1647.

\bibitem[Sugiyama et~al., 2018]{sugiyama2018graphkernels}
Sugiyama, M., Ghisu, M.~E., Llinares-L{\'o}pez, F., and Borgwardt, K. (2018).
\newblock graphkernels: {R} and {P}ython packages for graph comparison.
\newblock {\em Bioinformatics}, 34(3):530--532.

\bibitem[Wasserman and Faust, 1994]{wasserman1994social}
Wasserman, S. and Faust, K. (1994).
\newblock {\em Social {N}etwork {A}nalysis: {M}ethods and {A}pplications}.
\newblock Cambridge University Press.

\bibitem[Xu and Reinert, 2021]{xu2021stein}
Xu, W. and Reinert, G. (2021).
\newblock A {S}tein goodness-of-test for exponential random graph models.
\newblock In {\em International Conference on Artificial Intelligence and
  Statistics}, pages 415--423. PMLR.

\bibitem[Xu and Reinert, 2022]{xu2022agrasst}
Xu, W. and Reinert, G. (2022).
\newblock Agra{S}{S}t: Approximate graph stein statistics for interpretable
  assessment of implicit graph generators.
\newblock {\em arXiv preprint arXiv:2203.03673}.

\bibitem[Zachary, 1977]{zachary1977information}
Zachary, W.~W. (1977).
\newblock An information flow model for conflict and fission in small groups.
\newblock {\em Journal of Anthropological Research}, 33(4):452--473.

\end{thebibliography}


\clearpage

\appendix

\section{Additional details on background}\label{app:background}

\subsection{{The} graph kernel Stein statistic (gKSS)}
The graph Kernel Stein statistic (gKSS) has been proposed to {assess goodness of fit for}  the {family of} exponential random graph models (ERGMs).
ERGMs  are frequently used as parametric models for {social} network analysis \citep{wasserman1994social, holland1981exponential, frank1986markov}; 
{they include Bernoulli random graphs as well as stochastic blockmodels as special cases}. 
Here we restrict attention to undirected, unweighted simple graphs {on $n$ vertices}, 
{without self-loops or multiple edges.}
To define {such} an ERGM, we introduce the following notations.

Let $\G^{lab}_n$ be a set of vertex-labeled  graphs on $n$ vertices and,
for {$N =n(n-1)/2 $,}
{encode} $x \in \G^{lab}_n$ by an ordered collection of $\{0,1\}$ valued variables $x = (x_{ij})_{1 \le  i < j \le n} \in \{0,1\}^N$ {where} $x_{ij}=1$ {if and only if}
there is an edge between $i$ and $j$. 
{For a graph $H$ on at most $n$ vertices, let $V(H)$ denote the vertex set, 
and for $x\in\{0,1\}^N$, denote by $t(H,x)$  the number of
{\it edge-preserving} injections from $V(H)$ to $V(x)$; an injection $\sigma$ preserves edges if for all edges $vw$ of $H$ {with  $\sigma(v)<\sigma(w)$}, $x_{\sigma(v)\sigma(w)}=1$.
For  $v_H =| V(H)| \ge 3$  set 
$$
t_H(x)=\frac{t(H,x)}{n(n-1)\cdots(n-v_H+3)}.
$$
If $H{=H_1}$ is a single edge, then $t_H(x)$ is twice the number of edges of~$x$. In the exponent this scaling of counts matches \cite[Definition~1]{bhamidi2011mixing} and \cite[Sections~3 and~4]{chatterjee2013estimating}.
} 
An ERGM {
for the collection  $x\in \{0,1\}^{N}$
can be defined
} as follows, see  \cite{reinert2019approximating}.

\begin{Definition}
\label{def:ergm} 
Fix $n\in \N$ and $c \in \N$. {{L}et $H_1$ be a single edge {and f}or $l={2}, \ldots, c$ let}  $H_l$ be a connected graph on at most $n$ vertices; 
set $t_l(x) = t_{H_l}(x)$. For  $\beta = (\beta_1, \dots, \beta_c)^{\Trans} {\in \R^c}$ 
and
$t(x) =(t_1(x),\dots,t_c(x))^{\Trans} \in \R^c$ 
$X\in \G^{lab}_n$ follows  the exponential random graph model  $X\sim \operatorname{ERGM}(\beta, t)$ if for  $\forall x\in \G^{lab}_n$,
$$
    \P(X = x) = \frac{1}{\kappa_n(\beta)}\exp{\left(\sum_{l=1}^{c} \beta_l t_l(x) \right)}.
$$
Here $\kappa_n(\beta)$ is the normalisation constant.
\end{Definition}
{
The vector $\beta \in \R^k$ is the parameter vector and the statistics $t(x) =(t_1(x),\dots,t_c(x))^{\Trans} \in \R^c$  are sufficient statistics. 
}

Many random graph models can be set in this framework. The simplest example is the Bernoulli random graph (ER graph) {with edge probability $0 < p < 1$; in this case},  {$l=1$ and $H_1$ is a single edge}.
{ERGMs can use other statistic in addition to subgraph counts, and many ERGMs model directed networks.} 
Moreover, ERGM{s} can  model network with covariates such as using dyadic statistics to model group interactions between vertices \citep{hunter2008goodness}. 
{Here we restrict attention to the case which is treated 
in \cite{reinert2019approximating} {because}  it is for this case that a Stein characterization {is available}.
}

As the network size increases, the number of possible network configurations increases exponentially {in the number of possible edges}, making the normalisation constant $\kappa_n(\beta)$ {usually} prohibitive to compute in closed form.
{Classical} statistical inference on ERGM mainly relies on  MCMC type  methods that utilise the density ratio between proposed state and current state, where the normalisation constant cancels.
{However the Stein score-function operator framework does not require a normalising constant. In \cite{reinert2019approximating} a Stein operator for an ERGM is obtained which is of the form $\T_q f 
= \frac{1}{N}\sum_{s\in [N]} \T^{(s)}_q f$ where}
{the components of the Stein operator} 
{are} 
\begin{eqnarray}
 {\A_q^{(s)} f(x)}  \nonumber
 & \hspace{-0.2cm} = & \hspace{-0.13cm} q(x^{(s,1)}|{x_{-s}} )f (x^{(s,1)}) + q(x^{(s,0)}|{x_{-s}} )  f (x^{(s,0)})  - f(x) \nonumber\\
 & \hspace{-0.2cm} = & \hspace{-0.13cm} \E_{\{0,1\}}[f(X^{s},x_{-s}) | {x_{-s}} )] - f(x).\label{eq:stein_conditional_difference}
\end{eqnarray}
{Here $N = 
n(n-1)/2$ is the total number edges; $[N]:=\{1, \ldots, N\}$ denotes the 
set of vertex pairs; $x^{(s,1)}$ has the $s$-entry replaced of $x$ by $1$; 
$x_{-s}$ is the network $x$ with edge index $s$ removed, and} 
 $ \E_{\{0,1\}}$ refers to the expectation taken only over the value, {0 or 1,}  which ${X}_s$ takes on.
Hence, {with $S$ chosen uniformly at random from $[N]$, independently of all other variables,} 
\begin{align}\label{eq:stein_expectation}
\A_q f(x) 
&= \E_S \left[\E_{\{0,1\}}[f(X^{s},x_{-s}|x)]\right] - f(x)
. 
\end{align}

{It is easy to see that $ \E_q \A_q f(x) =0$ for all finite functions $f:\mathcal{G}_{n}^{lab} \to \R$.} 
Let $\H$ denote a RKHS with kernel $k$ and inner product $\langle \cdot, \cdot \rangle$. 
For a fixed {network}  $x$, we {next}  seek a function $f \in \H$, s.t. $\|f\|_{\H}\leq 1$, that best distinguishes the difference in Eq.\eqref{eq:stein_expectation} {when $X$ does not have distribution $q$.}
We
 define the {\it graph kernel Stein statistics} (gKSS) as
\begin{align}\label{eq:gkss}
    \operatorname{gKSS}(q;x) 
    & = \sup_{\|f\|_{\H}\leq 1} \Big|\E_S[\A^{(S)}_q f(x)] \Big|.
\end{align}
{It is often more convenient to consider $\operatorname{gKSS}^2(q;x) $. 
By the reproducing property of RKHS functions,  algebraic manipulation allows  the supremum to be computed in closed form:
\begin{align}\label{eq:gkss_quadratic_form}
{\operatorname{gKSS}}^2(q;x) =  \frac{1}{N^2} \sum_{s, s'\in [N]} h_x(s, s')
\end{align}
where 
$h_x(s, s') = \left\langle \A^{(s)}_q k(x,\cdot), \A^{(s')}_q k(\cdot,x)\right\rangle
.
$
}

{When the distribution of $X$ is known, the} expectation in Eq.\eqref{eq:stein_expectation} can be computed for networks with a small number of vertices, but {when the number of vertices is large, exhaustive evaluation is computationally intensive.}
For a fixed network $x$, 
{\cite{xu2021stein}} propose the following randomised Stein operator via edge re-sampling.
Let $B$ be the {fixed} number of edges to be re-sampled. {The}  re-sampled Stein operator is
\begin{equation}\label{eq:resample_kss}
    \widehat{\A_q^B} f(x) = \frac{1}B \sum_{b\in [B]} \A^{(s_b)}_q f(x)
\end{equation}
where ${b \in B}$ and $s_b$ are edge samples from $\{1, \ldots, N\}$, {chosen uniformly with replacement, independent of each other and of $x$.}
The  expectation of  $ \widehat{\A_q^B} f(x)$ with respect to the re-sampling is 
$$
\E_B [ \widehat{\A_q^B} f(x)  ]
{ = \E_S [ \A_q^{(S)} f(x)]} 
= \A_q f(x) 
$$
{with corresponding} 
 {re-sampling} gKSS 
\begin{align}\label{eq:gkss_resample}
    \widehat{\operatorname{gKSS}}(q;x) = \sup_{\|f\|_{\H}\leq 1} \Big|\frac{1}{B}\sum_{b\in [B]}\A^{(s_b)}_q f(x) \Big|.
\end{align}
This is a stochastic Stein discrepancy, see  \cite{gorham2020stochastic}. 
{{The supremum in Eq.\eqref{eq:gkss_resample} is achieved by} $$
f^*(\cdot) = \frac{\frac{1}{B}\sum_b \A_q^{(s_b)}k(x,\cdot)}{\left\| \frac{1}{B}\sum_a \A_q^{(s_a)}k(x,\cdot) \right\|}.
$$} 
Similar algebraic manipulations as for Eq.\eqref{eq:gkss_quadratic_form} {yield} 

\begin{align}\label{eq:gkss_resample_quadratic_form}
    \widehat{\operatorname{gKSS}}^2(q;x) =  \frac{1}{B^2} \sum_{b, b'\in [B]} h_x(s_b, s_{b'}).
\end{align}

\begin{algorithm}[th]
   \caption{Kernel Stein Test for ERGM}
   \label{alg:kernel_stein_monte_carlo}
\begin{algorithmic}[1]
\renewcommand{\algorithmicrequire}{\textbf{Input:}}
\renewcommand{\algorithmicensure}{\textbf{Objective:}}
\REQUIRE~~\\
    Observed network $x$; null model $q$; 
    RKHS kernel $k$; \\
    Re-sample size $B$; 
    confidence level $a$;
    {number of simulated networks} $l$;
\ENSURE~~\\
Test $H_0: x\sim q$ versus $H_1: x \not\sim q$.
\renewcommand{\algorithmicensure}{\textbf{Test procedure:}}
\ENSURE~~\\
\STATE Sample $\{s_1,\dots,s_B\}$ with replacement uniformly from $[N]$.
\STATE Compute $\tau =\widehat{\operatorname{gKSS}}^2(q;x)$ in Eq.\eqref{eq:gkss_resample_quadratic_form}.
\STATE Simulate $z_1,\dots,z_l \sim q$.
\STATE Compute $\tau_i =\widehat{\operatorname{gKSS}}^2(q;z_i)$ in Eq.\eqref{eq:gkss_resample_quadratic_form}. 
{again with re-sampling, choosing new samples 
 $\{s_{1, i},\dots,s_{B,i}\}$ uniformly from $[N]$} with replacement.
\STATE {Estimate} the {empirical}  $(1-a)$-quantile $c_{1-a}$ of $\tau_1,\dots,\tau_l$.
\renewcommand{\algorithmicrequire}{\textbf{Output:}}
\REQUIRE~~\\
Reject $H_0$ if $\tau > c_{1-a}$; otherwise do not reject.
\end{algorithmic}
\end{algorithm}
 
The ERGM can be readily simulated from an unnormalised density via MCMC, {see for example}  \cite{morris2008ergm}. {Suppose that $q$ is the distribution of ERGM$(\beta, t)$ and $x$ is the observed network for which we want to assess the fit to $q$}.

Then gKSS in Eq.\eqref{eq:gkss} captures the optimised Stein features over RKHS functions, which is a  comprehensive non-parametric summary statistics.
Let $z_1,\dots,z_n \sim p$ be simulated networks from the null distribution. For test function $f$, the Monte-Carlo test is to compare $f(x)$ against $f(z_1),\dots, f(z_n)$ and the p-value can be determined accordingly.
{The detailed gKSS algorithm is shown in \Cref{alg:kernel_stein_monte_carlo}.} 

\subsection{Approximate graph Stein statistics (AgraSSt)}\label{app:agrasst}

\paragraph{Approximate Stein operators}
Recall the Stein operator for ERGMs in Eq.\eqref{eq:stein_conditional_difference}, 
which depends on the conditional probabilit{ies} $q(x^{(s,1)}|{x_{-s}} )$ and $q(x^{(s,0)}|{x_{-s}} )$.
For implicit models and graph generators $G$, the conditional probabilit{ies}
required in the Stein operator $\mathcal{T}_{q}^{(s)}$
in Eq.\eqref{eq:stein_conditional_difference}
cannot be obtained without explicit knowledge of $q(x)$.  
Instead, {\cite{xu2022agrasst}} consider summary statistic $t(x)$ and the probabilities conditioned on $t(x)$, $$q_t(x^{(s,1)}):= \mathbb{P} (X^s = 1 | t(x_{{-s}}))$$ {(and analogously $q_t(x^{(s,0)})$), interpreting $q_t(x^{(s,\cdot)})$ }  as {a}  \emph{discrete score} {function}. The corresponding Stein operator based on $t(x)$ is defined {in \cite{xu2022agrasst}} as 
\begin{align*}
\mathcal{A}_{q,t}^{(s)} f(x) 
 =  q_t(x^{(s,1)})f( x^{(s,1)})  + q_t(x^{(s,0)})f(x^{(s,0)})  - f(x).
\end{align*}

{Given a large number for samples from the graph generator $G$}, 
the conditional edge probabilities 
$q_t(x^{(s,1)})$ can be estimated.
Using the Stein operator for conditional graph distributions, {\cite{xu2022agrasst}}  obtain the approximate Stein operators Eq.\eqref{eq:approx_cond_stein} and \eqref{eq:approx_stein} for an implicit graph generator $G$ by estimating 
$q_t(x^{(s,1)})$. 
Here $t(x)$ are user-defined statistics.
{In principle, any multivariate statistic $t(x)$ can be used in this formalism. However, estimating the conditional probabilities using relative frequencies {can} 
be computationally prohibitive when the graphs are very large and specific frequencies are rarely observed. Instead, {\cite{xu2022agrasst}}  consider simple summary statistics, such as} edge density which corresponds to $t=0$, {the bi}degree statistics or {the} number of neighbours connected to both vertices of $s$.  
{Here for a vertex pair $s=(i,j)$, with $deg_{-s}(i)$ denoting the degree of $i$ in the network $x$ with $j$ removed, the bidegree statistic is  $t(x_{-s}) = (deg_{-s}(i), deg_{-s}(j))$. The common neighbour statistic is  $t(x_{-s}) = {\vert \{ v \in V \mid v \text{ is neighbour of } i \text{ and } j \text{ in } x\}\vert}$.} 


AgraSSt {performs} 
model assessment using {an operator which approximates the} 
Stein operator
$\A_{q,t}^{(s)}$. 
{We define t}he approximate Stein operator for {the} conditional random graph by 
\begin{align}
\A_{\q,t}^{(s)} f(x) 
 =  \q_t(x^{(s,1)})f(x^{(s,1)})  + \q_t(x^{(s,0)})f( x^{(s,0)})  - f(x).
\label{eq:approx_cond_stein}
\end{align}
The 
{vertex-pair} averaged {approximate} Stein operator is
\begin{equation}\label{eq:approx_stein}
\A_{\widehat q, t} f(x) = \frac{1}{N} \sum_{s \in [N]} \A_{\q,t}^{(s)} f(x).
\end{equation}

{AgraSSt for implicit graph generators
} {is then defined in analogy to} 
gKSS in Eq.\eqref{eq:gkss}, as
\begin{align*}\label{eq:AgraSSt}
    \operatorname{AgraSSt}(\widehat q, t;x) 
 = \sup_{\|f\|_{\H}\leq 1} \Big|\frac{1}{N}\sum_s \A^{(s)}_{\widehat q,t} f(x)\Big|.
\end{align*}

\paragraph{Re-sampling Stein statistic}
Similar to Eq.\eqref{eq:resample_kss}, a computationally efficient operator for large $N$ 
is  derived  {in \cite{xu2022agrasst}} via re-sampling $B$ 
vertex-pairs $s_b$,  $b=1, \ldots, B$, from $\{1, \ldots, N\}$, {chosen uniformly with replacement, independent of each other and of $x$,}
which creates a randomised operator
to be re-sampled.
The re-sampled  
operator is
$${\widehat\A_{\q,t}^B} f(x) = \frac{1}B \sum_{b\in [B]} \A^{(s_b)}_{\q,t} f(x).$$
The  expectation of  $ {\widehat\A_{\q,t}^B} f(x)$ with respect to re-sampling is 
$
\E_B [ {\widehat\A_{\q,t}^B} f(x)  ]
{ = \E_S [ \A_{\q,t}^{(S)} f(x)]} 
= \A_{\q,t} f(x). 
$
The corresponding re-sampled AgraSSt is
\begin{align*}
    \widehat{\operatorname{AgraSSt}}(\widehat q, t;x) = \sup_{\|f\|_{\H}\leq 1} \Big|\frac{1}{B}\sum_{b\in [B]}\A^{(s_b)}_{\widehat q,t} f(x) \Big|.
\end{align*}

{Similar to Eq.\eqref{eq:gkss_resample_quadratic_form}, the squared version of $\widehat{\operatorname{AgraSSt}}$ admits the following quadratic form,
\begin{align*}
    \widehat{\operatorname{AgraSSt}}^2({\widehat q,t};x) = \frac{1}{B^2}\sum_{b, b'\in [B]} \widehat h_x(s_b, s_{b'}),
\end{align*}
where 
$\widehat h_x(s, s') = \langle \A^{(s)}_{\widehat q,t} k(x,\cdot), \A^{(s')}_{\widehat q,t} k(\cdot,x)\rangle_{\H}.$
}
{
{Theoretical guarantees for this operator are given in} 
in \citet{xu2021stein}.}


\subsection{Graph kernels}\label{app:graph_kernel}
For a vertex-labeled graph {$x = \{ x_{ij} \}_{1 \le i , j \le n} \in \mathcal{G}^{lab}$}, with label range { $\{1, \ldots ,c \} = [c]$}, denote the vertex set by $V$ {and} the edge set by $E$. {With abuse of notation we write $x=(V(x), E(x))$.} 
Consider {a} vertex-edge mapping $\psi  :V  \cup E  \rightarrow [c]$. {In this paper we  use the following graph kernels.}

\paragraph{Gaussian vertex-edge  histogram 
graph {kernels}}
{The} {vertex-edge label histogram} 
$h = (h^{111},h^{211},\dots, h^{ccc})$  has as components
$$h^{l_1l_2l_3} {(x)}  = \left|\{v\in V {(x)} , (v,u)\in E {(x)}  \, | \,  \psi(v,u) = l_1, \psi(u)= l_2,\psi(v)=l_3\}\right|,$$ for $l_1, l_2, l_3 \in [c]$;  {it is a combination of vertex label counts and edge label counts.} 
{Let 
$\langle h (x) , h (x') \rangle = \sum_{l_1,l_2,l_3} h (x)^{l_1,l_2,l_3}h {(x')}^{l_1,l_2,l_3}$.
{Following} \cite{sugiyama2015halting},}
the Gaussian Vertex-Edge  Histogram (GVEH)
graph kernel between two graphs
{$x, x'$} 
is defined as 
$$
k_{\small GVEH}(x, x';\sigma) = \exp{\left\{-\frac{\|h(x) - h(x')\|^2}{2\sigma^2}\right\}}. 
$$
{The GVEH kernel is a special case of histogram-based kernels}  for assessing graph similarity {using feature maps, which are introduced} in \cite{kriege2016valid}.  {Adding a Gaussian RBF as in} \cite{sugiyama2015halting}, {yielding the GVEH kernel, significantly improved problems such as classification accuracy, see} 
\citep{kriege2020survey}.
{In our implementation, as in \cite{sugiyama2018graphkernels}, $\psi$ is induced by the vertex index. If the vertices are indexed by $i \in [n]$  then  the label of vertex $v_i$ is 
$\psi(v_i)=i$; for edges,  $\psi(u,v)=1$ if $(u,v)\in E$ is an edge and $0$ otherwise.}

\paragraph{Random walk graph 
kernels} 
A $K$-step random walk (KRW) graph kernel \citep{sugiyama2015halting} is built as follows. Take $A_{\otimes}$
as the adjacency matrix of the direct (tensor) product $G_{\otimes} = (V_{\otimes},E_{\otimes},\psi_{\otimes})$ \citep{gartner2003graph} between $x$ and $x'$ such that {vertex labels match and edge labels match:} 
$$
V_{\otimes} = \{(v,v')\in V \times V' \mid \psi(v) = \psi'(v')\},
$$
$$
E_{\otimes} = \{((v,u),(v',u')))\in E \times E'\,  \mid  \, \psi (v,u) = \psi (v',u')\},
$$
and {use} the corresponding label mapping
$\psi_{\otimes}(v,v') = \psi(v) = \psi'(v')
$; $\psi_{\otimes}((v,v'),(u,u')) = \psi(v,u) = \psi'(v',u')
$. {With}  input parameters $(\lambda_0, \dots, \lambda_K)$, {the} $K$-step random walk kernel between two graphs
{$x, x'$}  is defined as 
$$
k_{\otimes}^{(K)}(x,x') = \sum_{i,j=1}^{ |V_{\otimes}|}\left[\sum_{t=0}^K \lambda_t A_{\otimes}^{\top}\right]_{i,j}.
$$

A geometric random walk (GRW) kernel between two graphs
{$x, x'$}  takes the $\lambda$-weighted infinite sum from the random walk:
$$
k_{GRW}(x, x' ) =  \sum_{i,j=1}^{ |V_{\otimes}|} \left[(I - \lambda A_{\otimes})^{-1}\right]_{i, j} . 
$$
{In our implementation we choose, $\lambda_l = \lambda, \forall l=1,\dots, k$ and $\lambda=\frac{1}{3}$.}

\paragraph{Shortest path 
graph kernels} {Introduced by  \cite{borgwardt2005shortest}, the shortest path (SP) kernels are based on a Floyd transformation of the graph $x$. 
The Floyd
transformation $F$  turns the original graph into the so-called shortest-path graph $y=F(x)$;  the  graph $y$ is a complete graph with vertex set $V$ with each edge labelled by the shortest distance in $x$ between the vertices on either end of the edge.} 
{For two networks  $x$ and $x'$}  the 1-step random walk kernel { $k^1_{\otimes}$} between the shortest-path graphs {$y = F(x)$ and $y'=F(x')$ } gives the  shortest-path (SP) kernel between $x$ and $x'$;  
$$
k_{SP}(x,x') = k^1_{\otimes}(y, y').
$$
 Lemma 3  in \cite{borgwardt2005shortest} showed that this kernel is  positive definite. 
 

\paragraph{Weisfeiler-Lehman 
graph kernels} Weisfeiler-Lehman (WL) graph kernels {have been}  proposed by \cite{shervashidze2011weisfeiler}; {these kernels are based on the Weisfeiler-Lehman test for graph isomorphisms and involve}   counting  matching subtrees between two given graphs. Theorem 3 in \cite{shervashidze2011weisfeiler} showed the positive definiteness of {these}  kernels.  {In our implementation, we adapted an} efficient implementation from the $\mathtt{graphkernel}$ package \citep{sugiyama2018graphkernels}. 


\section{Additional experiments and discussions}\label{app:exp}

\subsection{Power performance}

{Here we}  provide {further} {results}
in {addition}
to
the experiments {presented} in the main text. 
All the experiments shown in this section {are} based on test level $a = 0.05$, network size $n=20$ and  re-sample size $B=200$. For both gKSS and AgraSSt tests, we obtain $n_{M1} = 100$ trials for each setting to obtain the rejection rates. 
For AgraSSt, we simulate $n_{M0} = 100$ to estimate the conditional distribution $\q_t$. The Monte Carlo sample size $l=200$ are used to simulated the null distribution.

\begin{figure*}[ht!]
    \centering
    \includegraphics[width=0.33\textwidth]{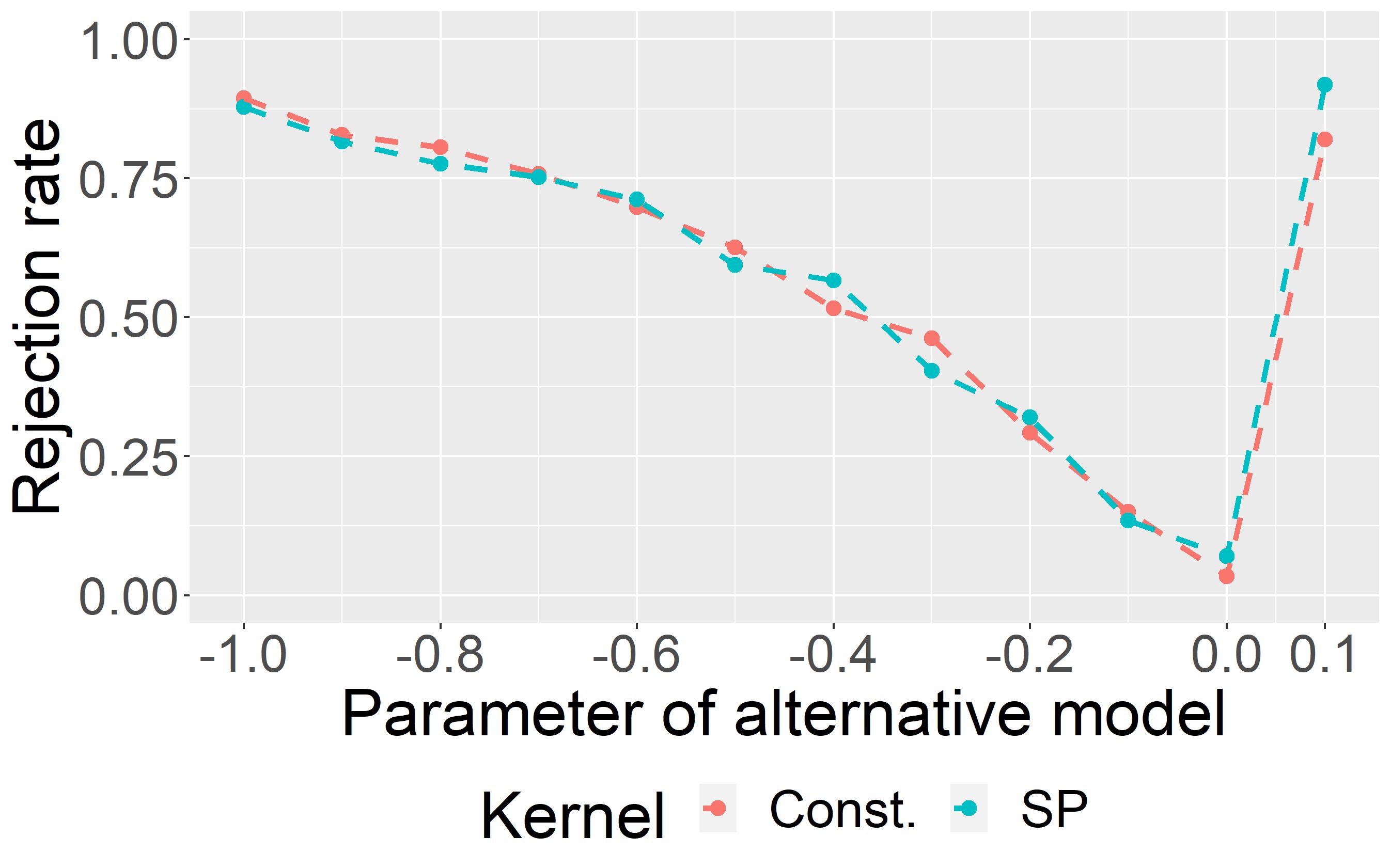}{\includegraphics[width=0.33\textwidth]{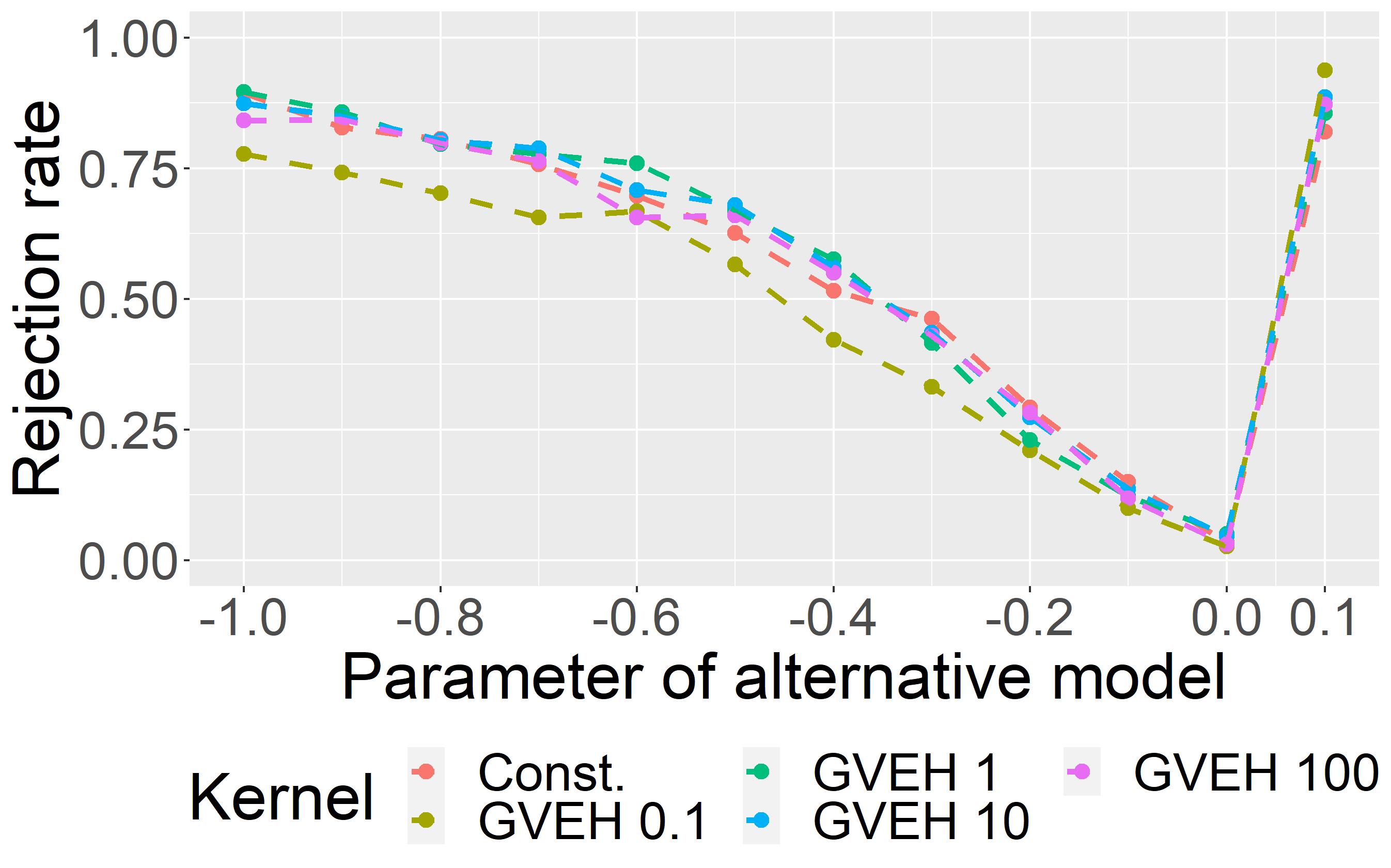}}\includegraphics[width=0.33\textwidth]{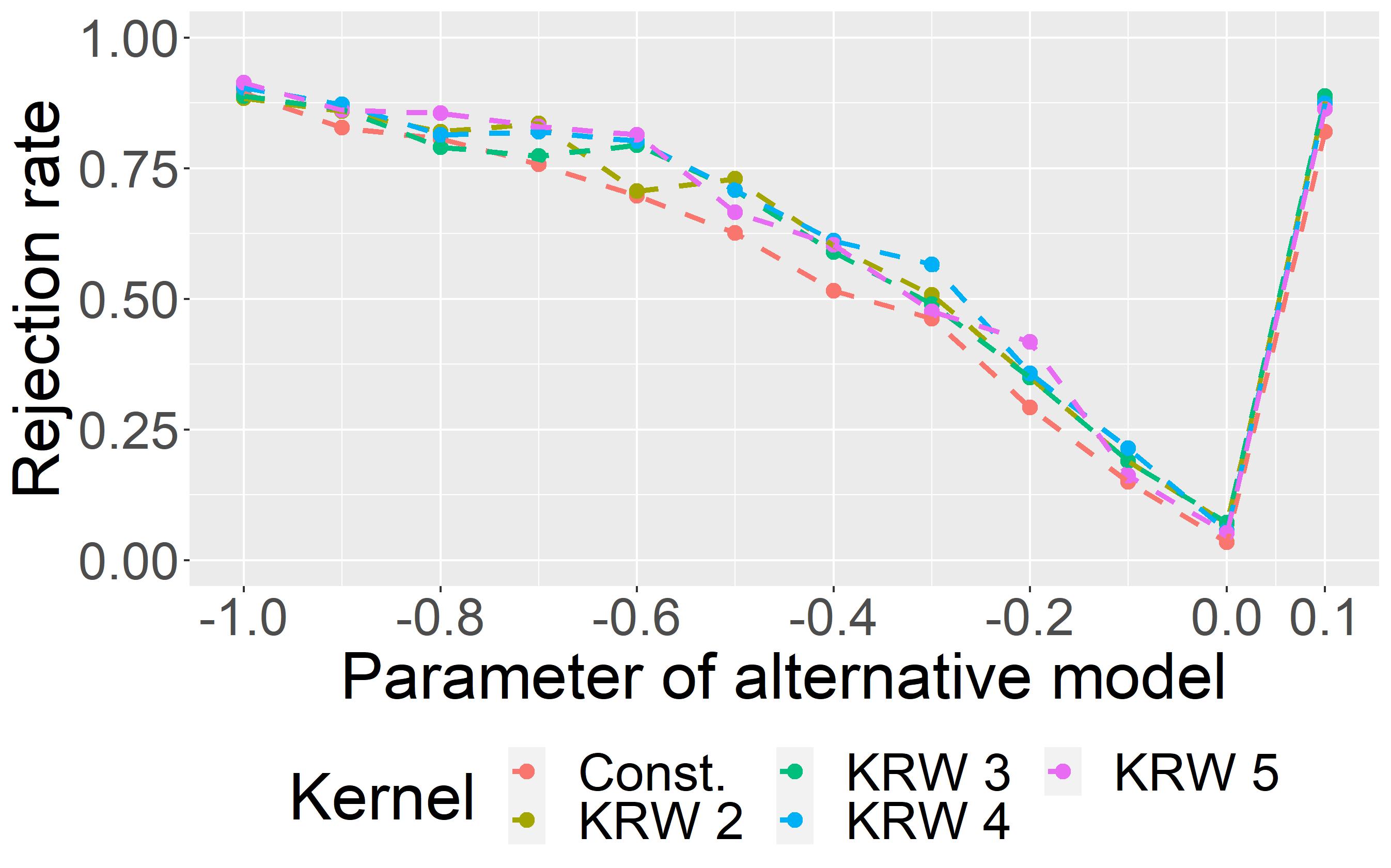}
    \caption{
    Kernel experiments in {the} setting of \Cref{fig:e2s}: gKSS for E2S {model}  with $\beta_2$ perturbed.
    }
    \label{fig:e2s-morekernel}
\end{figure*}

\begin{figure*}[t!]
    \centering
    {\includegraphics[width=0.33\textwidth]{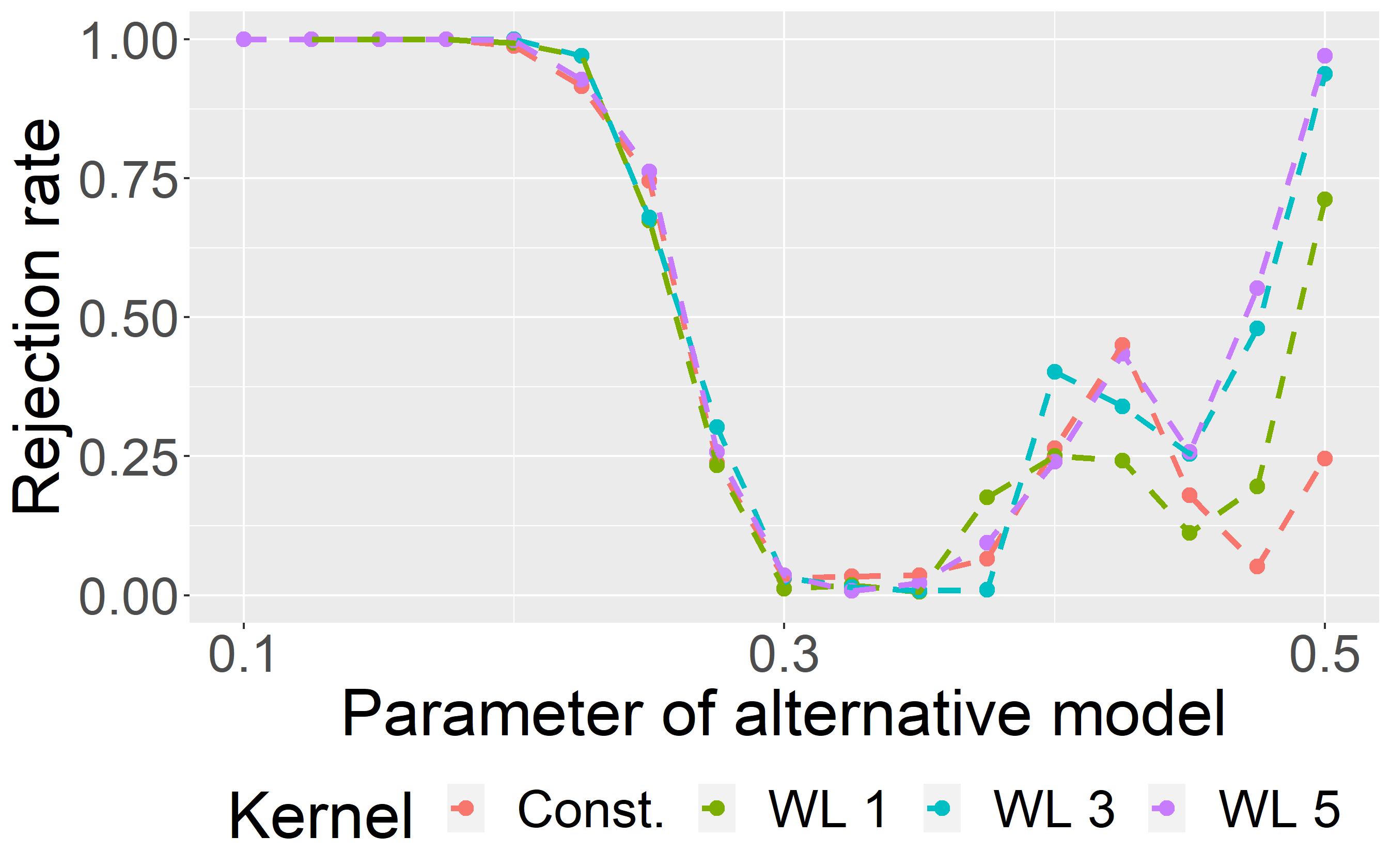}}\includegraphics[width=0.332\textwidth]{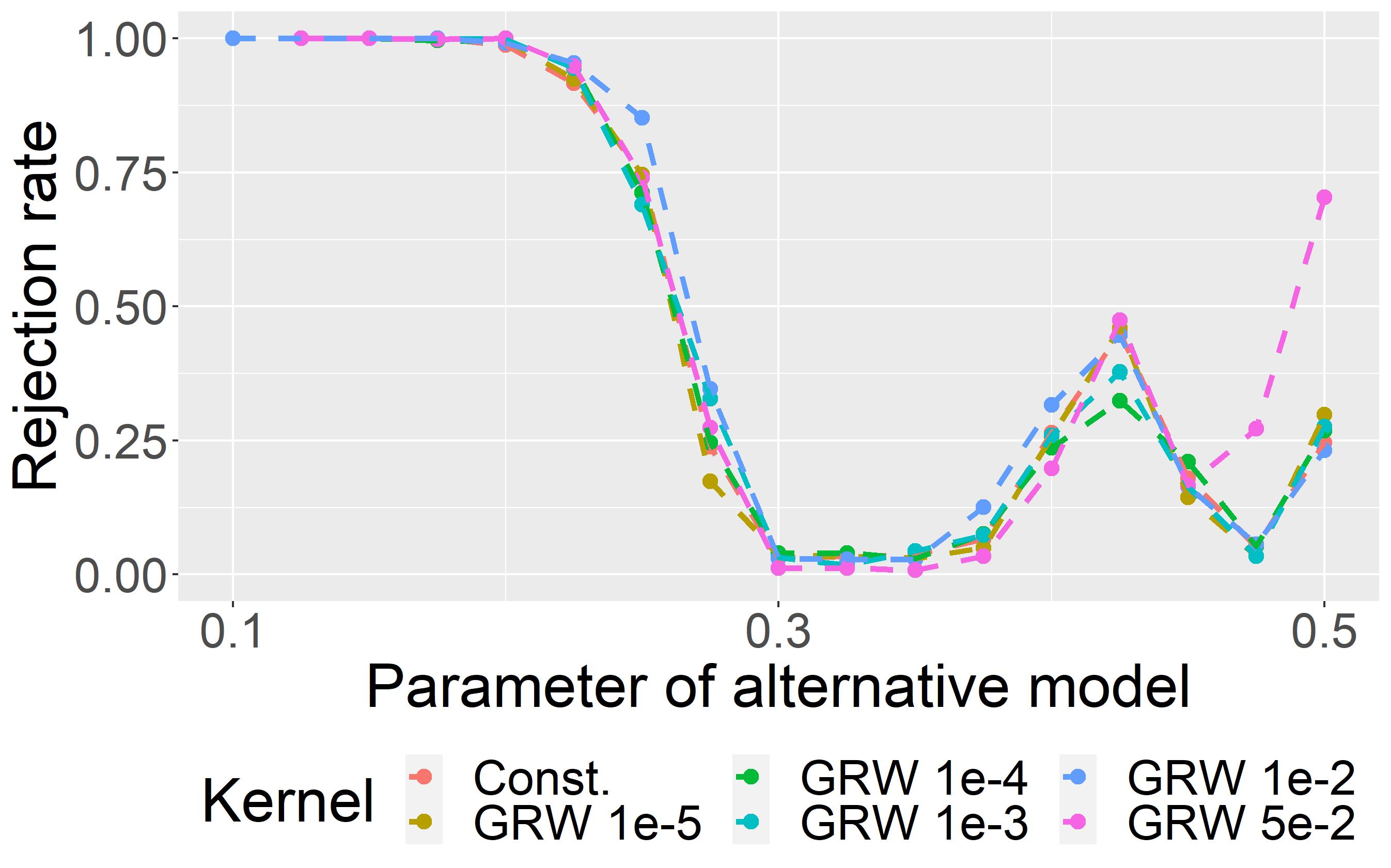}
    {\includegraphics[width=0.33\textwidth]{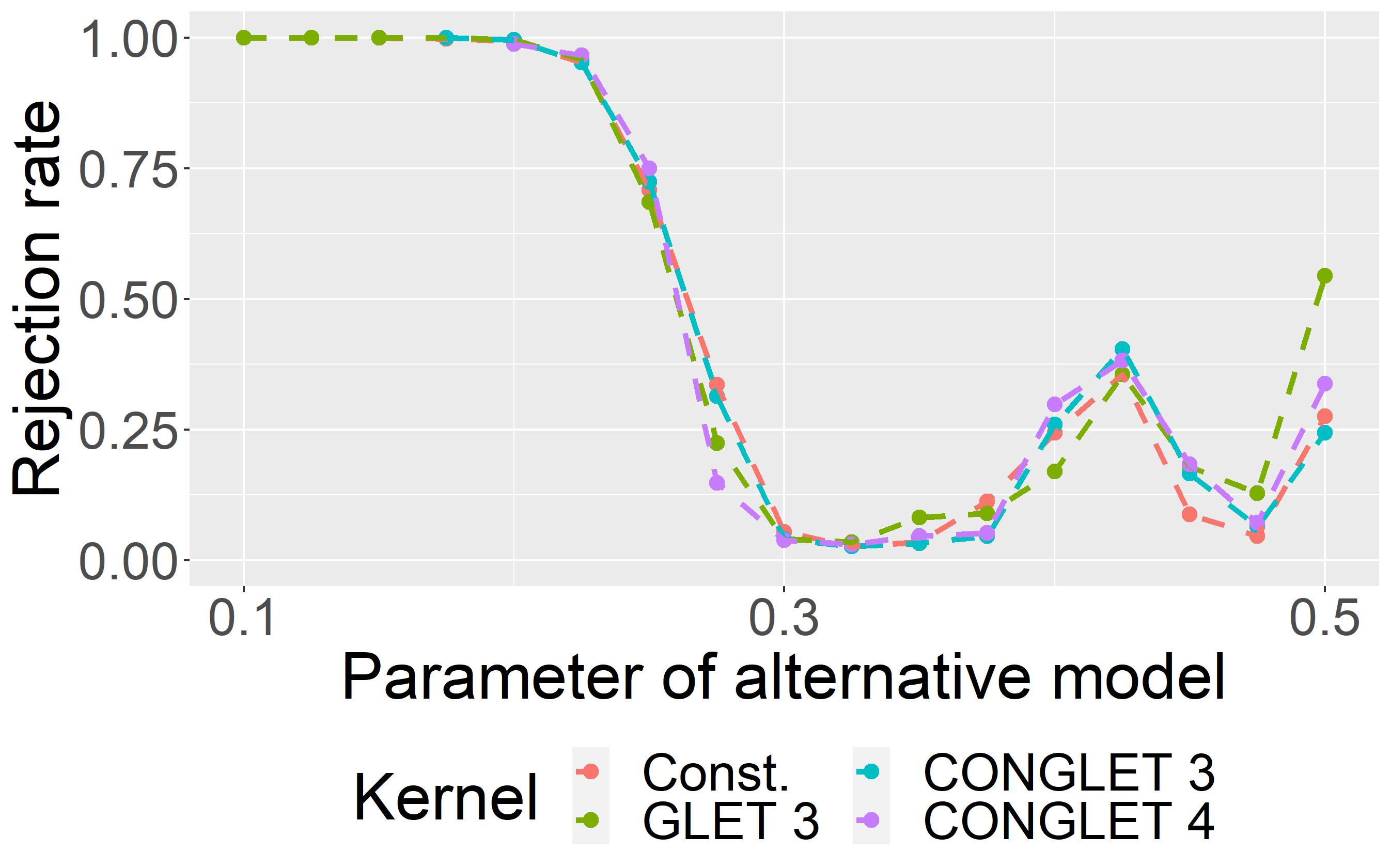}}
    \caption{
        Kernel experiments in the  setting of \Cref{fig:grg}: AgraSSt for {the} GRG model with  alternative;  
    $t(x)$ is set to be bivariate (vertex) degree vector.
    }
    \label{fig:grg-bideg}
\end{figure*}

\subsubsection{Additional experiments on {the} E2S {model}} In \Cref{fig:e2s-morekernel}, we show the rejection rate for SP, GVEH and KRW kernels in the same E2S setting as presented in \Cref{fig:e2s}. All these kernels have 
similar performance to the constant kernel, with GVEH kernels {being} more sensitive 
to parameter choice than the KRW kernels. 

\begin{figure}[t!]
    \centering
    {\includegraphics[width=0.32\textwidth]{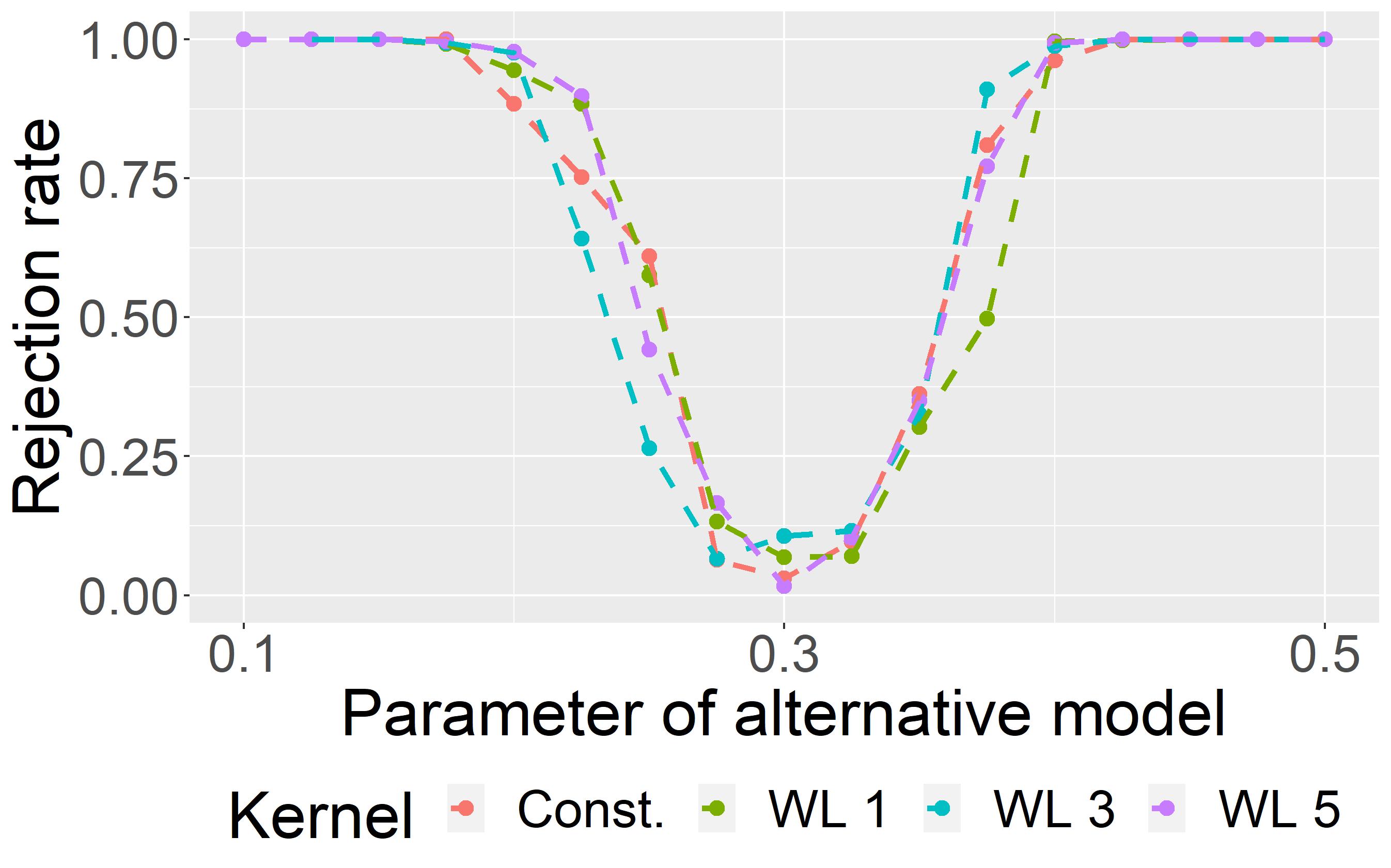}}\includegraphics[width=0.32\textwidth]{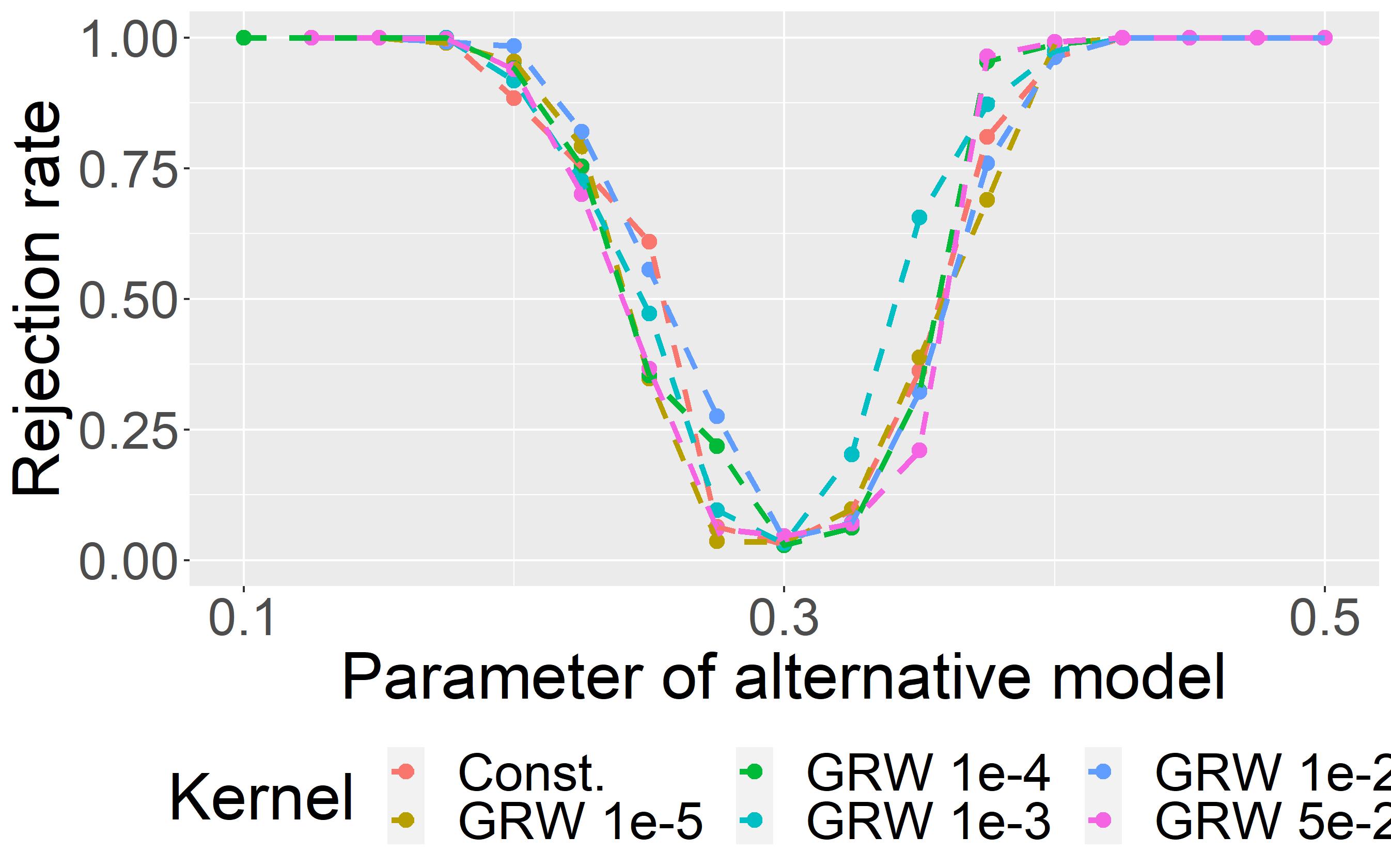}
    {\includegraphics[width=0.32\textwidth]{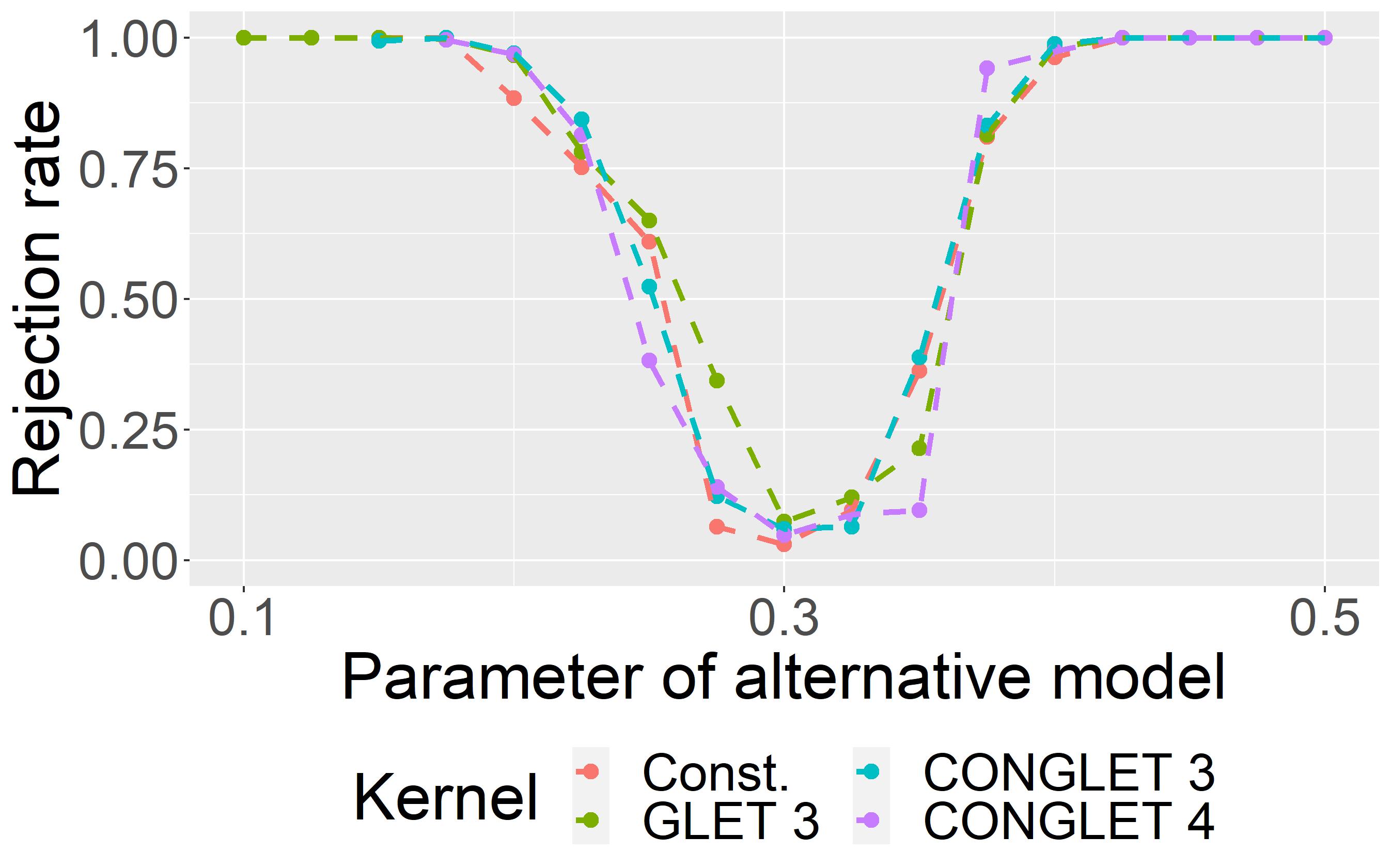}}
        {\includegraphics[width=0.32\textwidth]{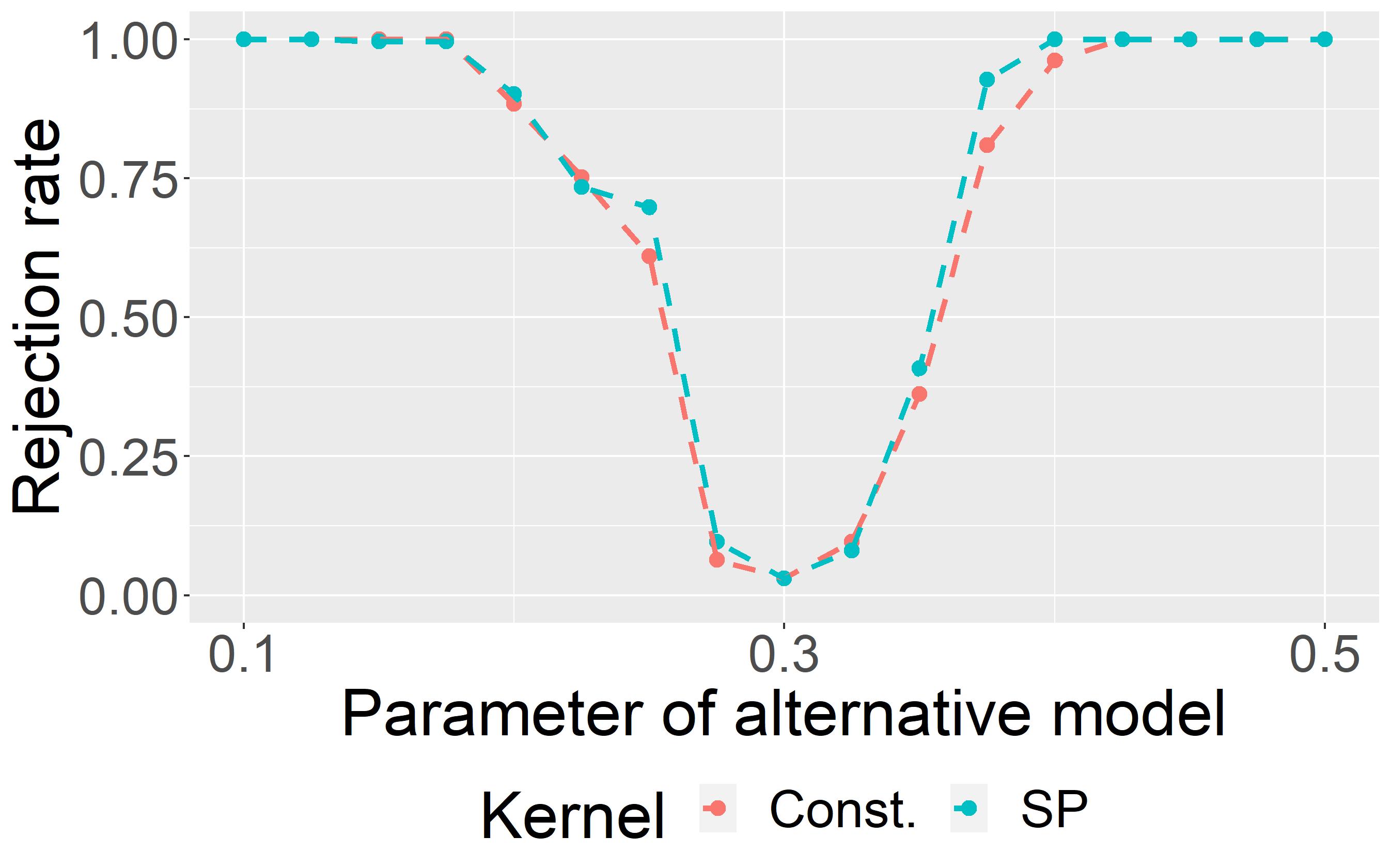}}
\includegraphics[width=0.325\textwidth]{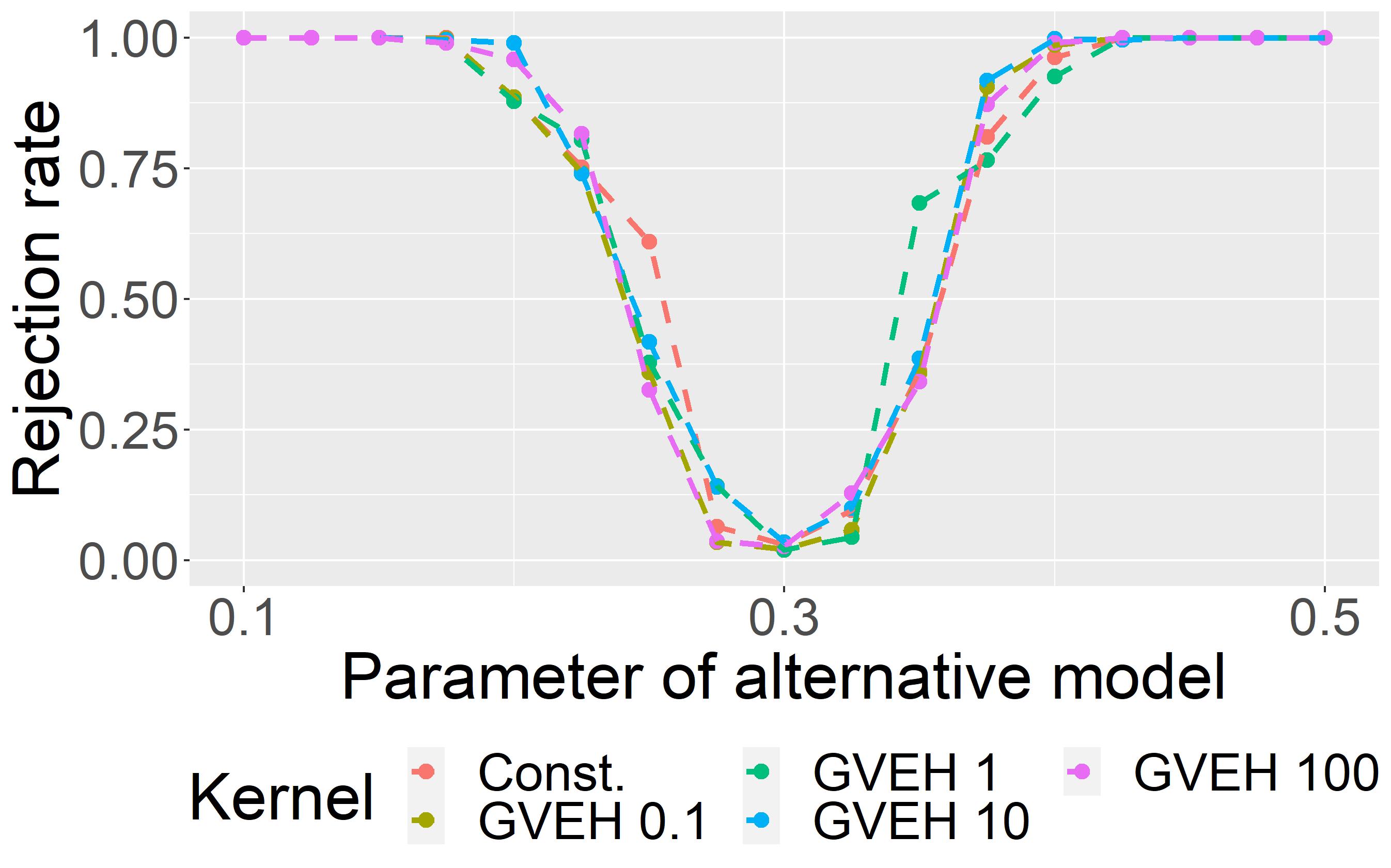}
    {\includegraphics[width=0.32\textwidth]{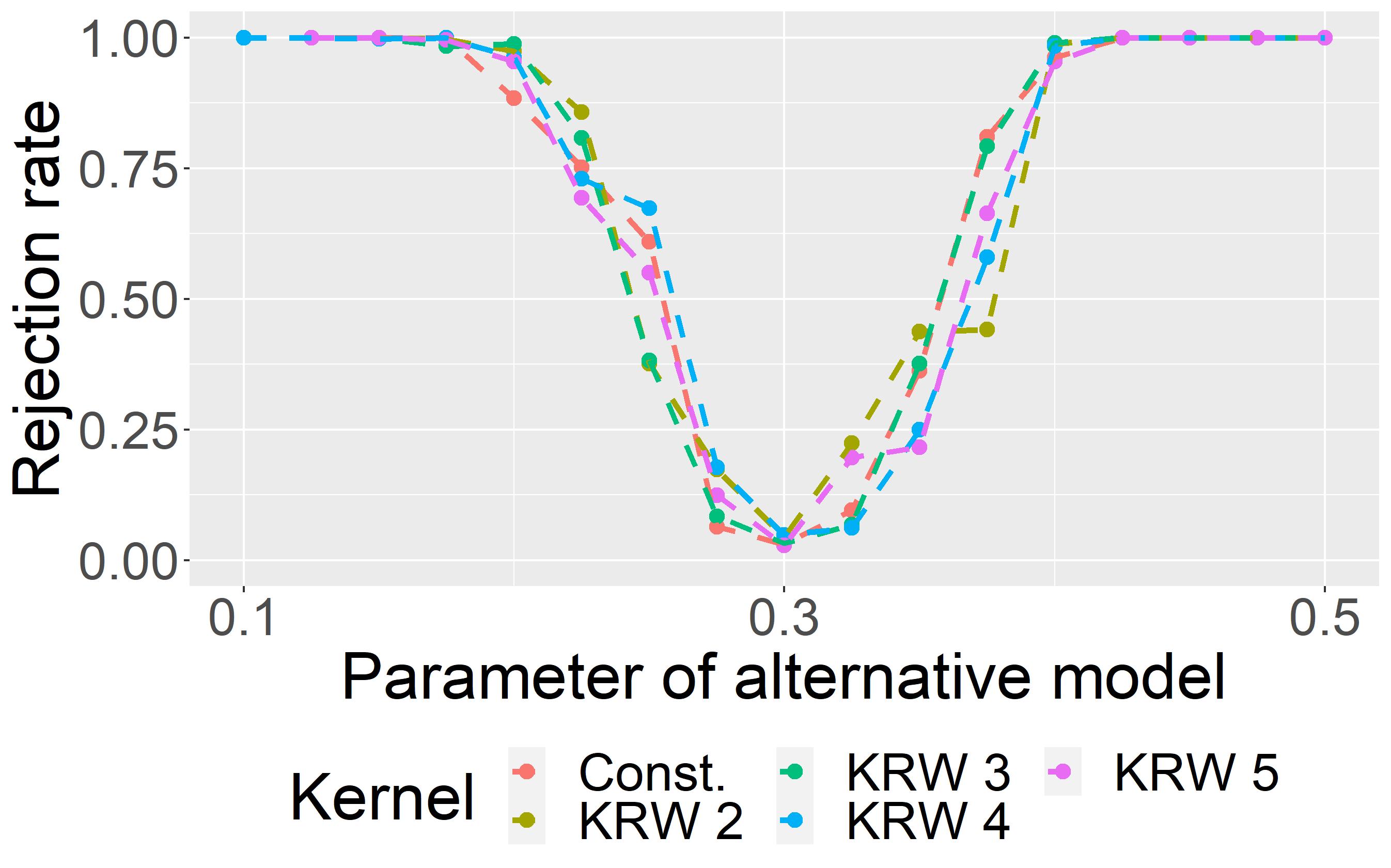}}
    \caption{
AgraSSt for {the} GRG {model} with $t(x)$ being the edge density.
    }
    \label{fig:grg-density}
\end{figure}

\begin{figure*}[t!]
    \centering
    {\includegraphics[width=0.32\textwidth]{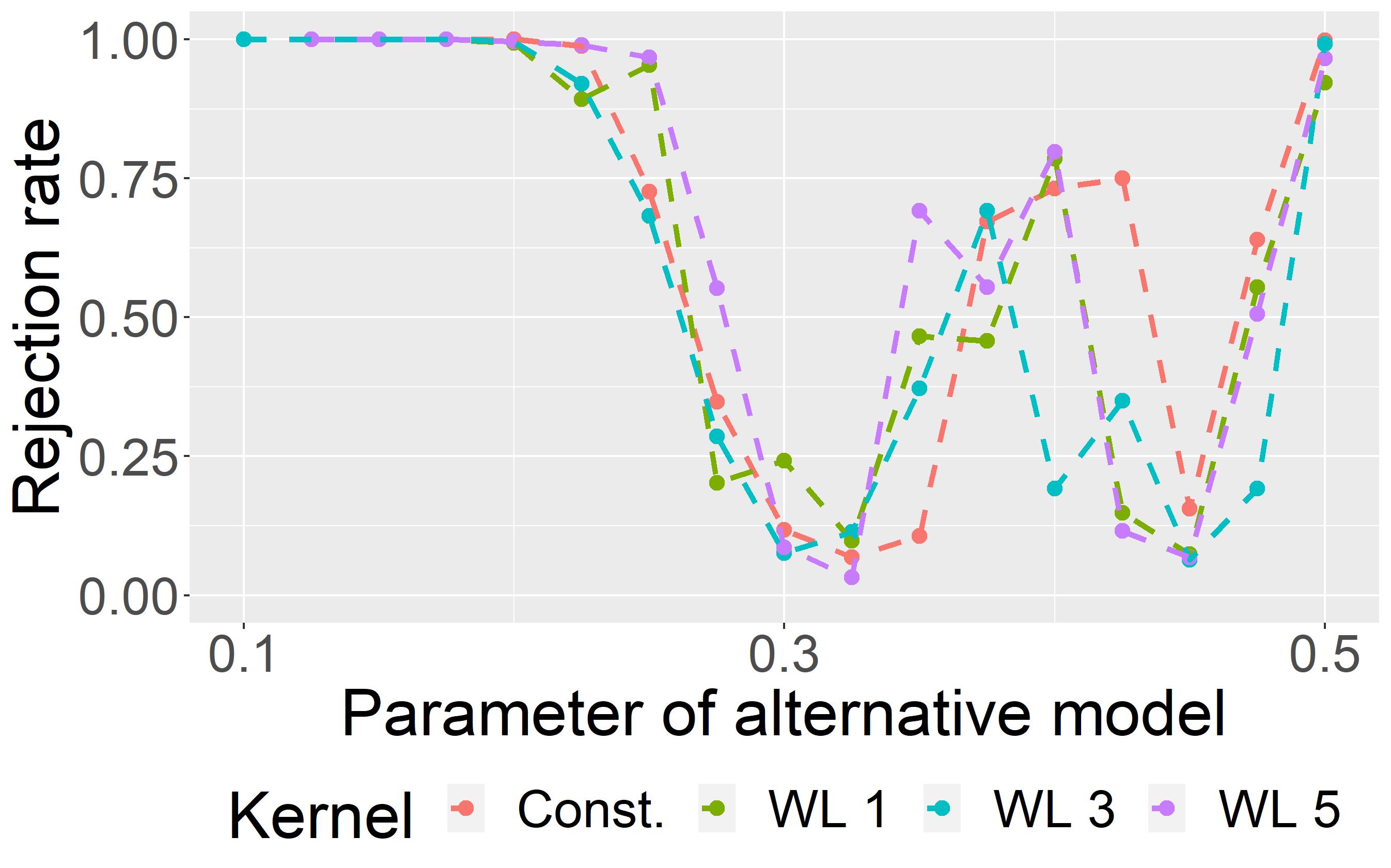}}\includegraphics[width=0.325\textwidth]{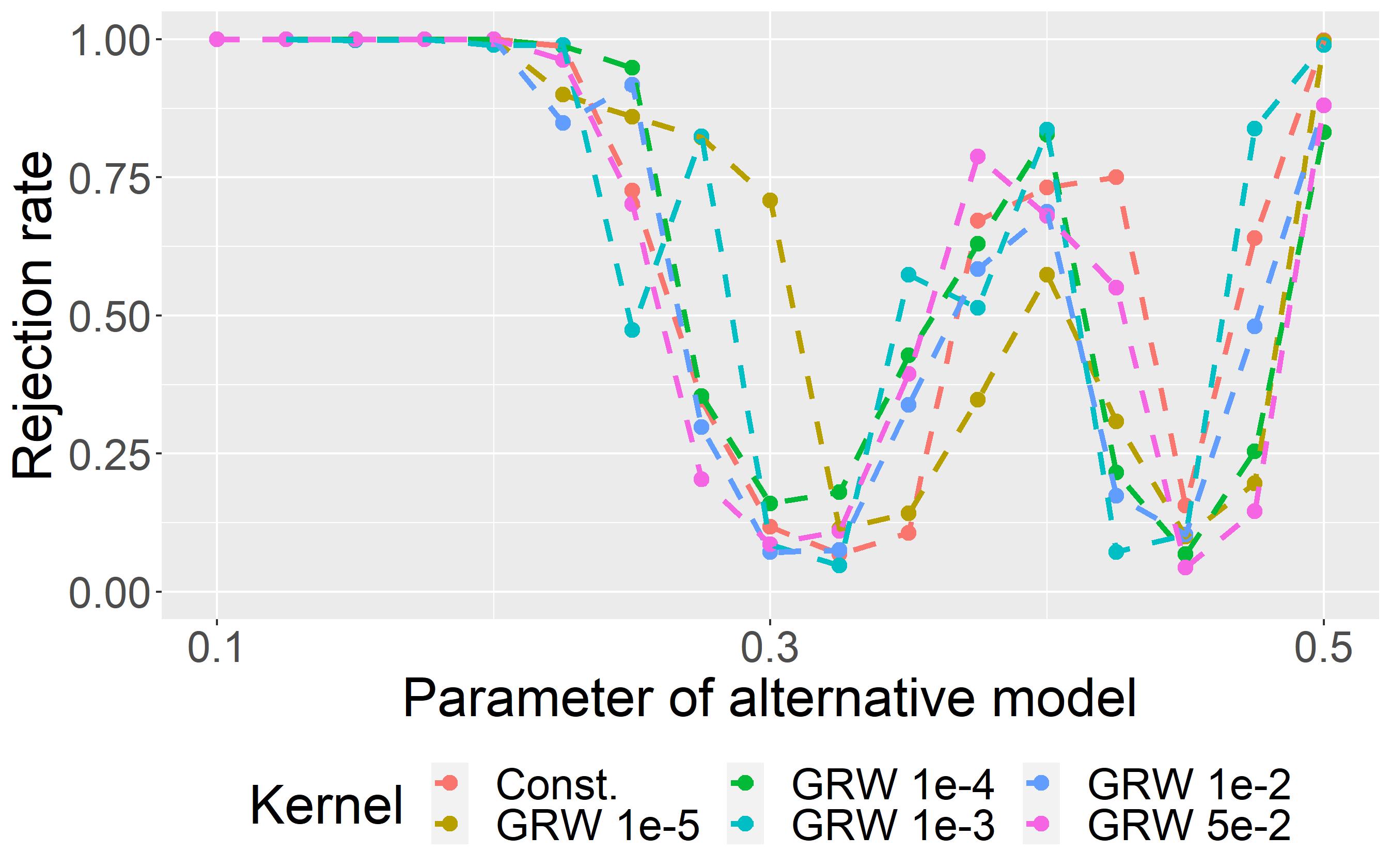}
    {\includegraphics[width=0.33\textwidth]{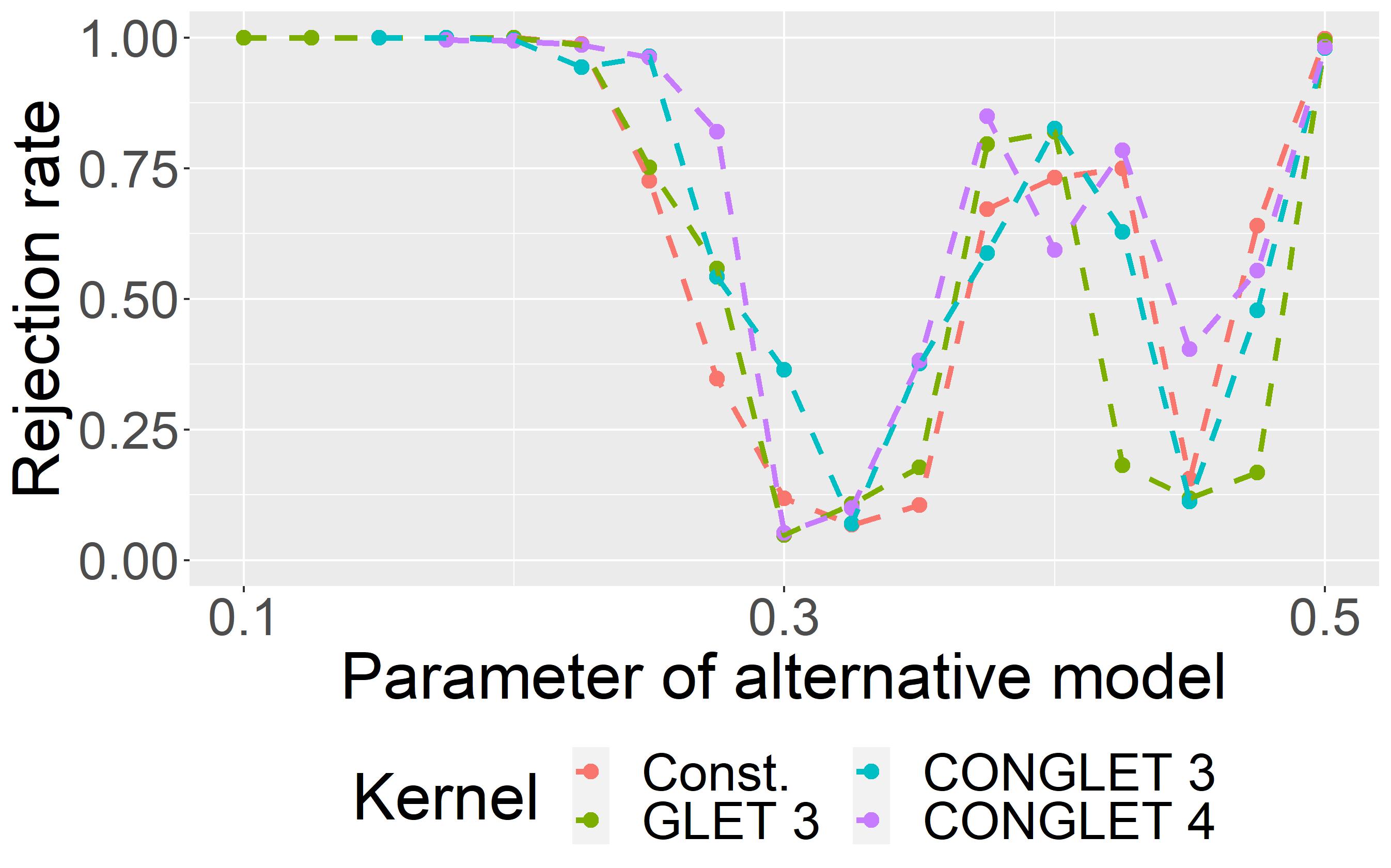}}
        {\includegraphics[width=0.32\textwidth]{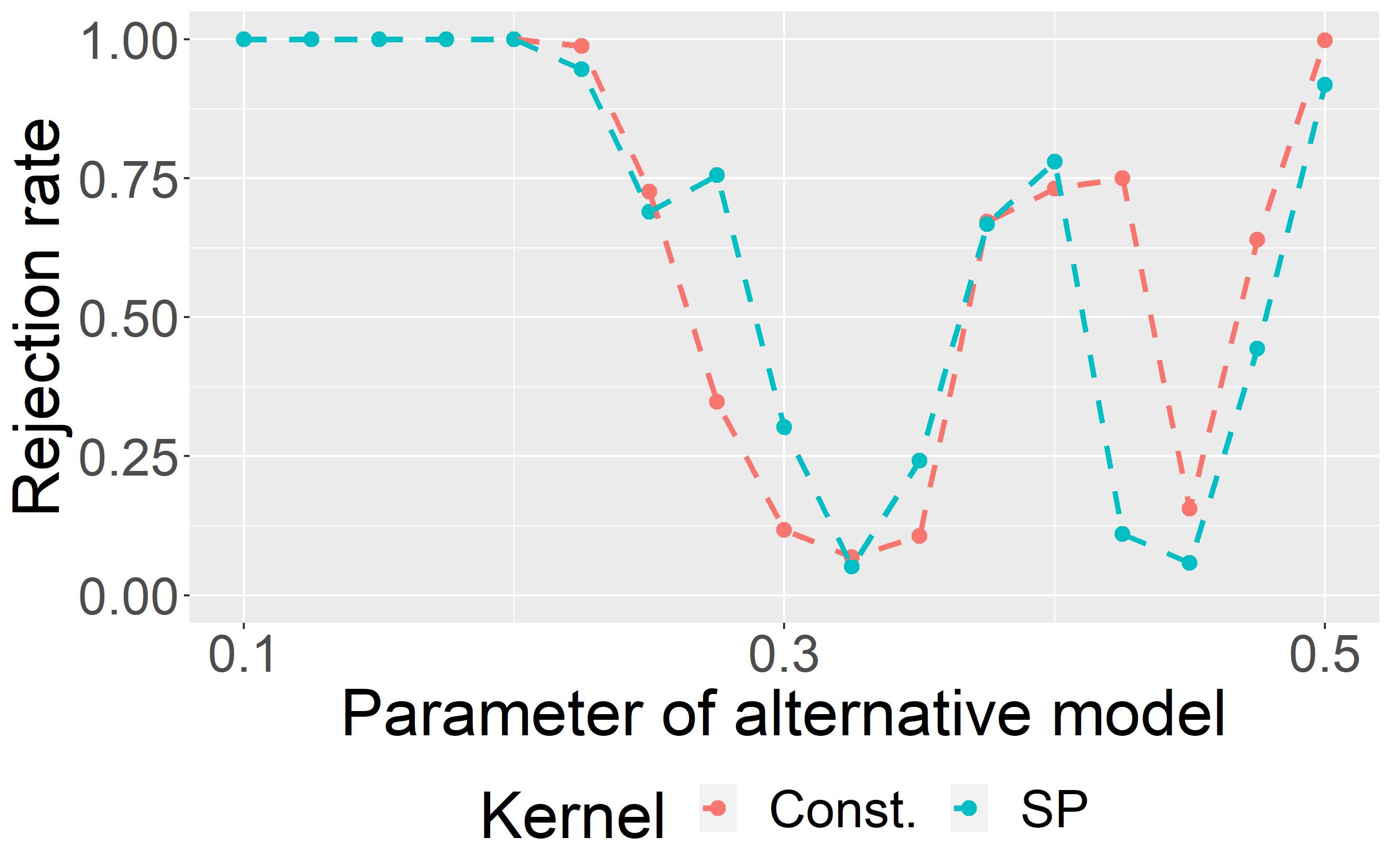}}
\includegraphics[width=0.325\textwidth]{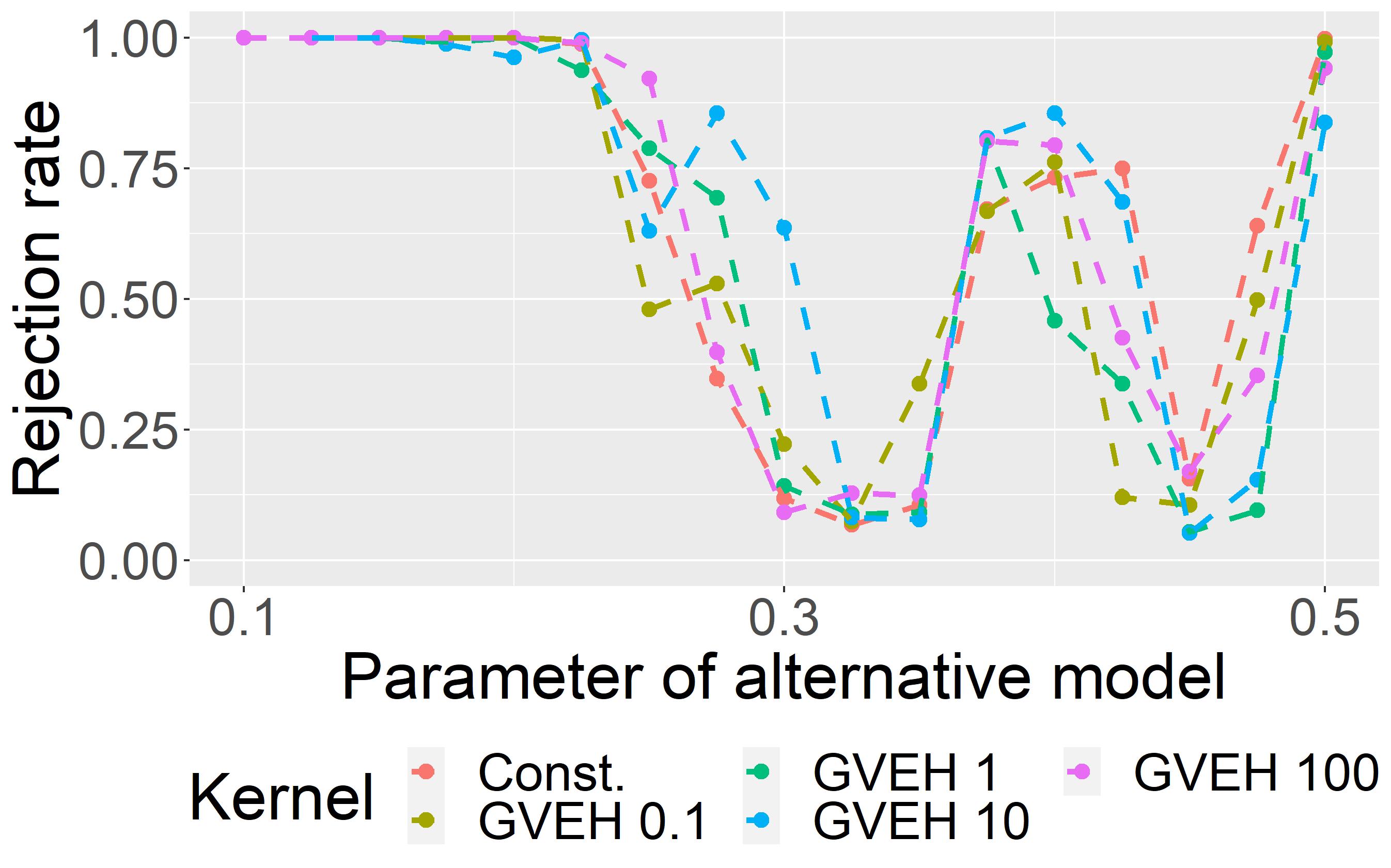}
    {\includegraphics[width=0.33\textwidth]{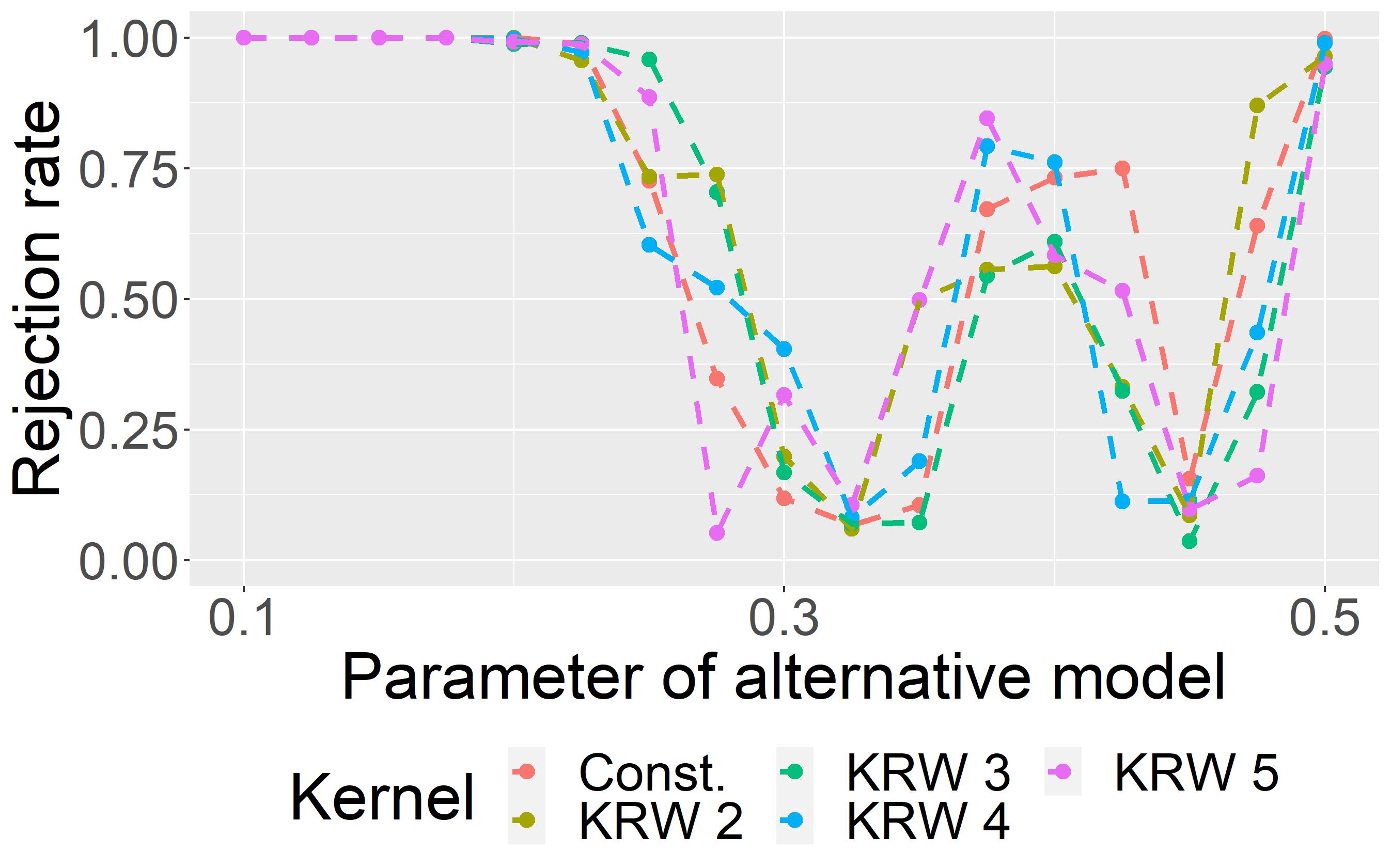}}
    \caption{
AgraSSt for {the} GRG {model} with $t(x)$ being the number of common neighbours.
    }
    \label{fig:grg-comnb}
\end{figure*}

{
\subsubsection{Additional GRG experiments: {GRG {models} on {the torus}}}

In the main text the results of the experiments using as $t$ the bivariate vertex degree vector are shown. In \Cref{fig:grg-density} we show results for using as $t$ the average density in the sample,. The type 1 error is controlled under all kernels; the kernels perform similarly. 

\Cref{fig:grg-comnb} shows the behaviour of the kernels using the common neighbour statistic. The behaviour is similar to the bi-degree statistic, in showing an additional
dip. The constant kernel and the shortest path kernel have lowest rejection rate not at the true value. {The connected graphlet kernel with graphlet size 3} also suffers from this issue. 
}

\subsubsection{Additional GRG experiments: {GRG {models} on {a} 
unit square}}

\begin{figure*}[htp!]
    \centering
    {\includegraphics[width=0.33\textwidth]{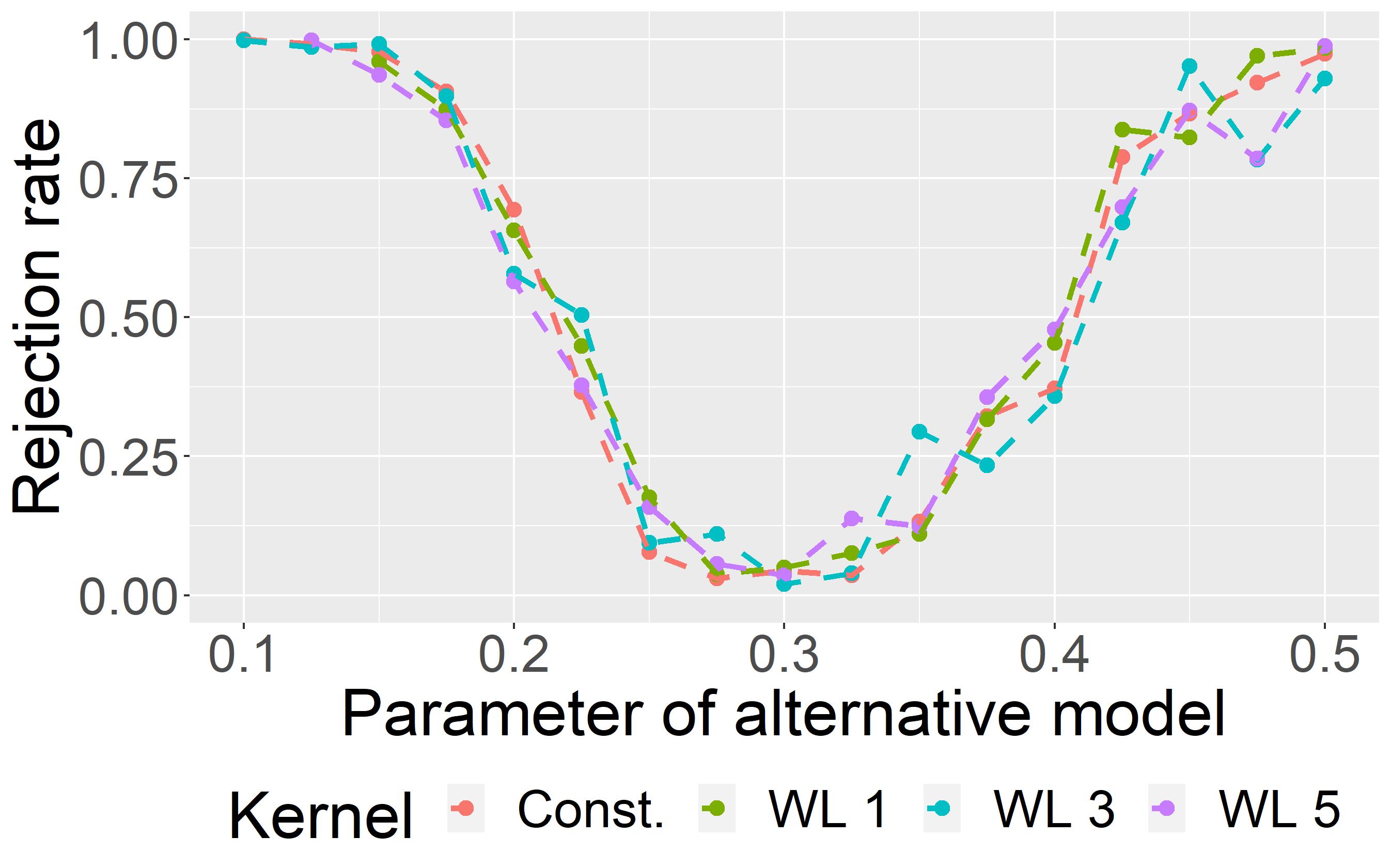}}\includegraphics[width=0.332\textwidth]{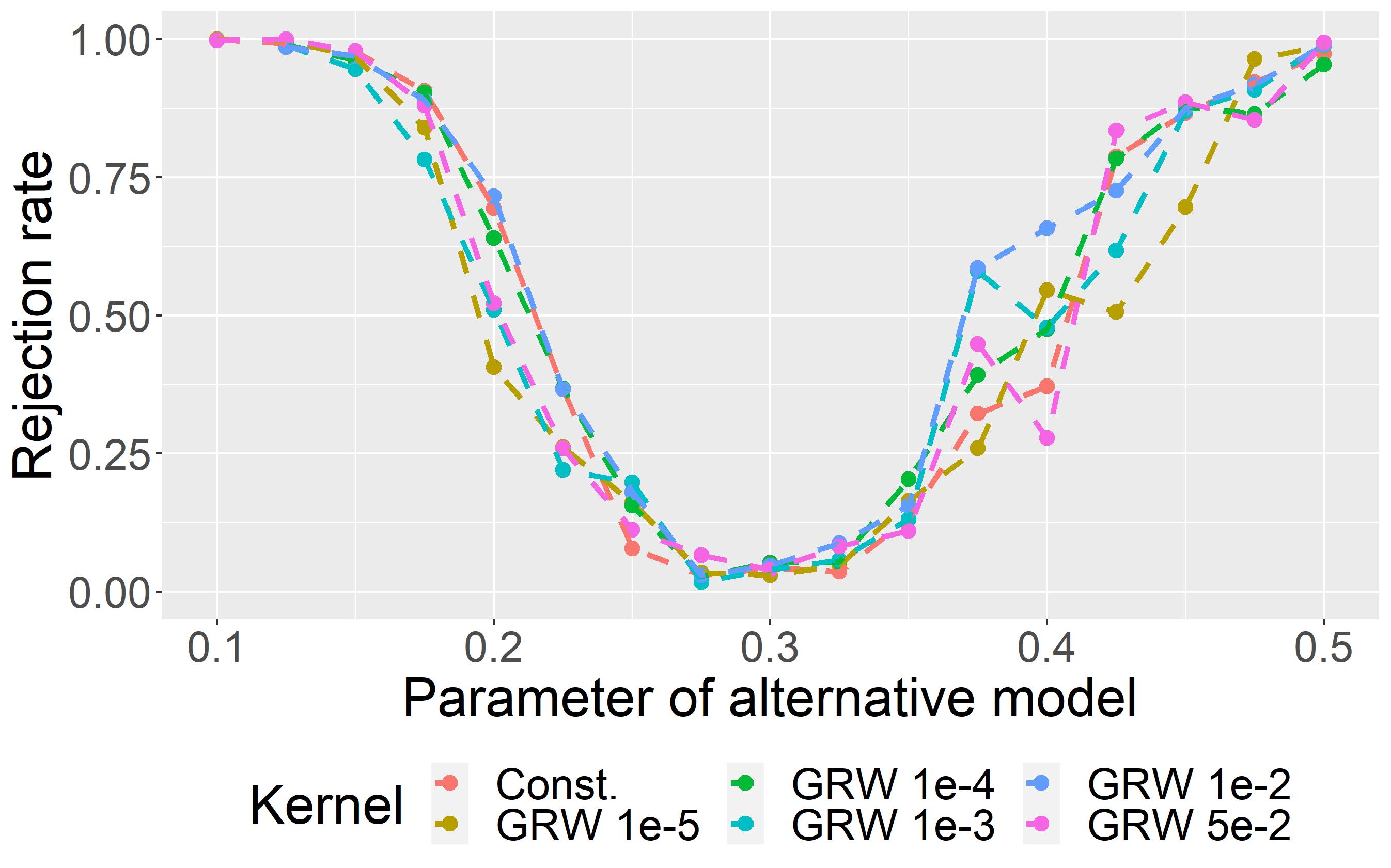}
    {\includegraphics[width=0.33\textwidth]{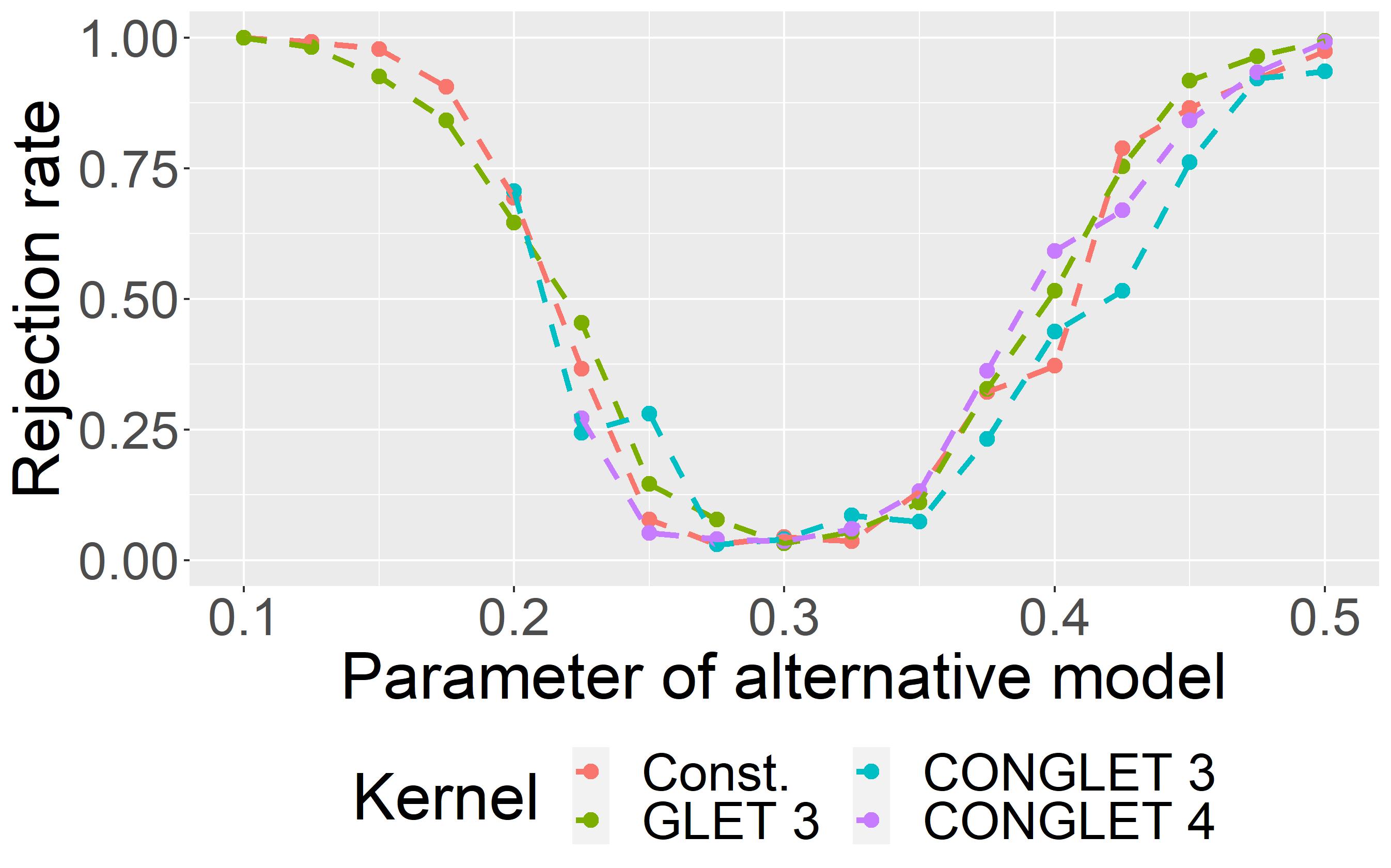}}
        {\includegraphics[width=0.32\textwidth]{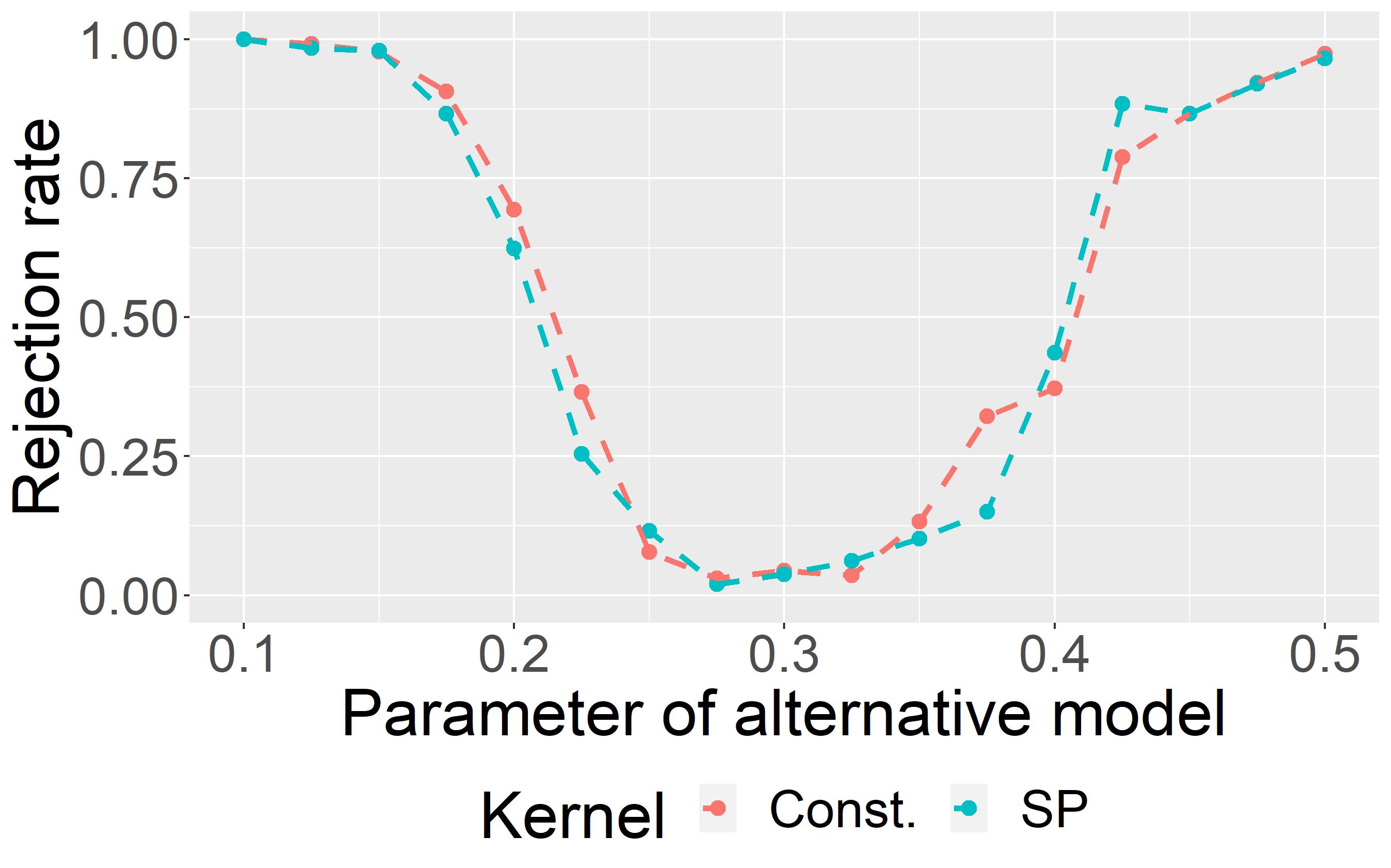}}
\includegraphics[width=0.325\textwidth]{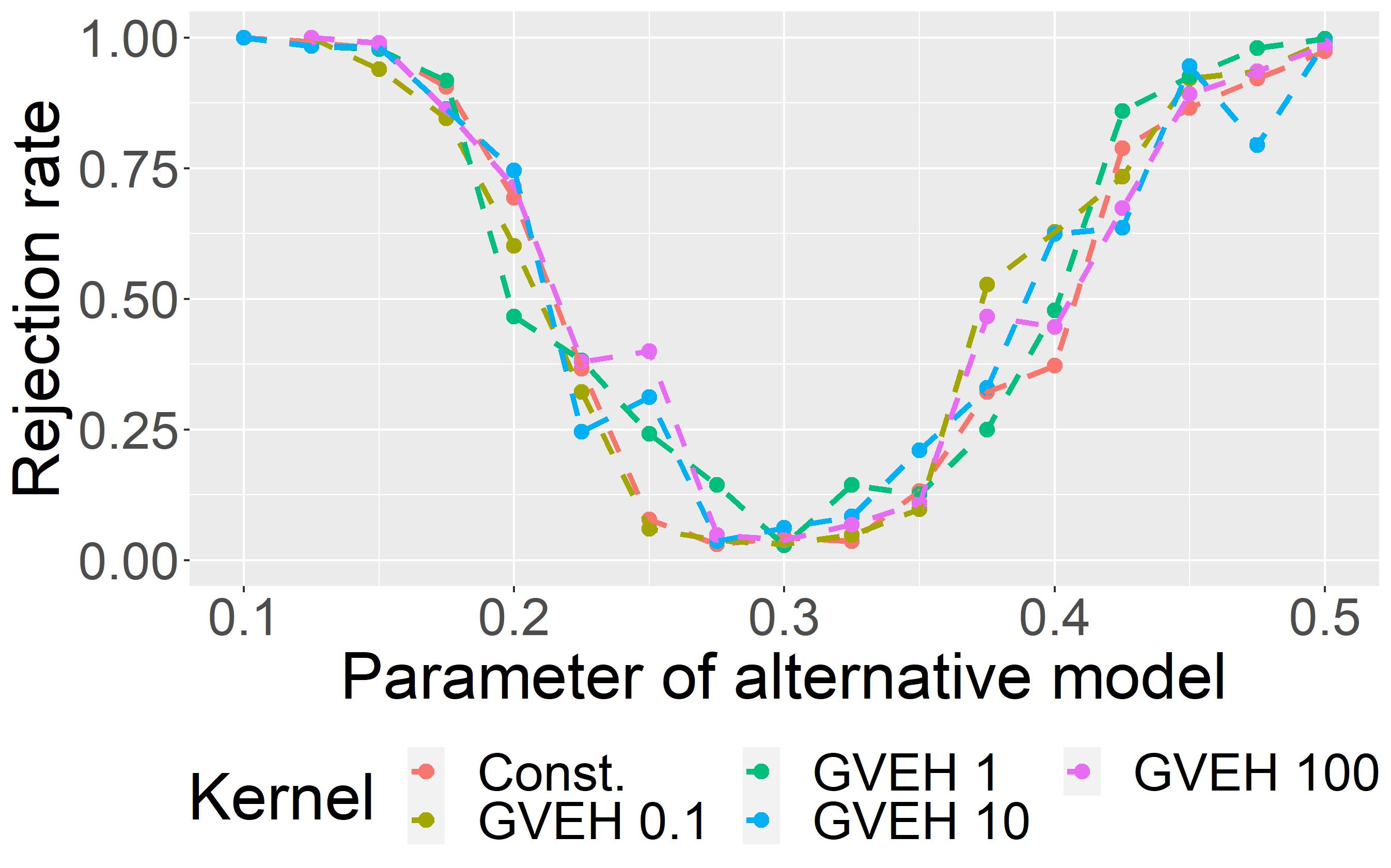}
    {\includegraphics[width=0.32\textwidth]{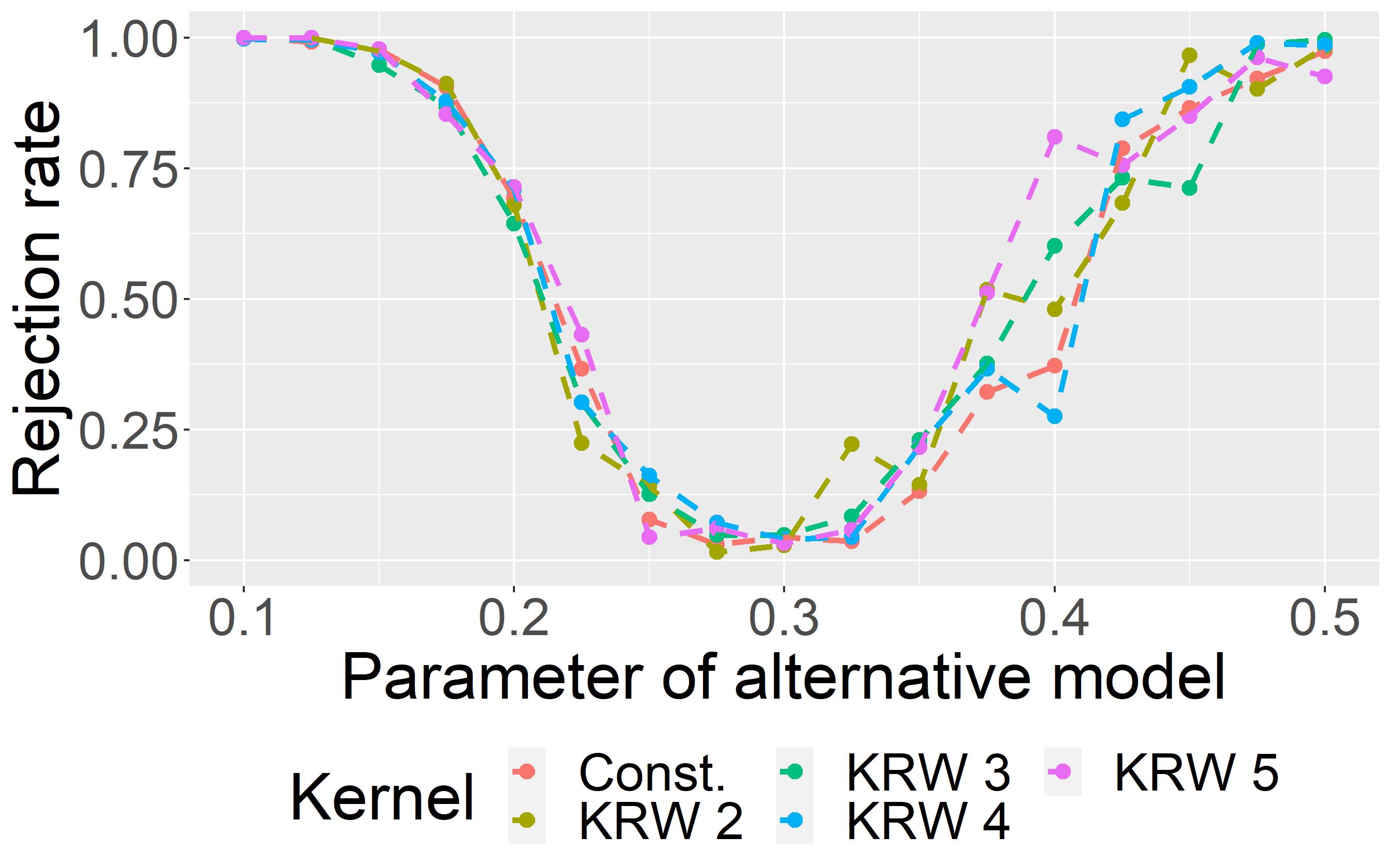}}
    \caption{
AgraSSt for GRG {on the 2-dimensional} 
unit square  with $t(x)$ being the edge density.
}
    \label{fig:grg-nt-density}
\end{figure*}

\begin{figure*}[htp!]
    \centering
    {\includegraphics[width=0.33\textwidth]{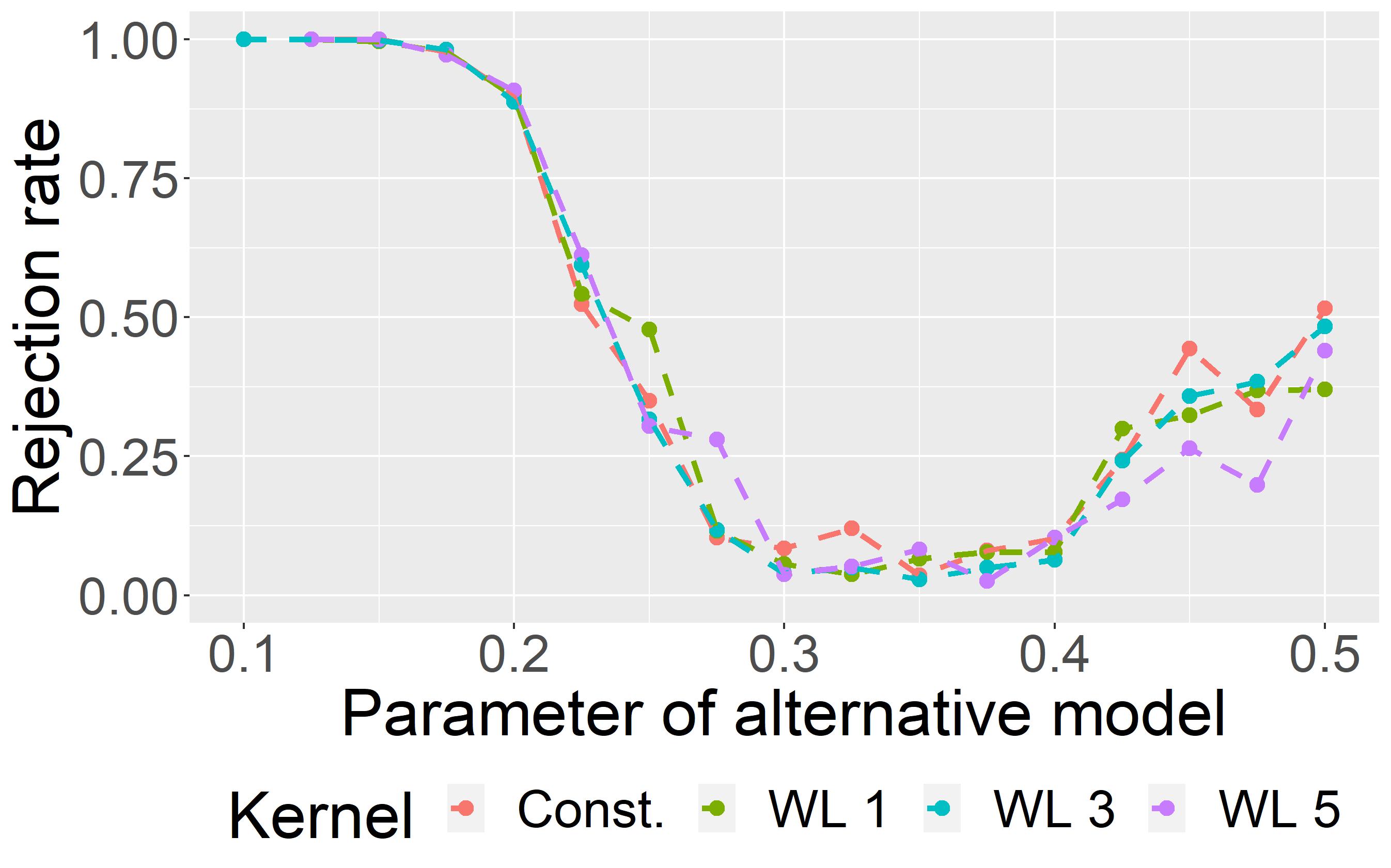}}\includegraphics[width=0.332\textwidth]{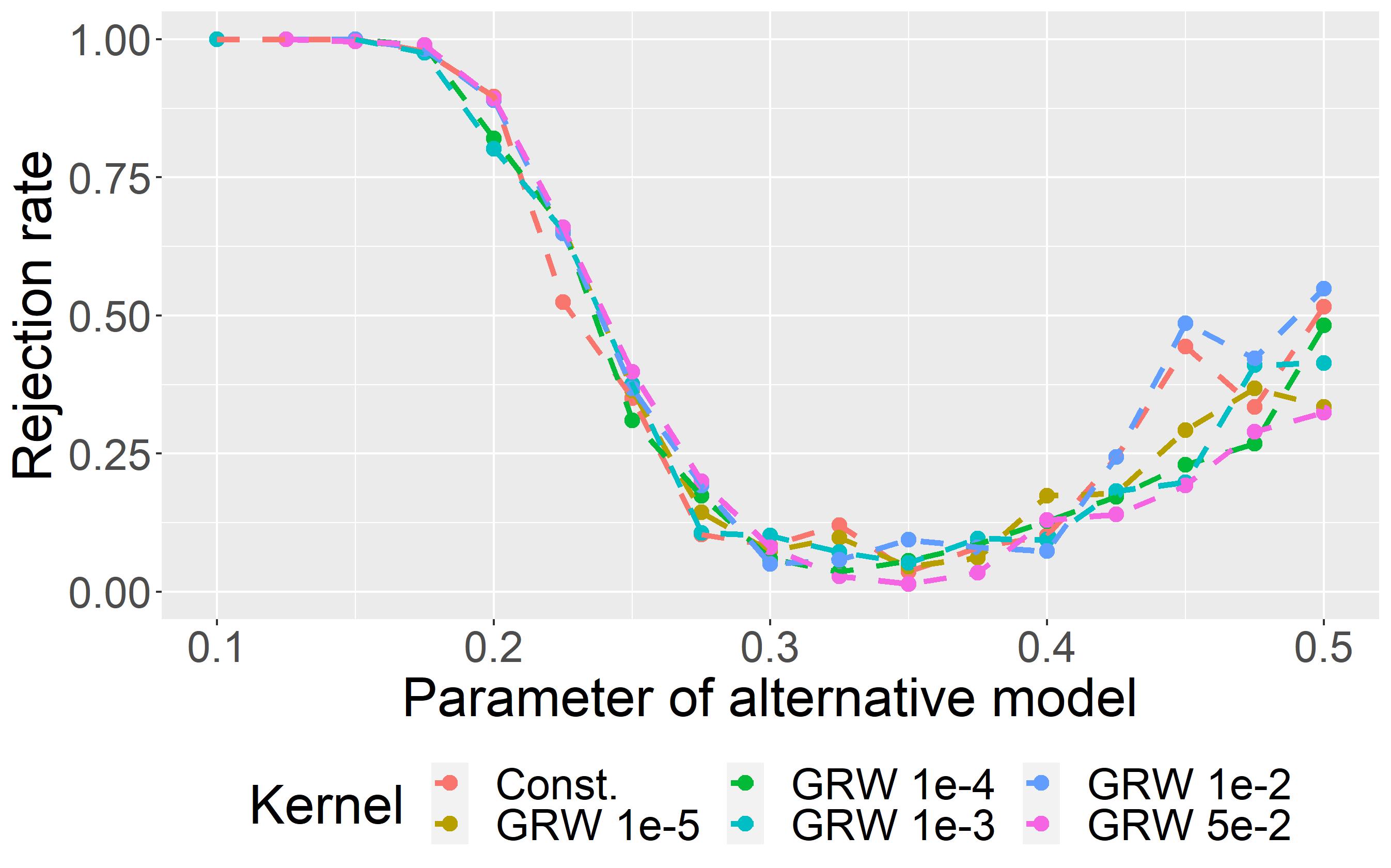}
    {\includegraphics[width=0.33\textwidth]{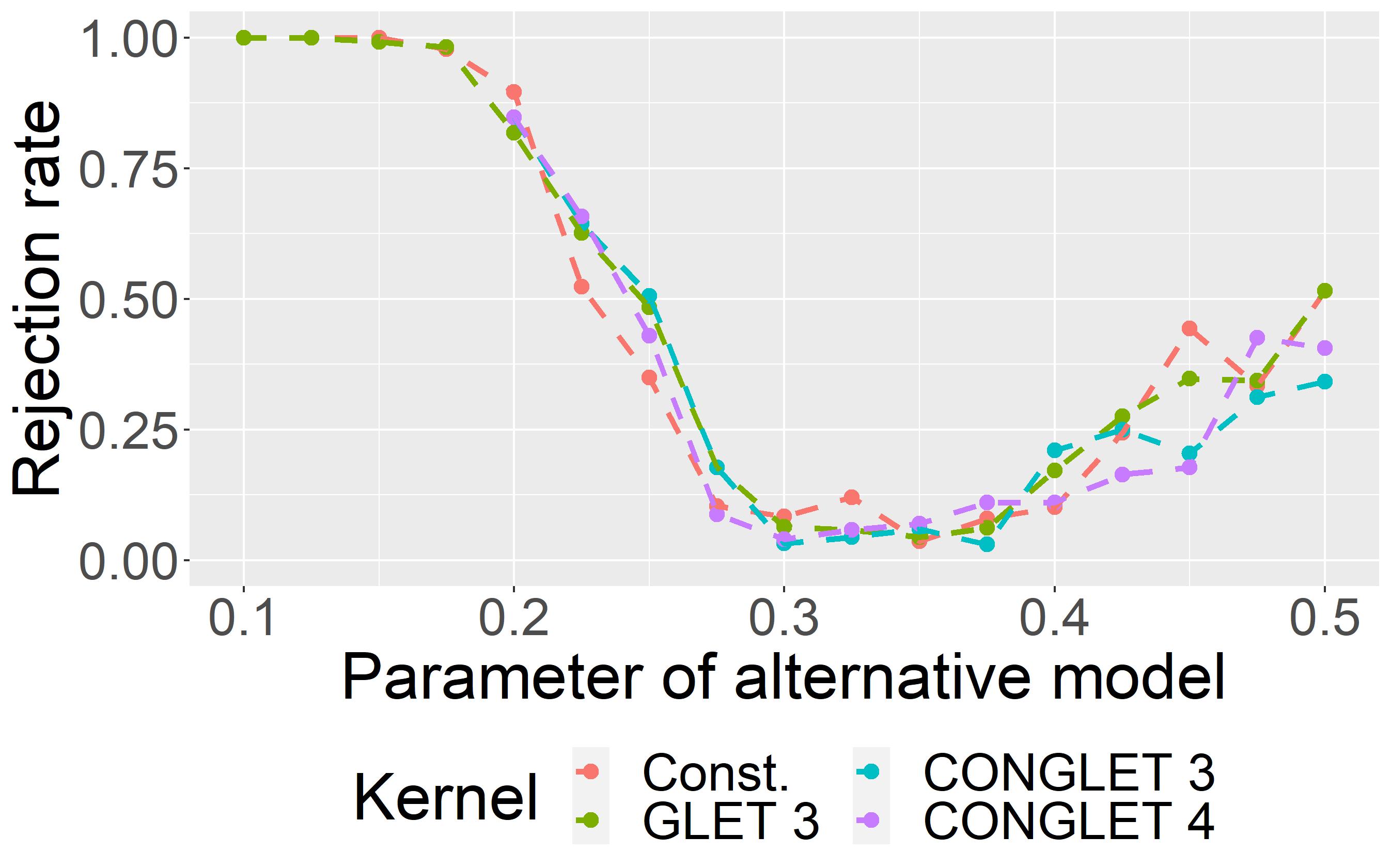}}
        {\includegraphics[width=0.32\textwidth]{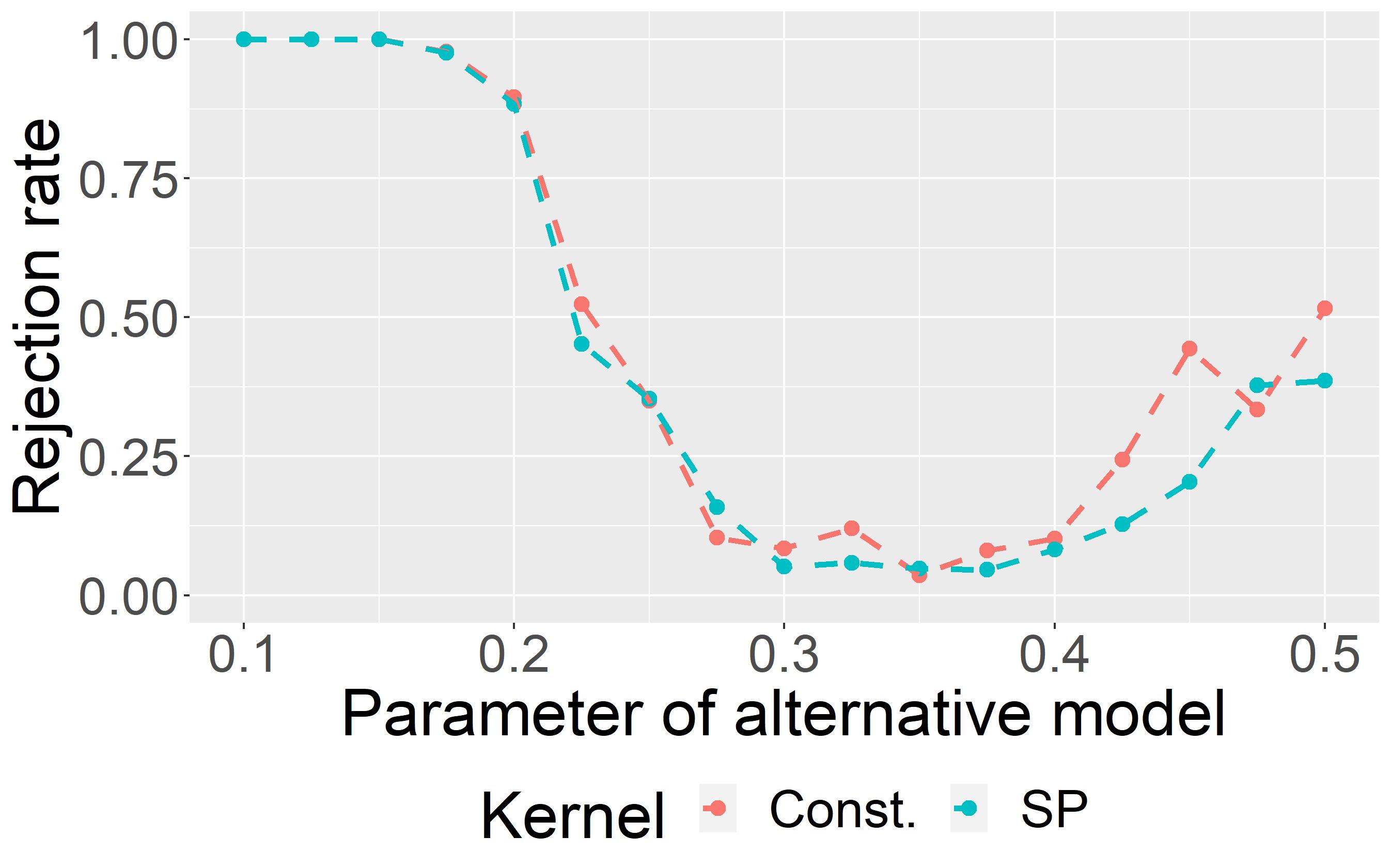}}
\includegraphics[width=0.325\textwidth]{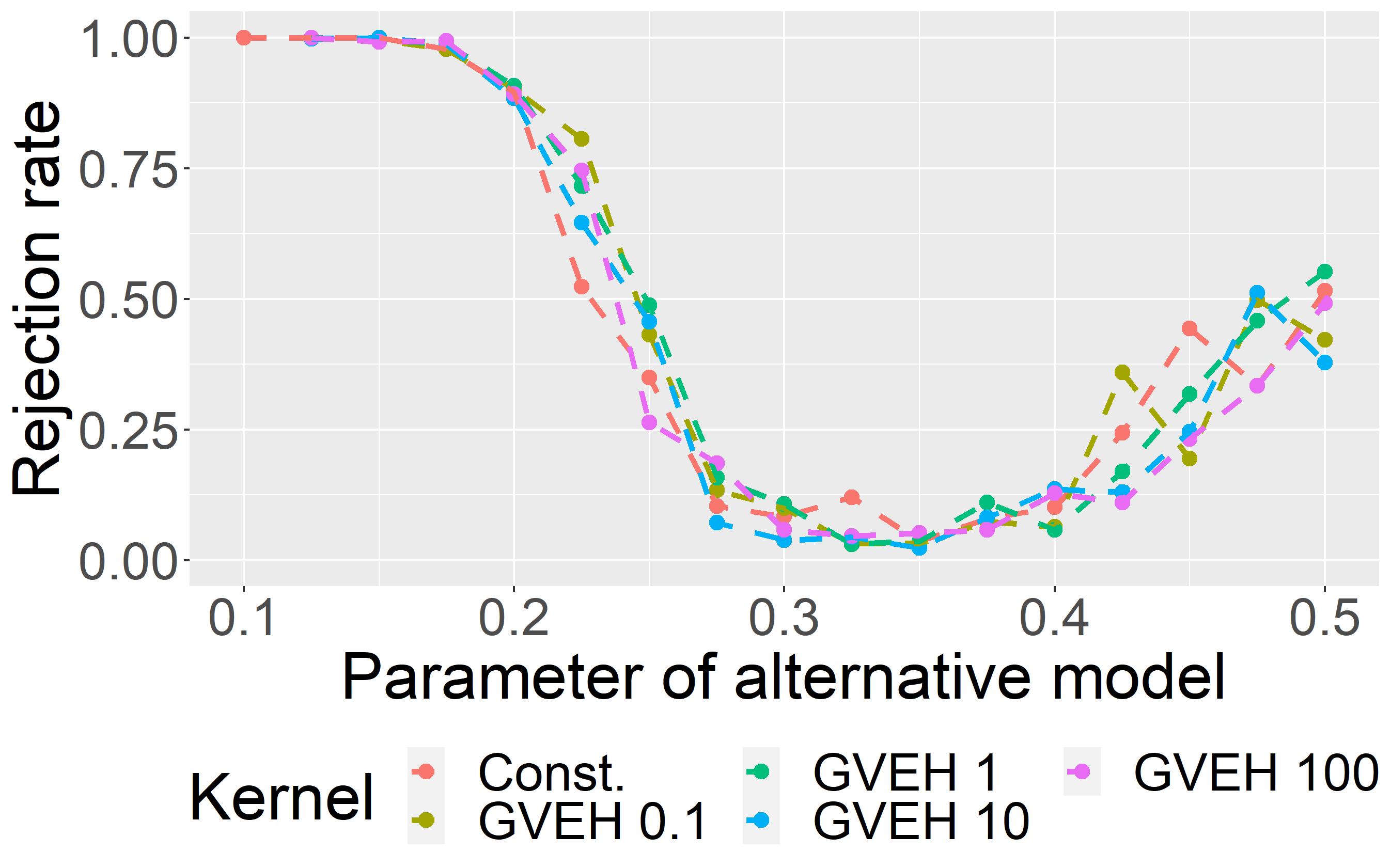}
    {\includegraphics[width=0.32\textwidth]{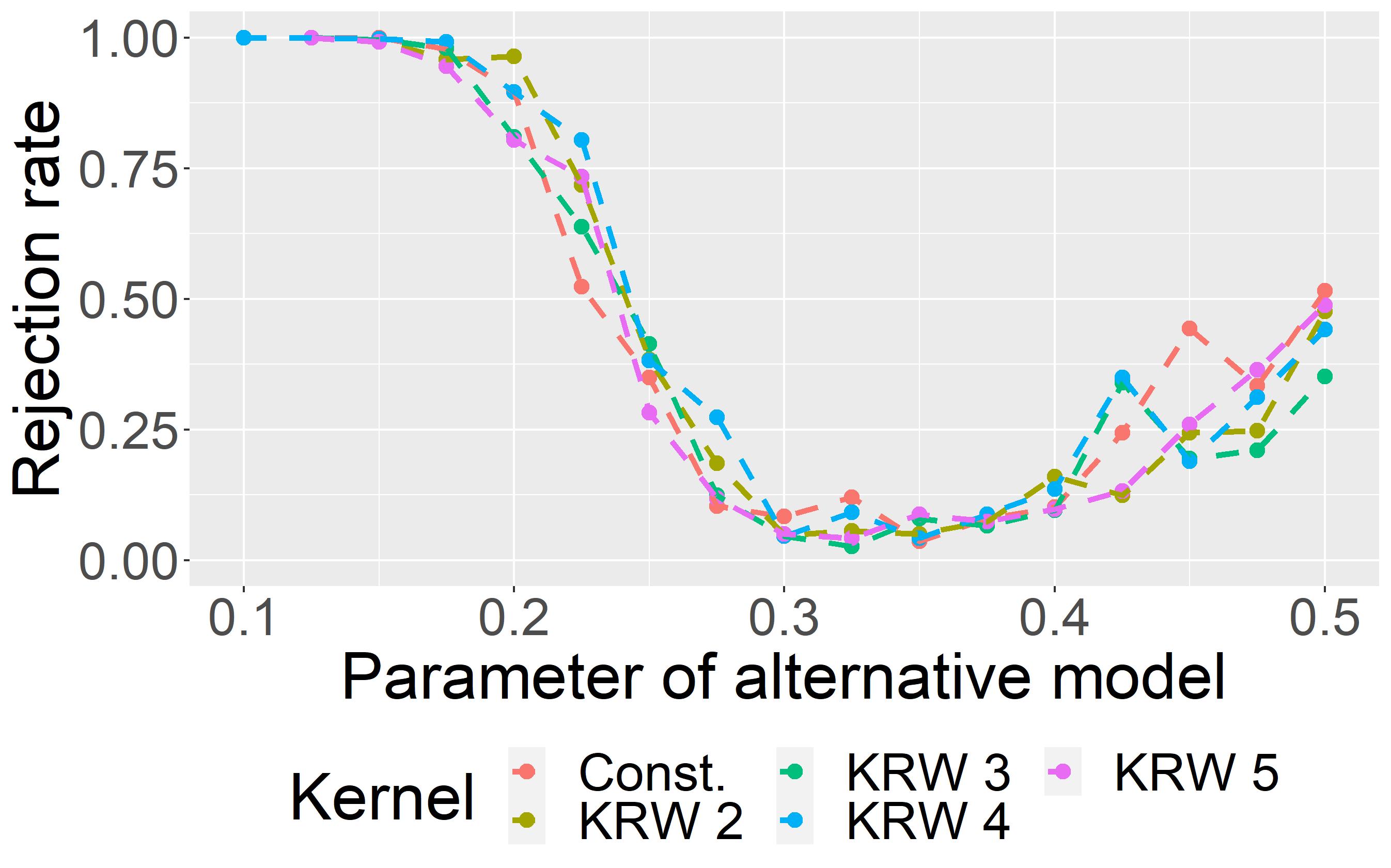}}
    \caption{
AgraSSt for GRG {on the 2-dimensional} 
unit square with $t(x)$ being the bivariate degree vector.
    }
    \label{fig:grg-nt-bideg}
\end{figure*}

\begin{figure*}[htp!]
    \centering
    {\includegraphics[width=0.32\textwidth]{figure/GRG-nt/Dense_WL_n20_B200.jpg}}\includegraphics[width=0.32\textwidth]{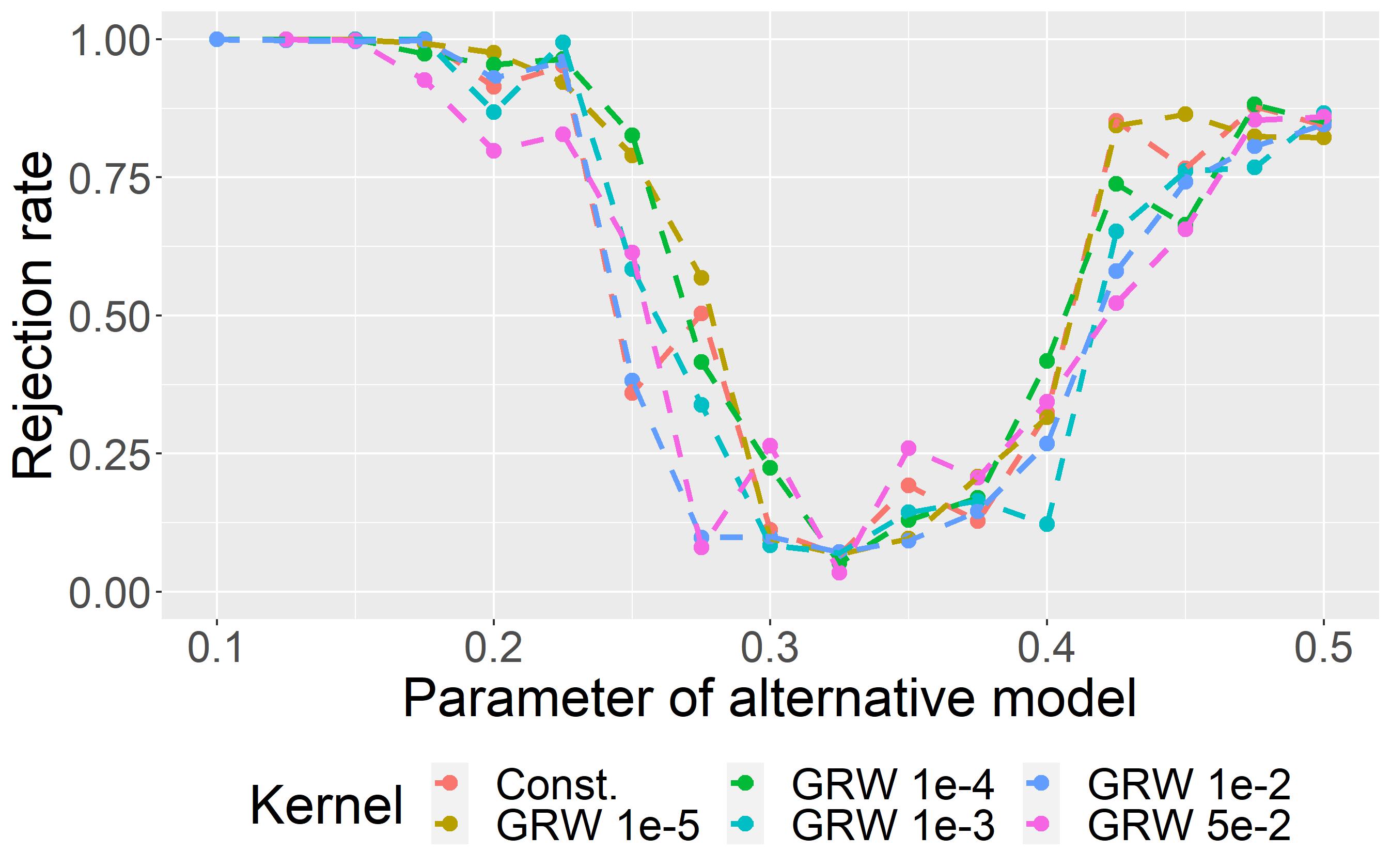}
    {\includegraphics[width=0.32\textwidth]{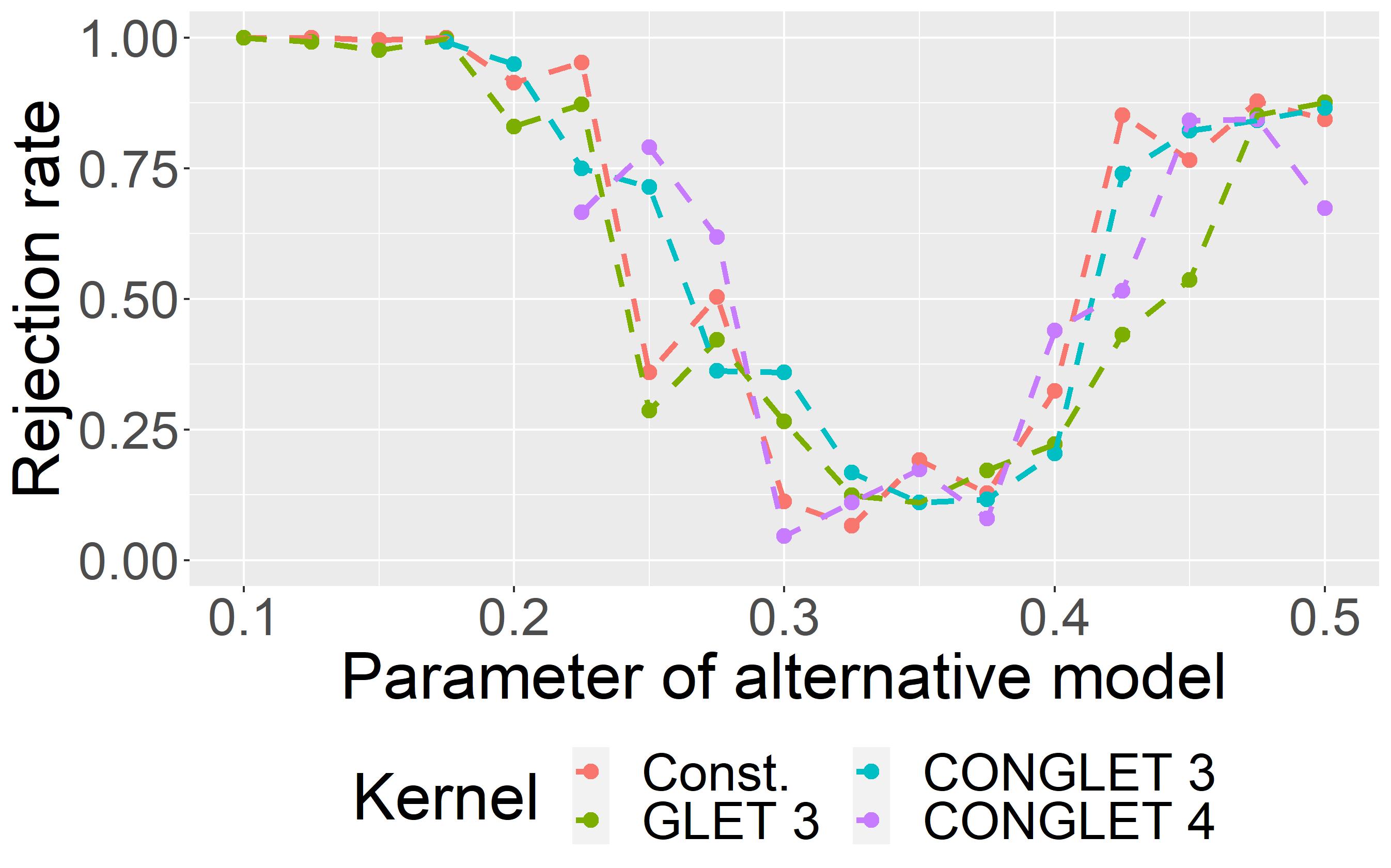}}
        {\includegraphics[width=0.32\textwidth]{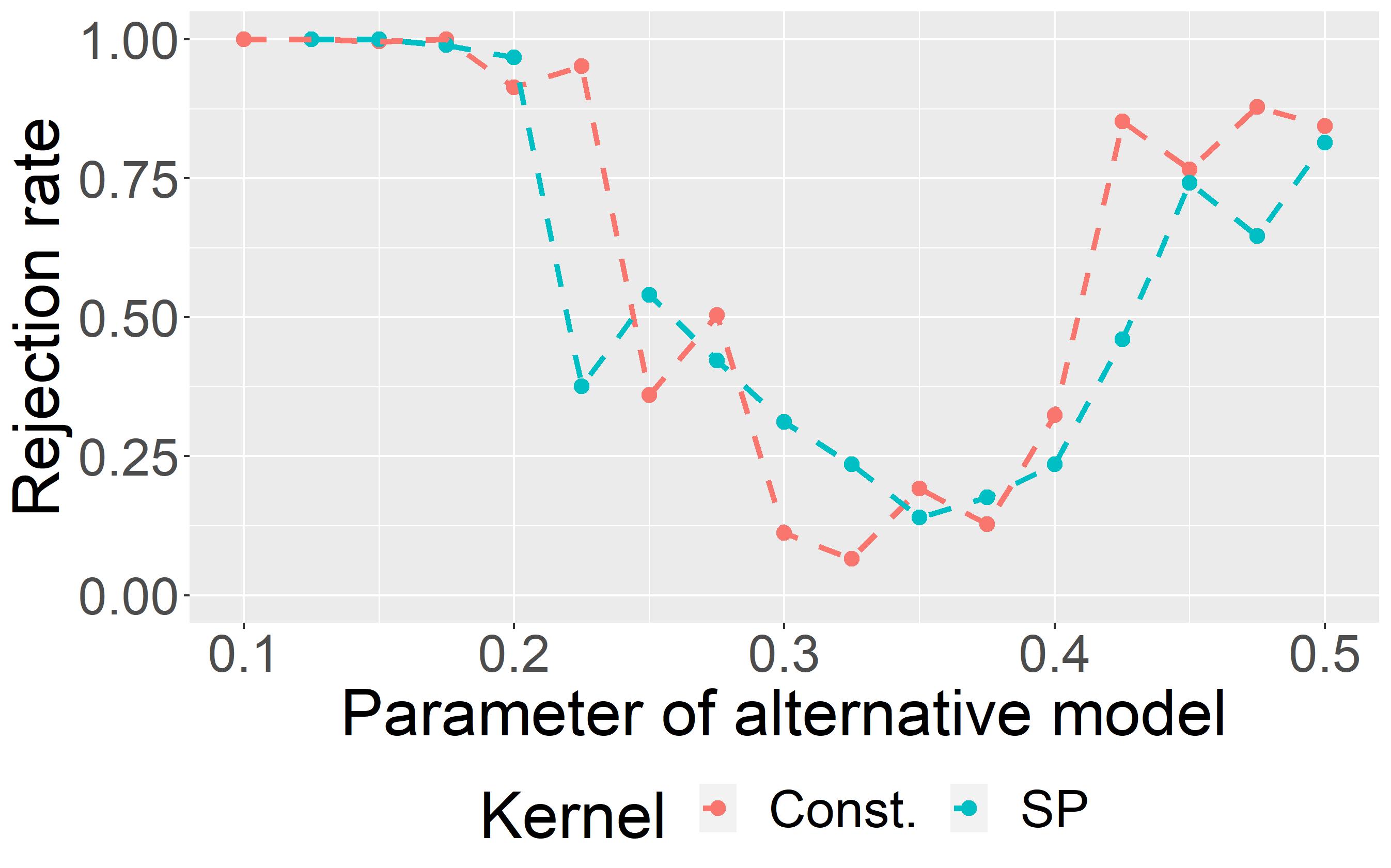}}
\includegraphics[width=0.325\textwidth]{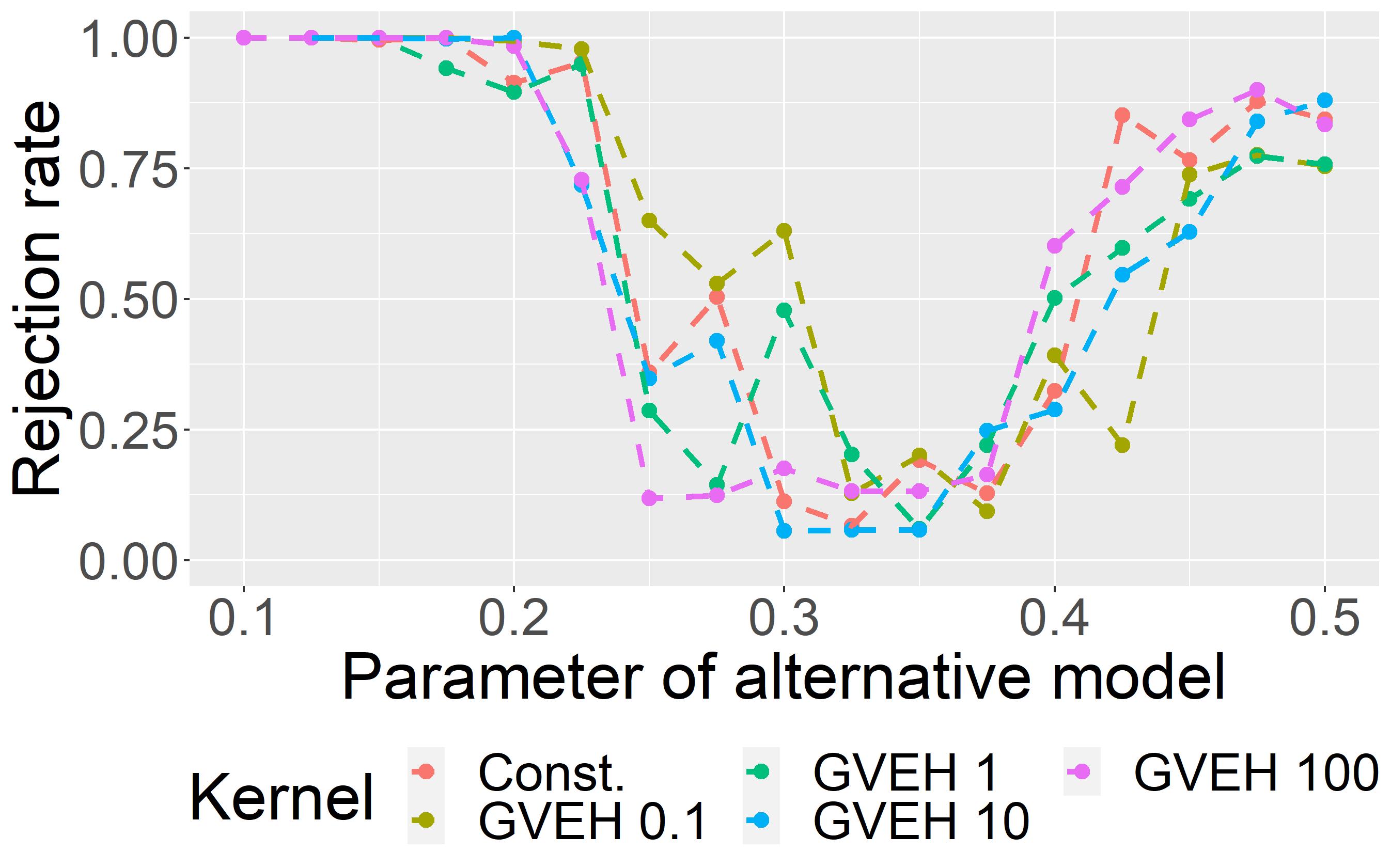}
    {\includegraphics[width=0.32\textwidth]{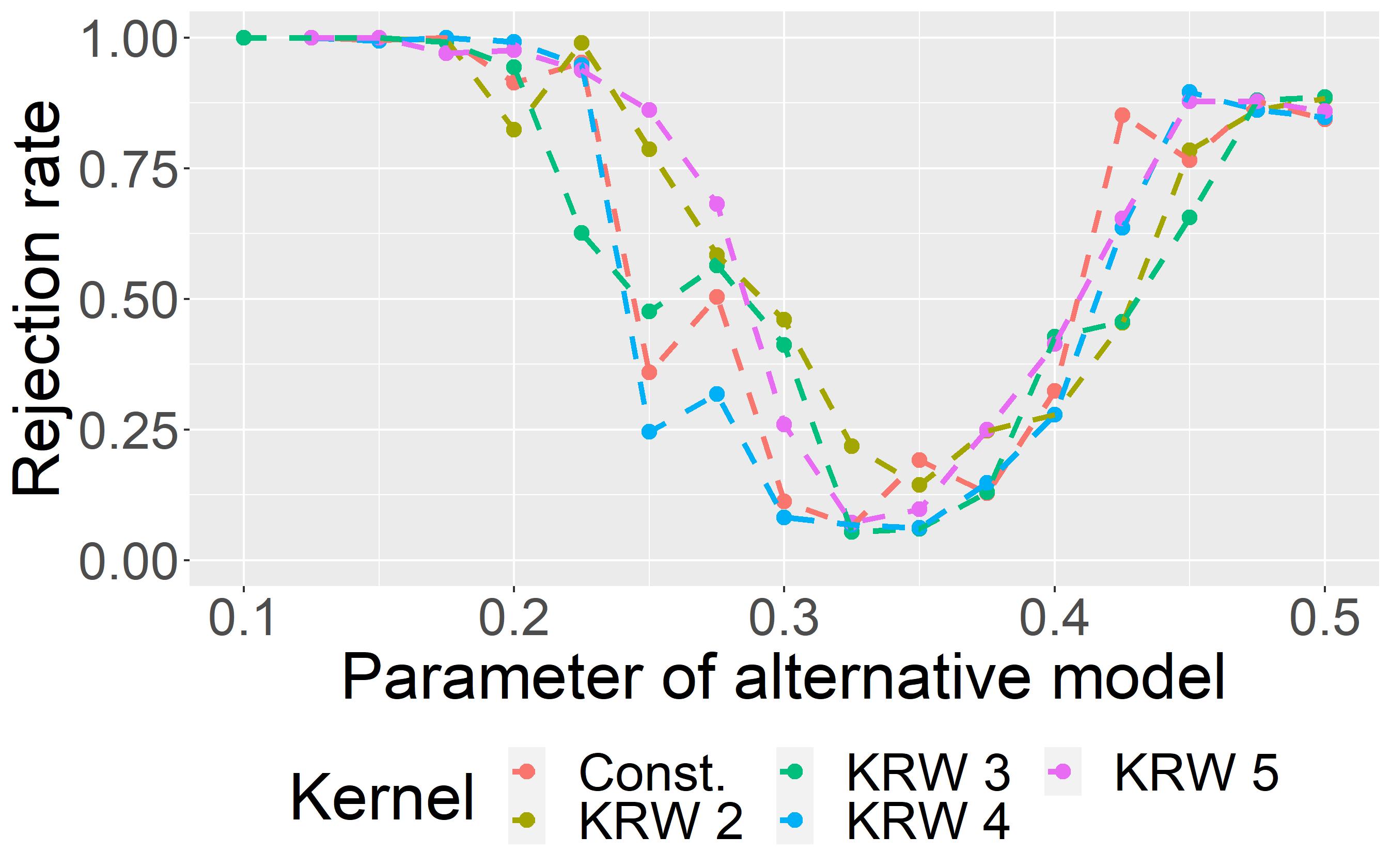}}
    \caption{
AgraSSt for GRG {on the 2-dimensional} 
unit square with $t(x)$ being the number of common neighbours.
    }
    \label{fig:grg-nt-comnb}
\end{figure*}

{In the next set of experiments, i}nstead of {on a} torus, we place the vertices on {the 2-dimensional unit square to} 
generate the geometric random graph models. The null {model} {has} 
radius $r=0.3$ while the alternative {model{s}} {have a different $r$}.
Experimental results with AgraSSt {are}  shown in \Cref{fig:grg-nt-density}, \Cref{fig:grg-nt-bideg} and \Cref{fig:grg-nt-comnb}.
{In this example all the {tested}  kernels
show a similar behaviour, with sparse alternatives easier to distinguish than denser alternatives.}

\subsubsection{
AgraSSt 
{for} geometric random graphs {on a torus: further details}}  \label{app:explain}

{For the geometric random graph on a torus, we observe a spike and subsequent dip for the bidegree statistic as well as for the common neighbour statistic, which occurs at around $r_{M1} = 0.45$.} 
{We hypothesize that th{is} phenomenon 
stems from the torus structure of the underlying space; {the behaviour on the unit square does not show this pattern}. {The torus effect} 
can be most easily seen for the common neighbour statistic. Consider two vertices placed on the unit square and imagine circles around these vertices of radius $r$, {as in \Cref{fig:explanation}}. The area of their intersection equals the probability that a randomly placed vertex is a common neighbour of the two. For small $r$, this area is large if the vertices are very close to each other and small or zero if the vertices are far apart. Hence, conditional on two vertices having a common neighbour the probability that two vertices are only a distance smaller than $r$ apart is large, and thus the probability that they are themselves connected is large as well. However, for larger $r$, the area of overlap can be large even though both vertices are far away due to the circles wrapping around the torus (see  \Cref{fig:explanation}). Thus, the information about the number of common neighbours may become less informative for larger radius $r$.

\begin{figure}[ht!]
    \centering
    		\includegraphics[width=0.34\textwidth]{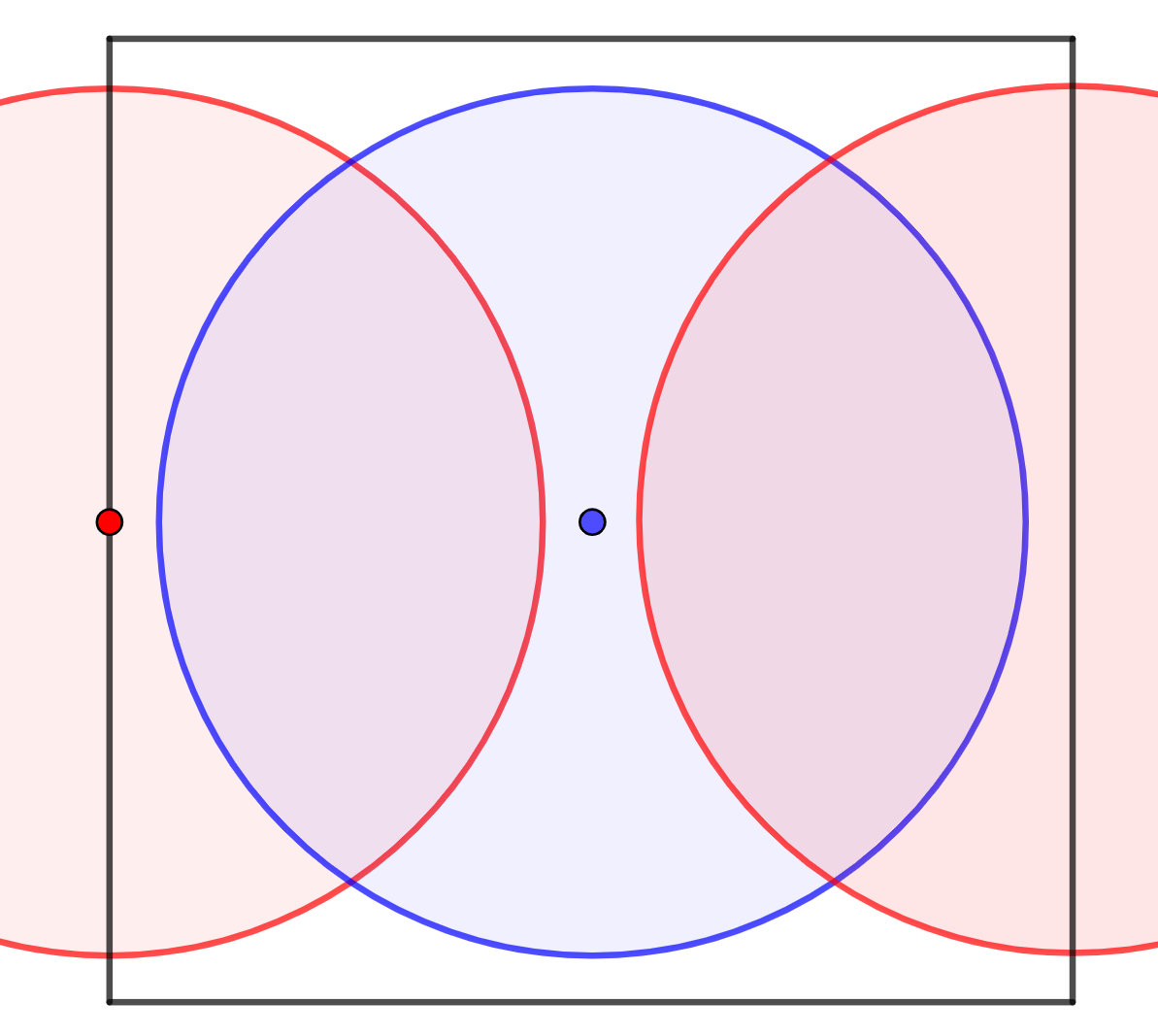}
    		\includegraphics[width=0.35\textwidth]{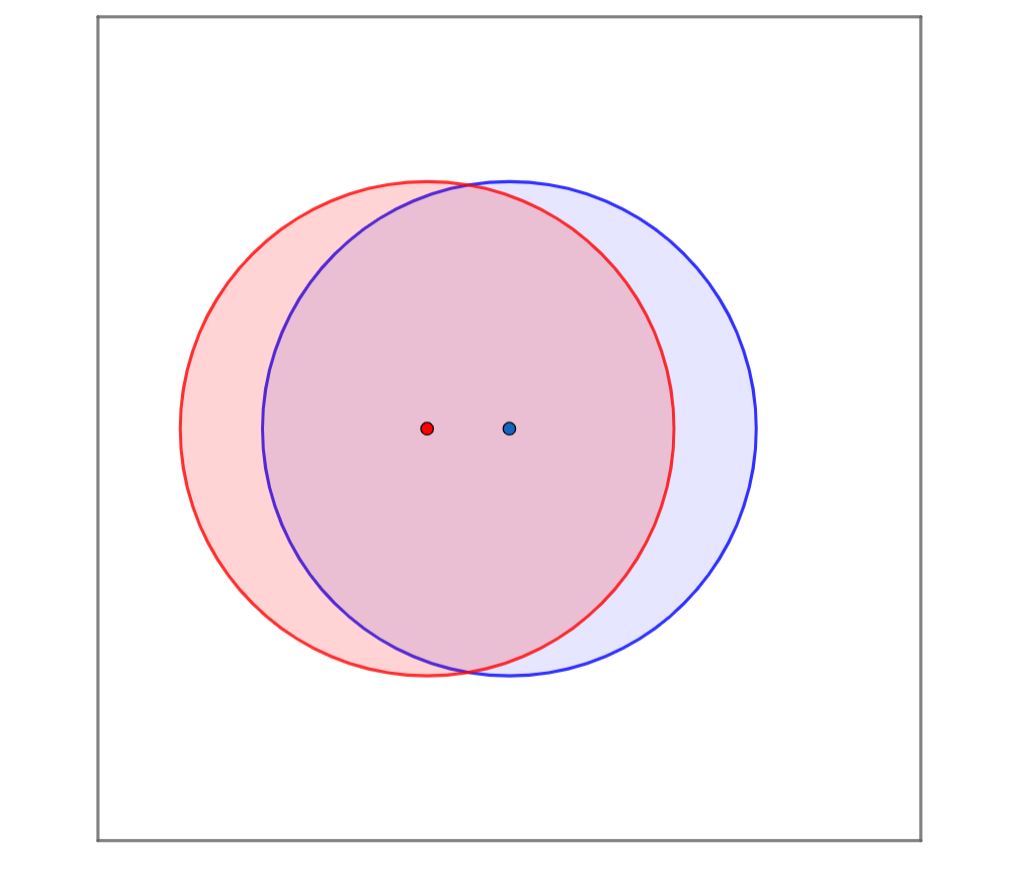}
    		\caption{\footnotesize
		Two vertices{, in red and blue,} and corresponding surrounding circles of radius
		are displayed{: left for $r= 0.45$, right for $r= 0.3$}. {Left:} Even though the vertices have a distance of $0.5$ and {are} 
		not 
		connected in the Geometric Random Graph 
		with radius $r=0.45$, their area of overlap is quite large because the red circle extends over the boundary of the unit square. Without the torus structure, the area of overlap between the two circles would only be half as large.
		{Right: Here $r=0.3$ and the two vertices are connected in the Geometric Random Graph model; the overlap of the two circles around them is similar in size to the overlap on the left-hand side. Thus, based on the overlap alone the two models are difficult to distinguish.} 
	} \label{fig:explanation} 
\end{figure}

This effect does not occur in the geometric random graph model on the two-dimensional unit square in the Euclidean space as the circles do not wrap around the edges of the square. Thus, compared to the torus topology, the area of overlap differs more depending on whether vertices are close or far away, even when $r$ is large. However, when the radius is large, {larger}
portions of the circles may be cut off by the square. Hence, the difference in intersection area is proportionally bigger between a pair of close and a pair of distant vertices, when $r$ is smaller. This explains why the rejection rates, seen in Figure \ref{fig:grg-comnb} have only one dip at the true null value $r=0.3$, but increase faster for {decreasing}  radius than for {increasing} radius.} 

\subsection{An additional experiment: Barabasi-Albert networks}\label{app:experiment-ba} 

\begin{figure*}[t!]
    \centering
    {\includegraphics[width=0.33\textwidth]{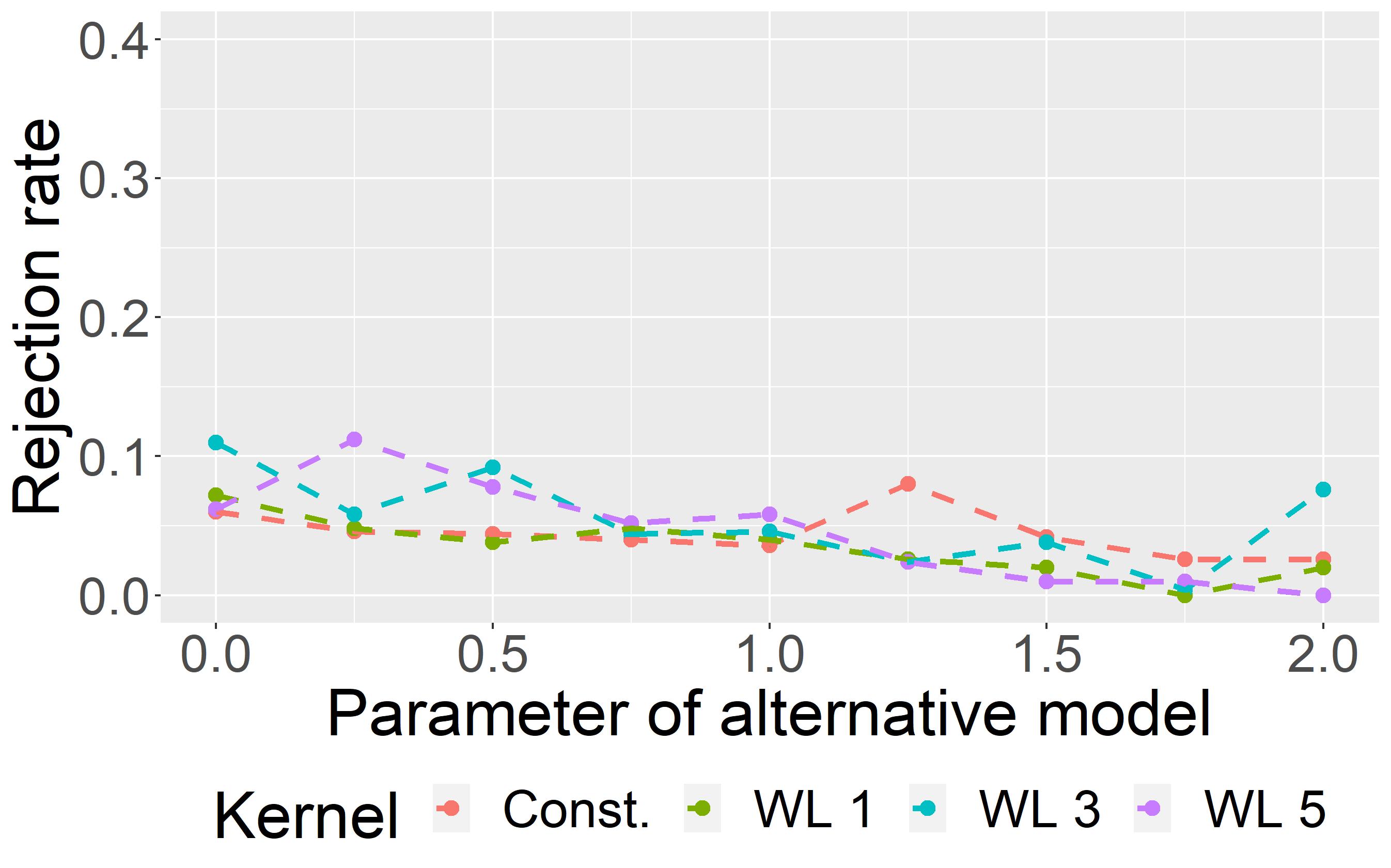}}\includegraphics[width=0.332\textwidth]{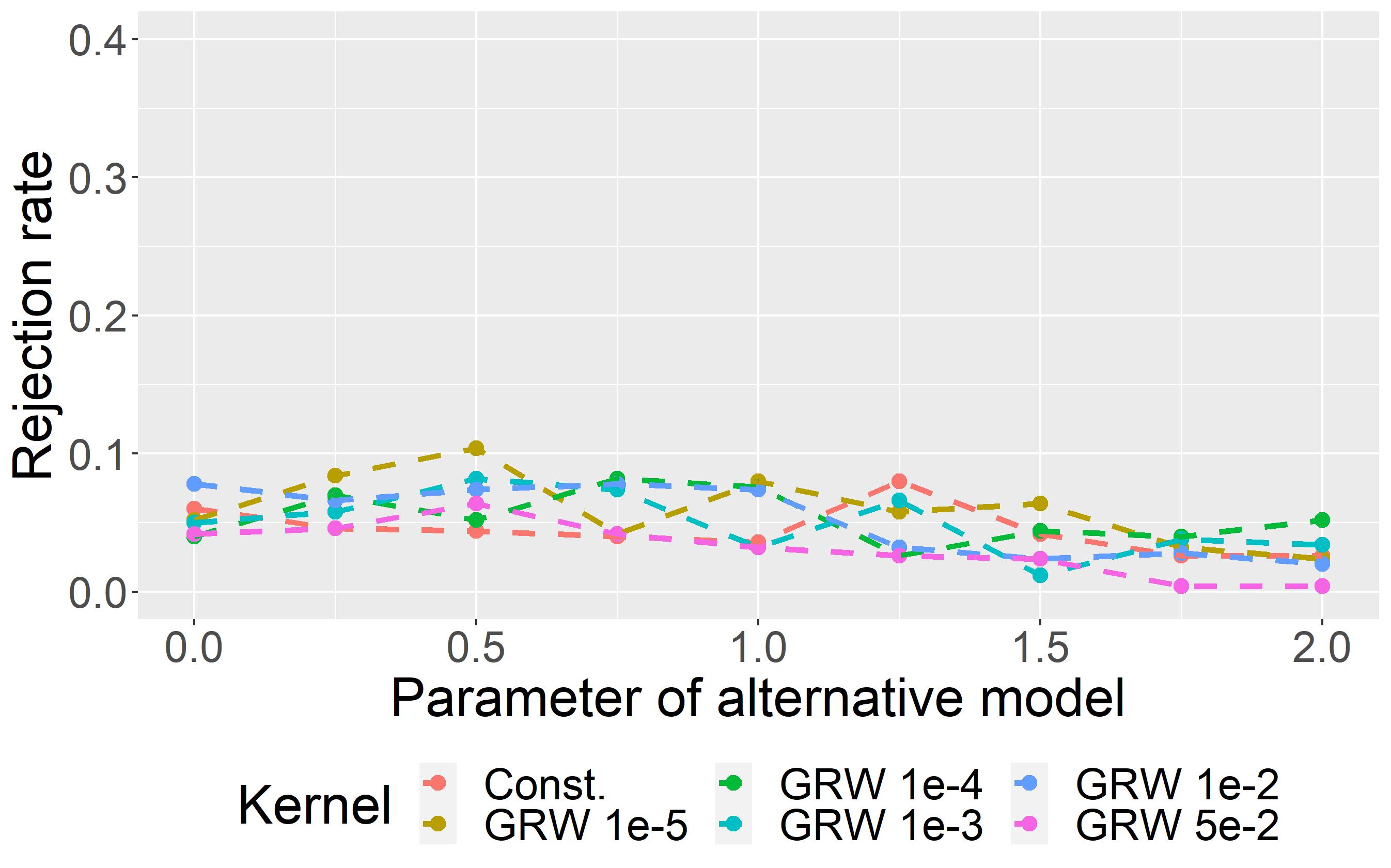}
    {\includegraphics[width=0.33\textwidth]{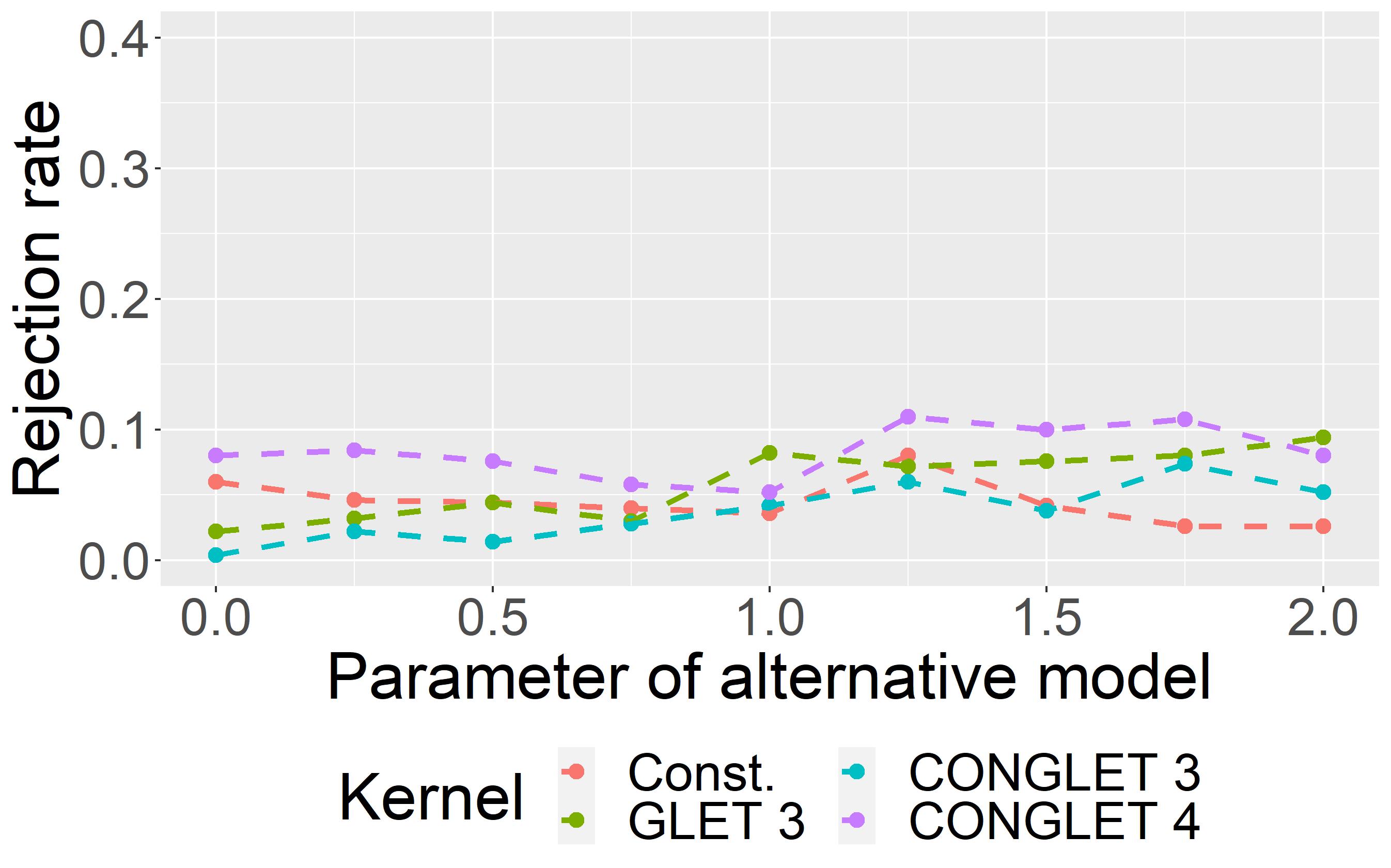}}
        {\includegraphics[width=0.32\textwidth]{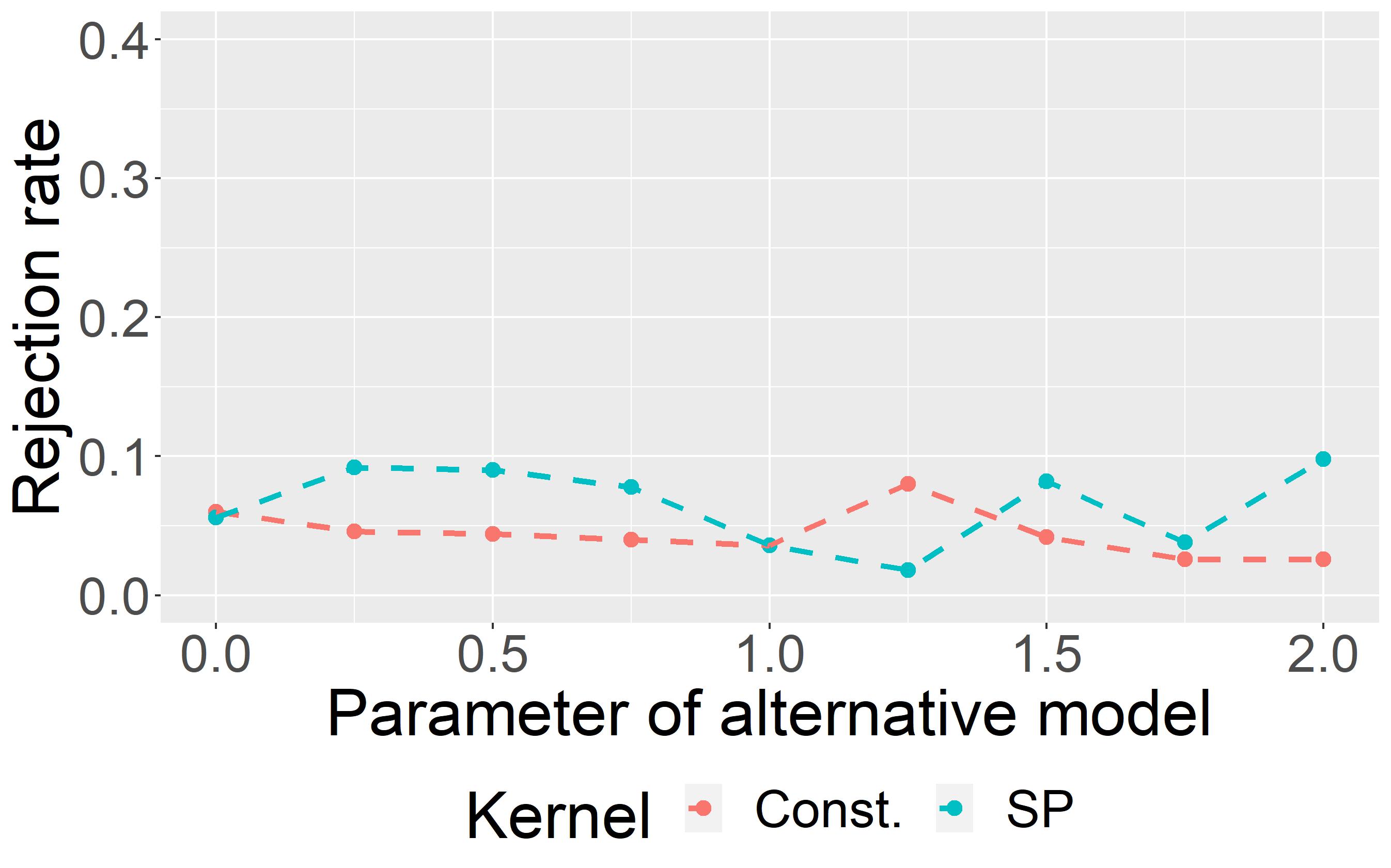}}
\includegraphics[width=0.325\textwidth]{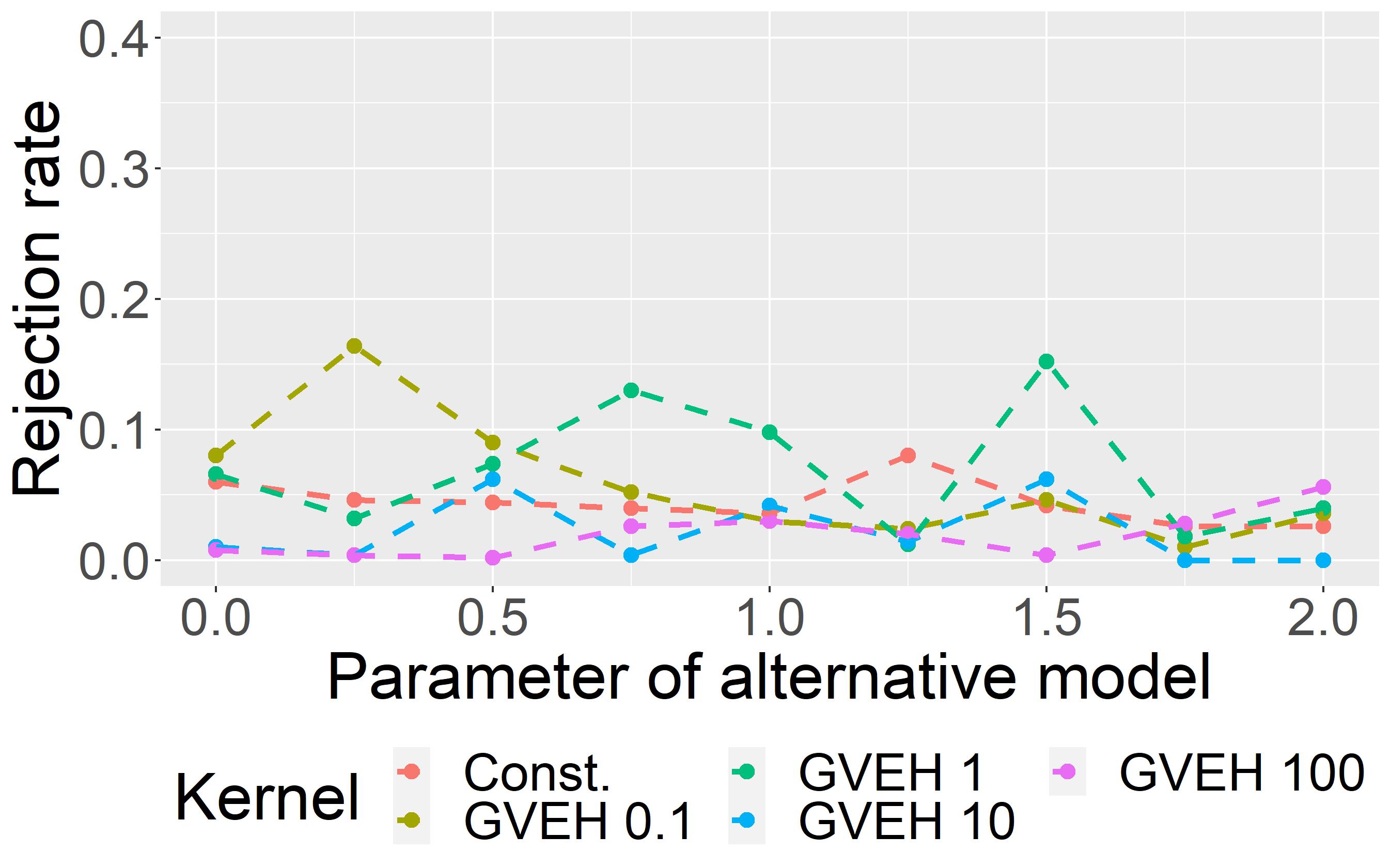}
    {\includegraphics[width=0.33\textwidth]{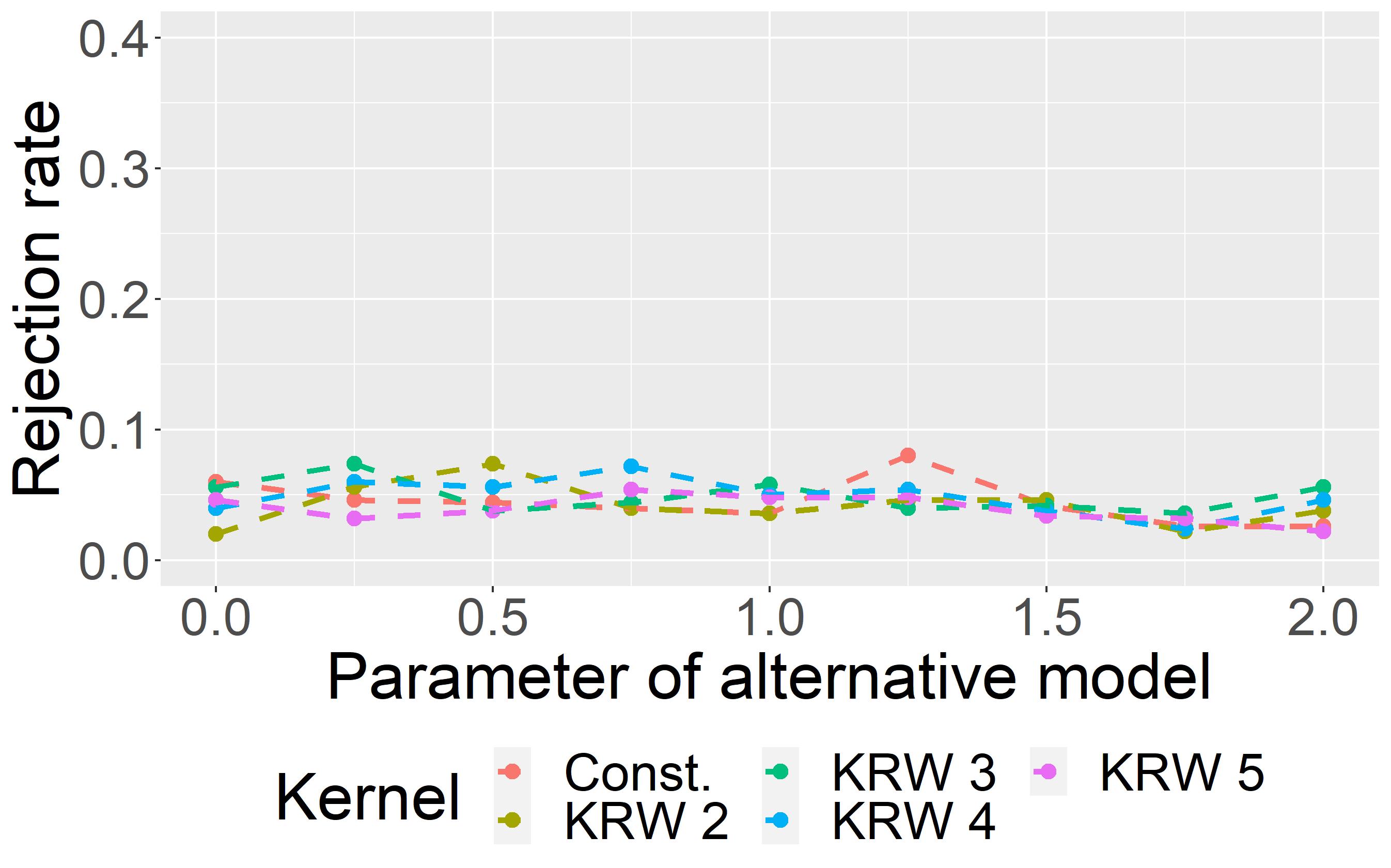}}
    \caption{
AgraSSt for BA model  with $m=1$; $t(x)$ being the edge density. }
    \label{fig:ba1-density}
\end{figure*}

\begin{figure*}[t!]
    \centering
    {\includegraphics[width=0.33\textwidth]{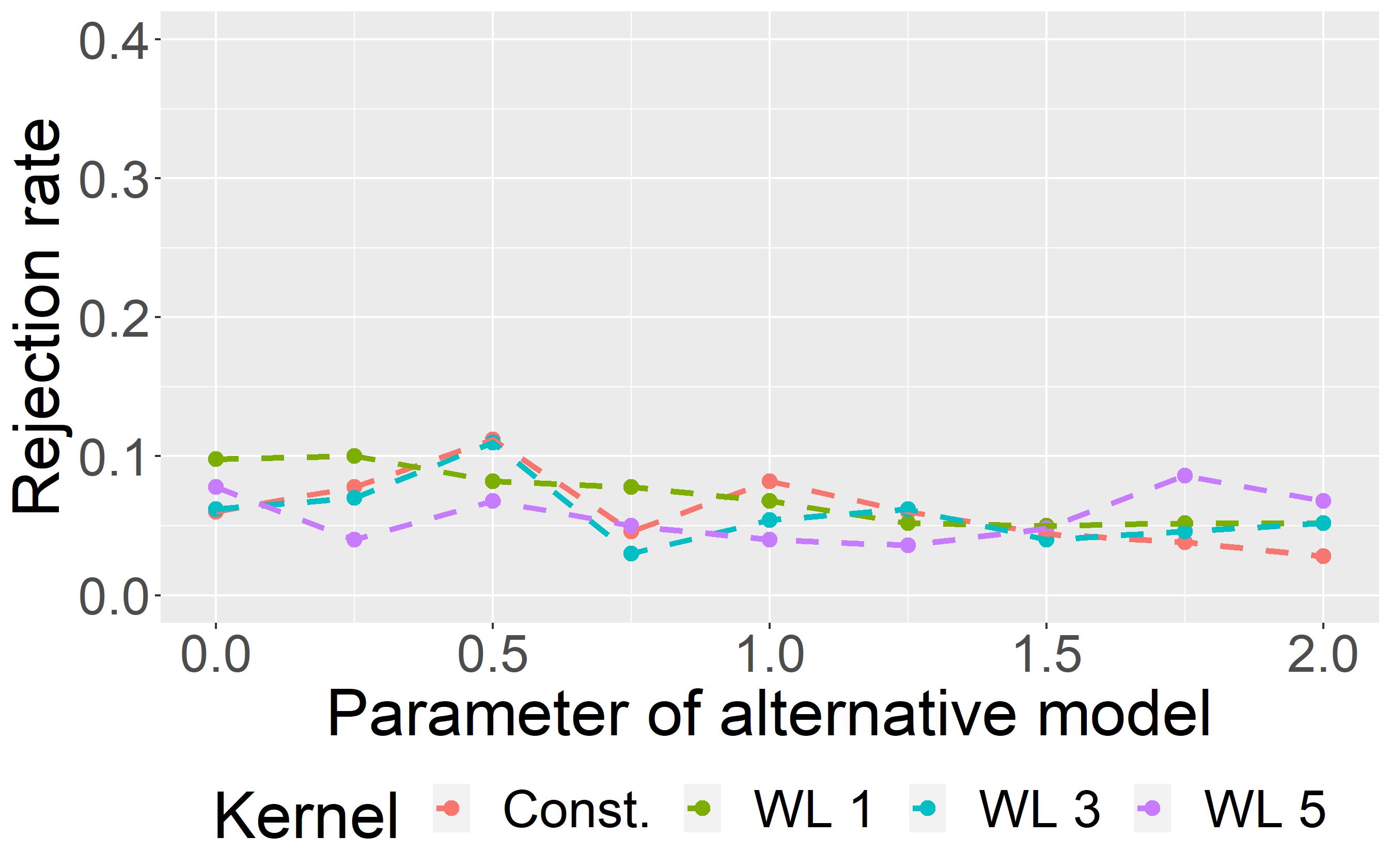}}\includegraphics[width=0.332\textwidth]{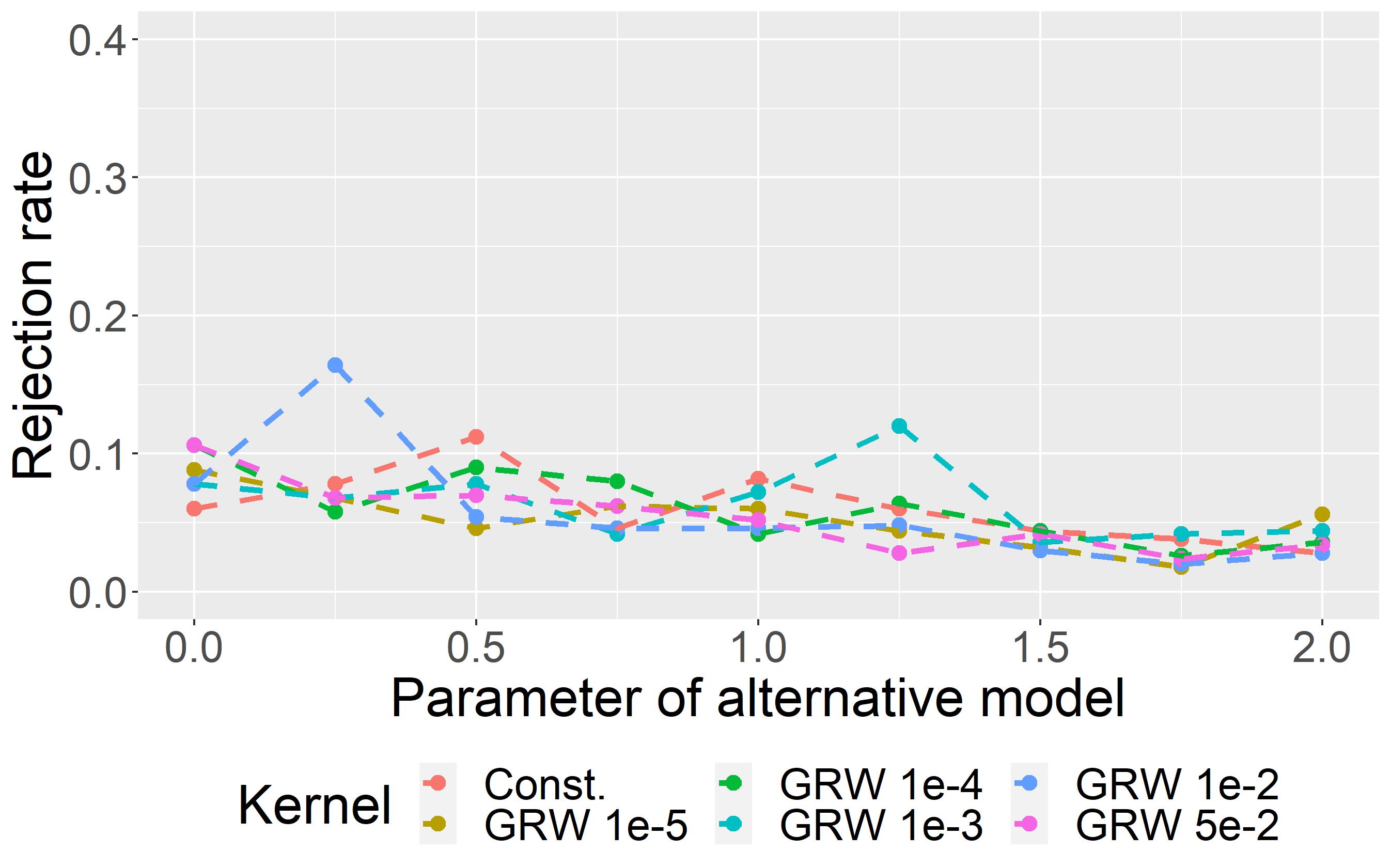}
    {\includegraphics[width=0.33\textwidth]{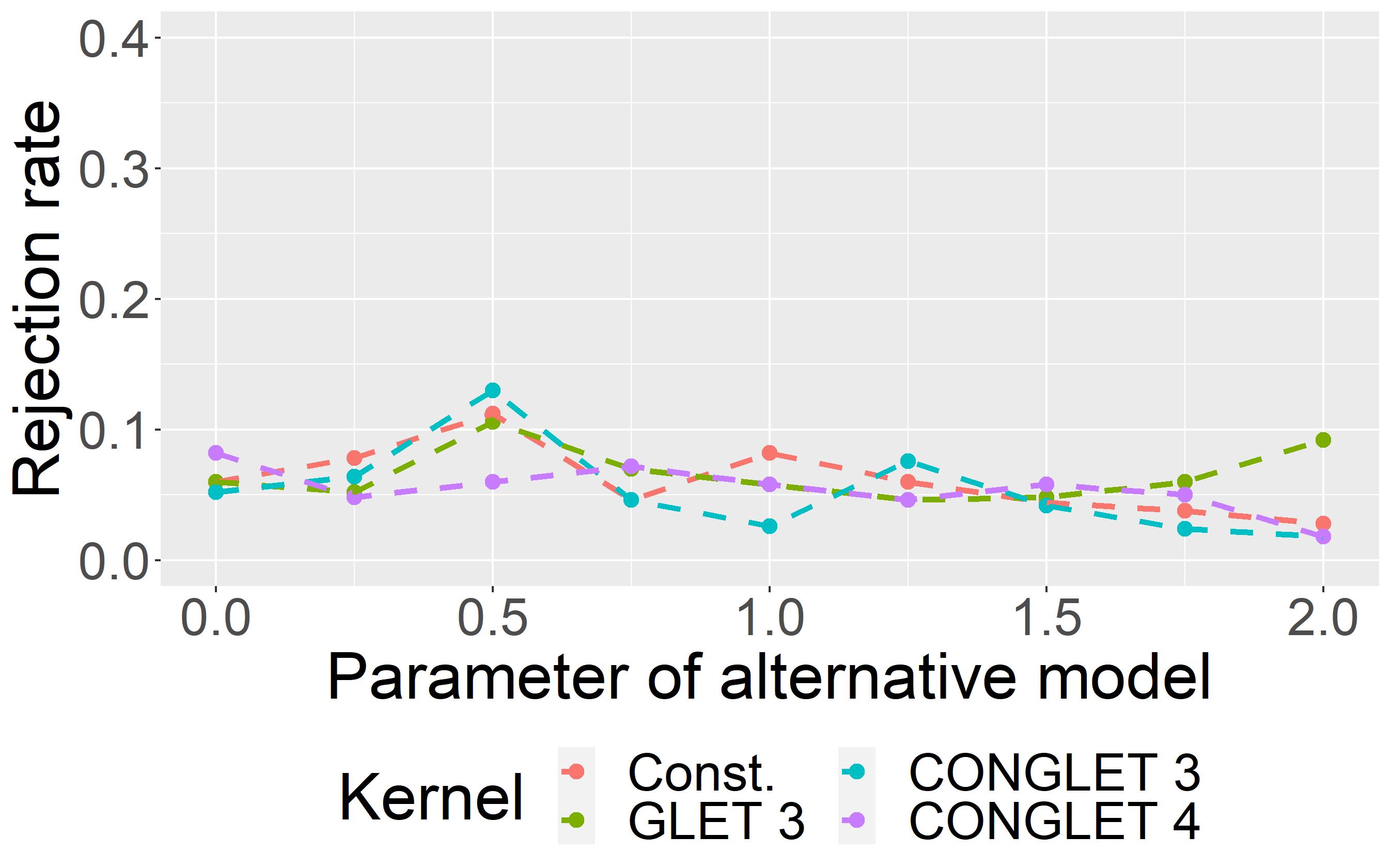}}
        {\includegraphics[width=0.32\textwidth]{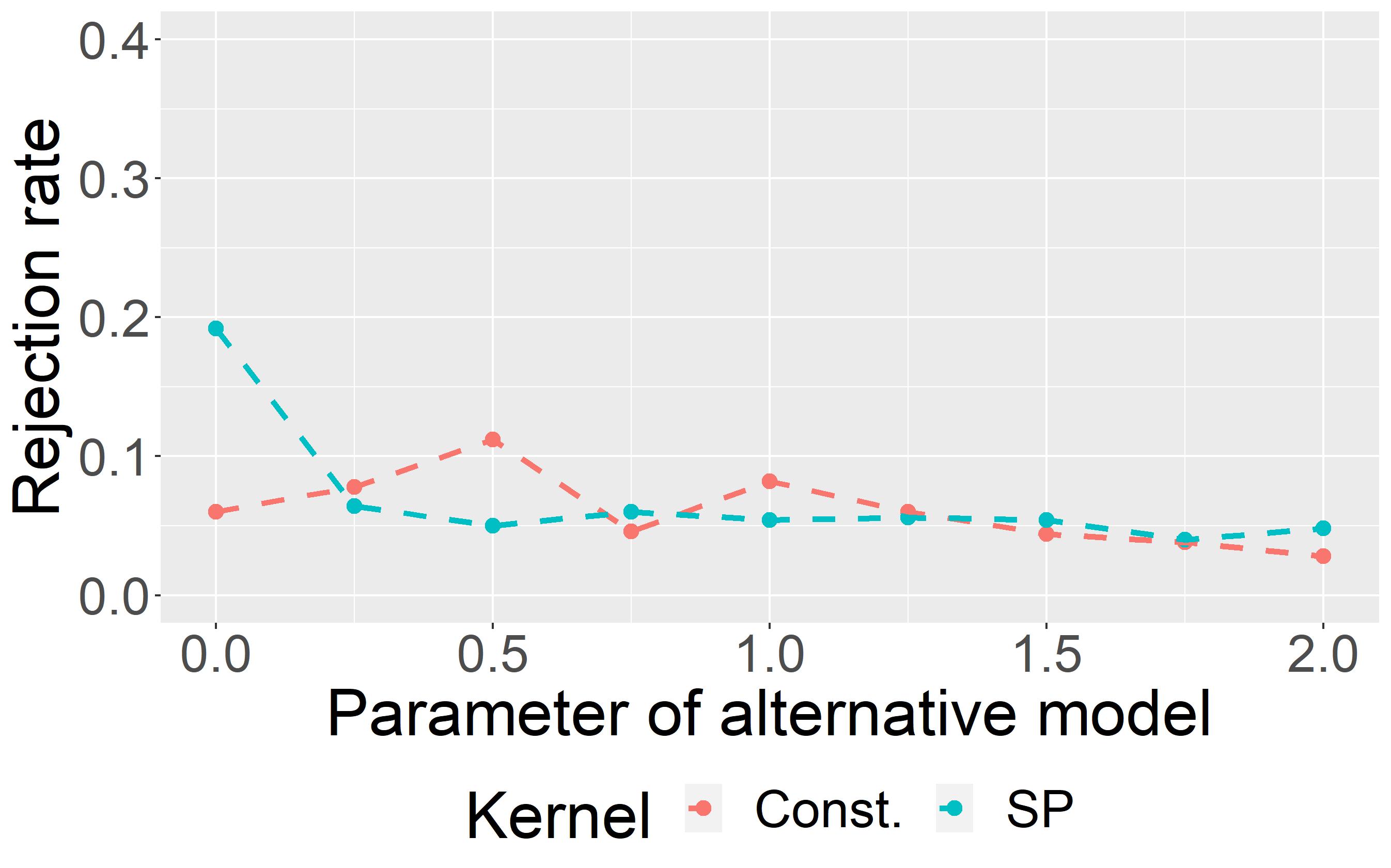}}
\includegraphics[width=0.325\textwidth]{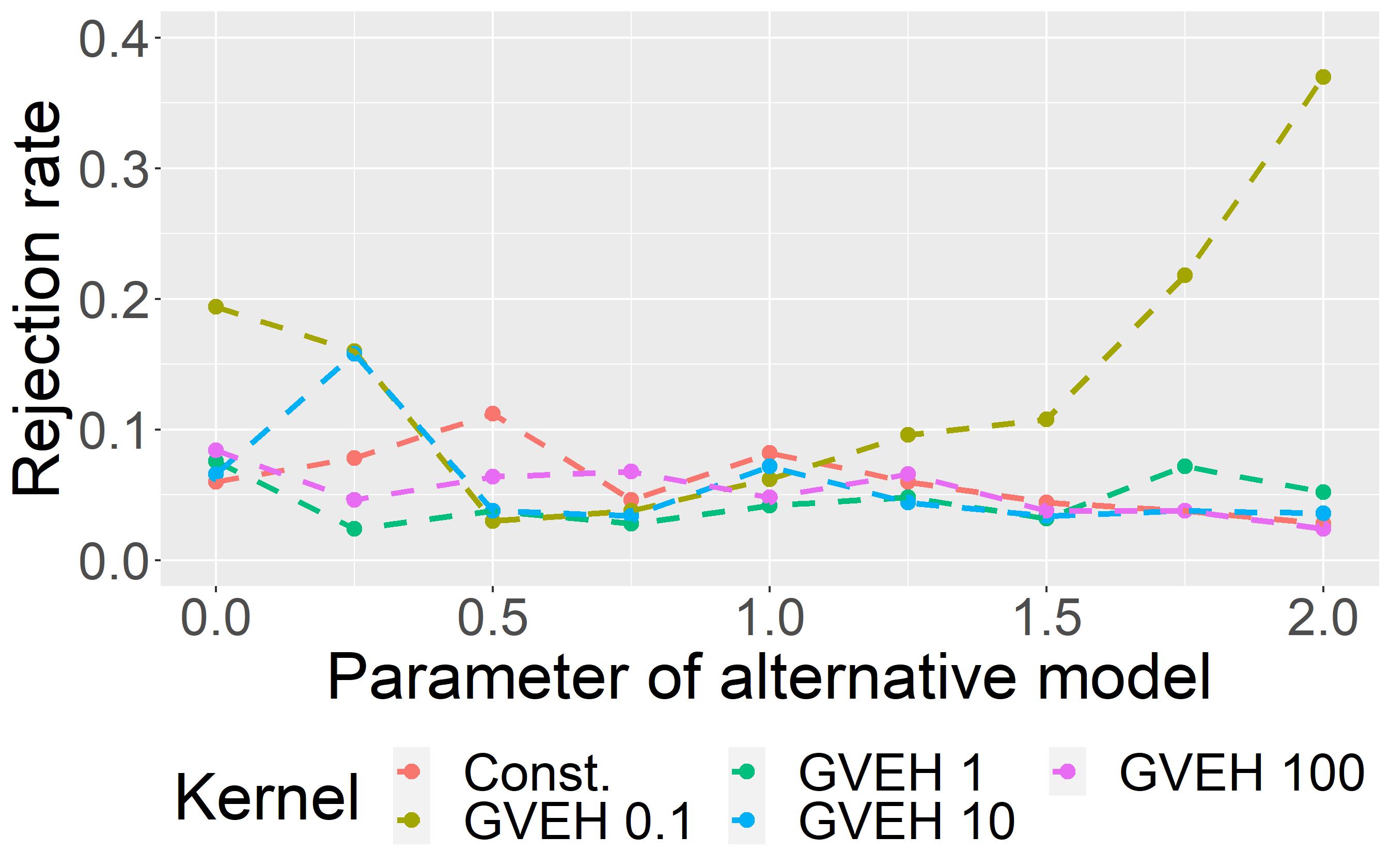}
    {\includegraphics[width=0.33\textwidth]{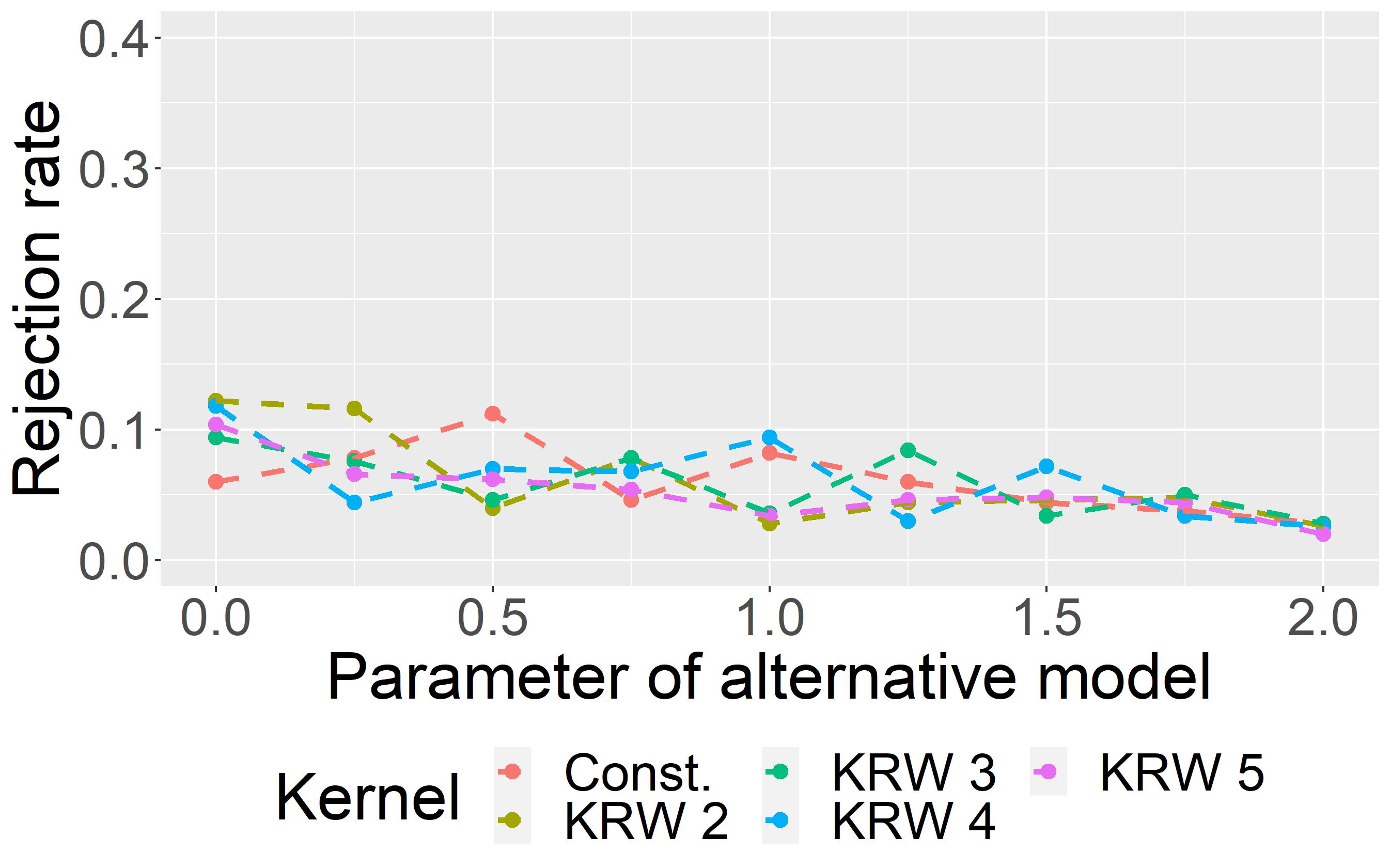}}
    \caption{
    AgraSSt for BA model with $m=1$; $t(x)$ being the bivariate degree vector.}
    \label{fig:ba1-bideg}
\end{figure*}

\begin{figure*}[t!]
    \centering
    {\includegraphics[width=0.33\textwidth]{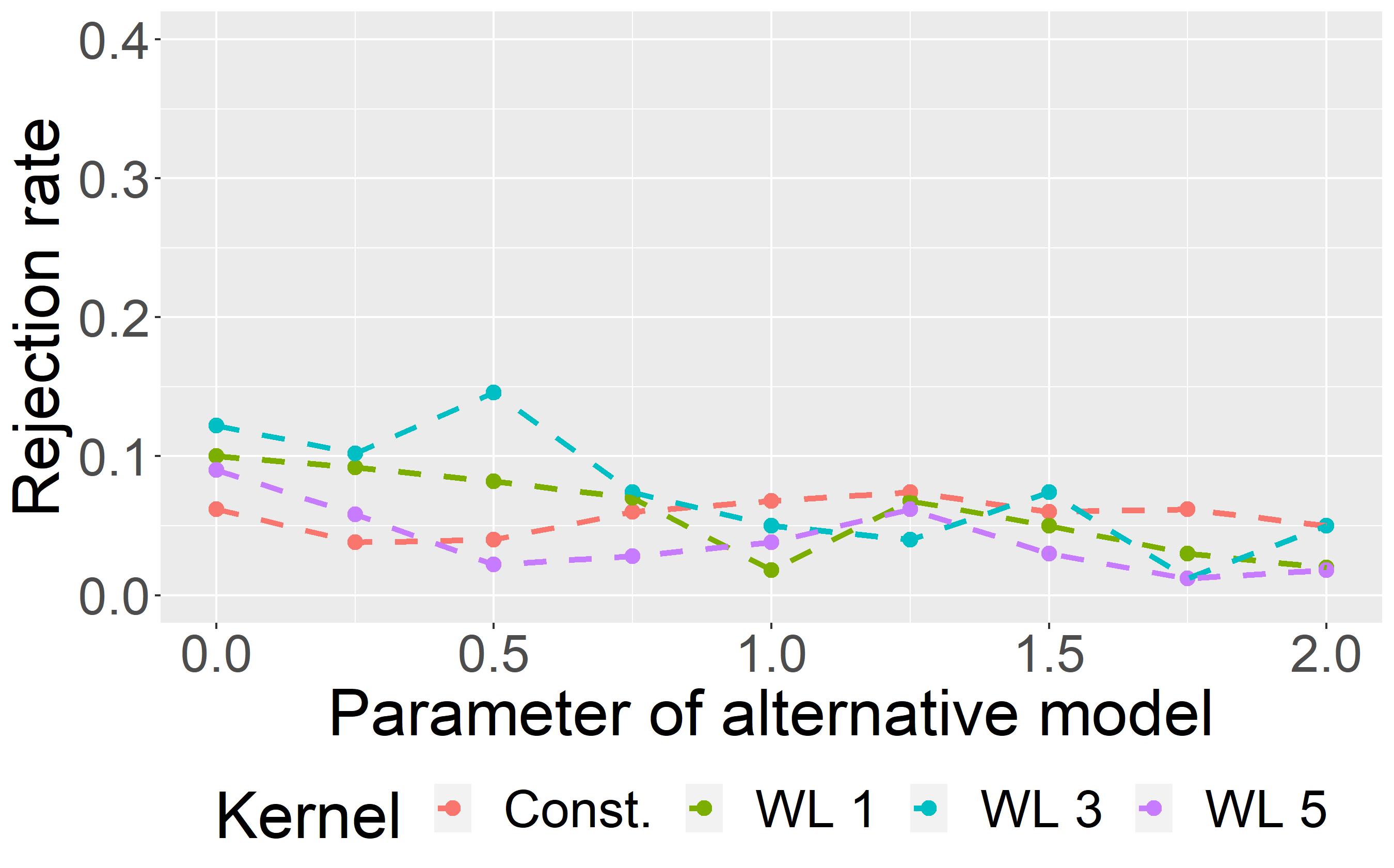}}\includegraphics[width=0.332\textwidth]{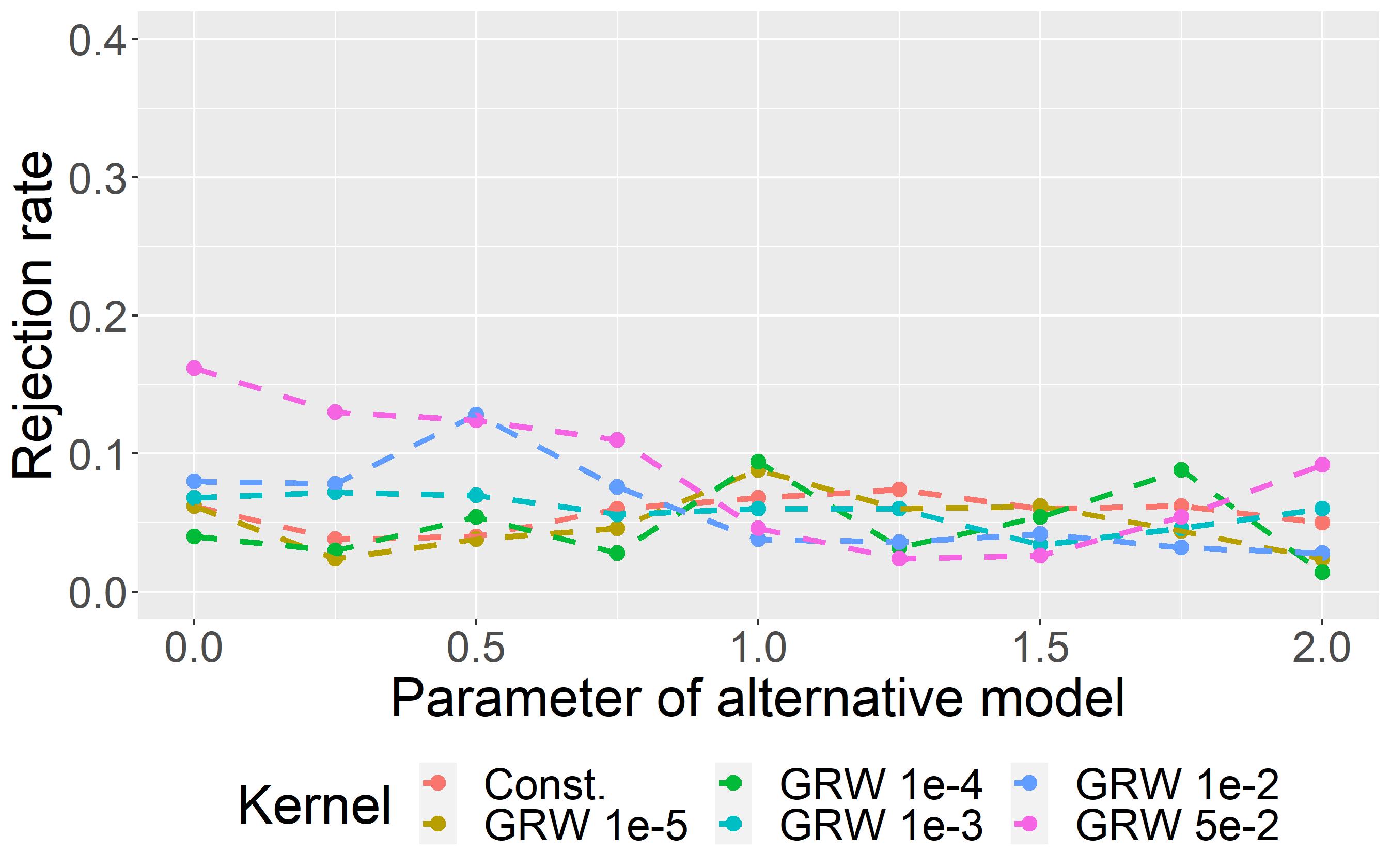}
    {\includegraphics[width=0.33\textwidth]{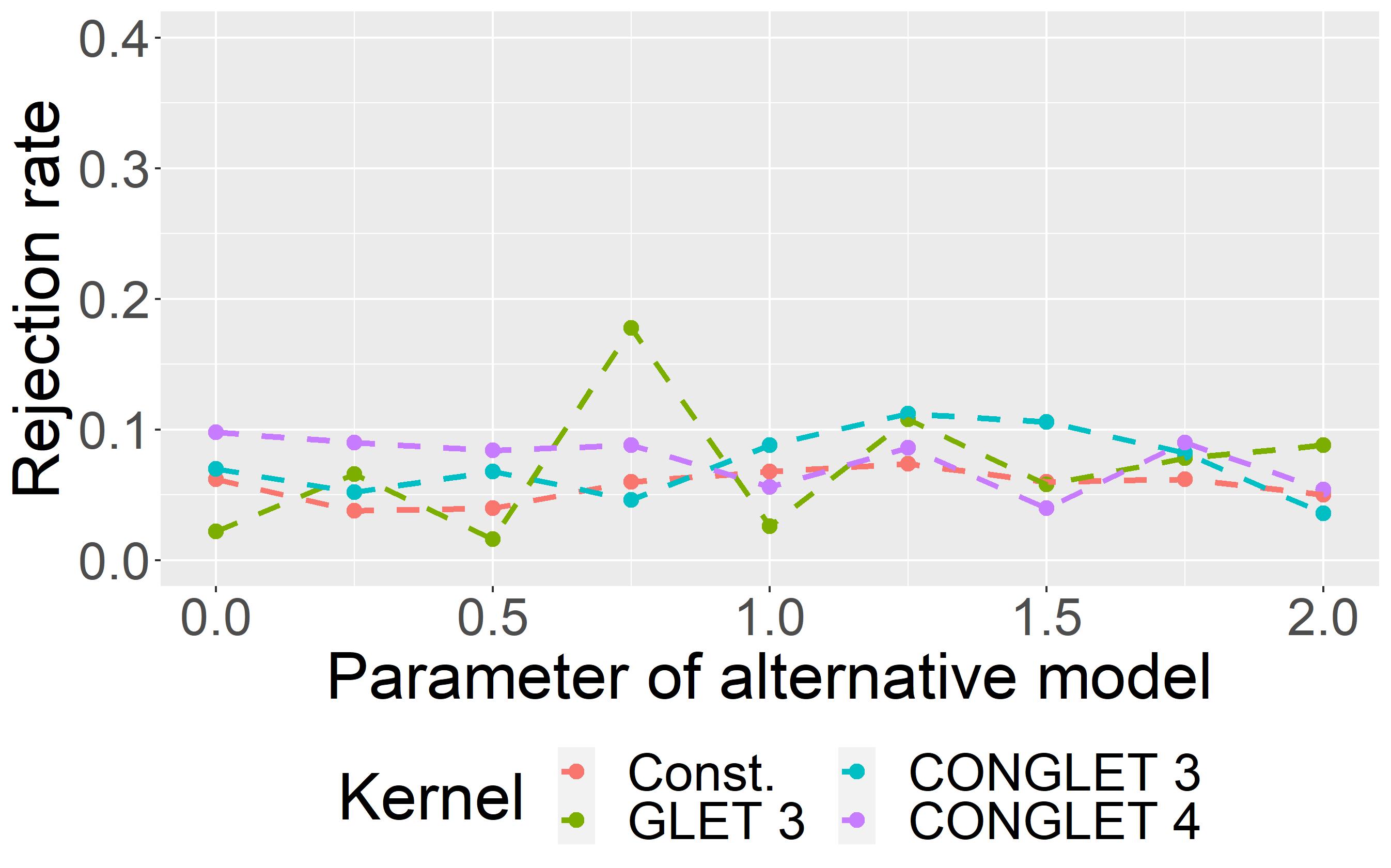}}
        {\includegraphics[width=0.32\textwidth]{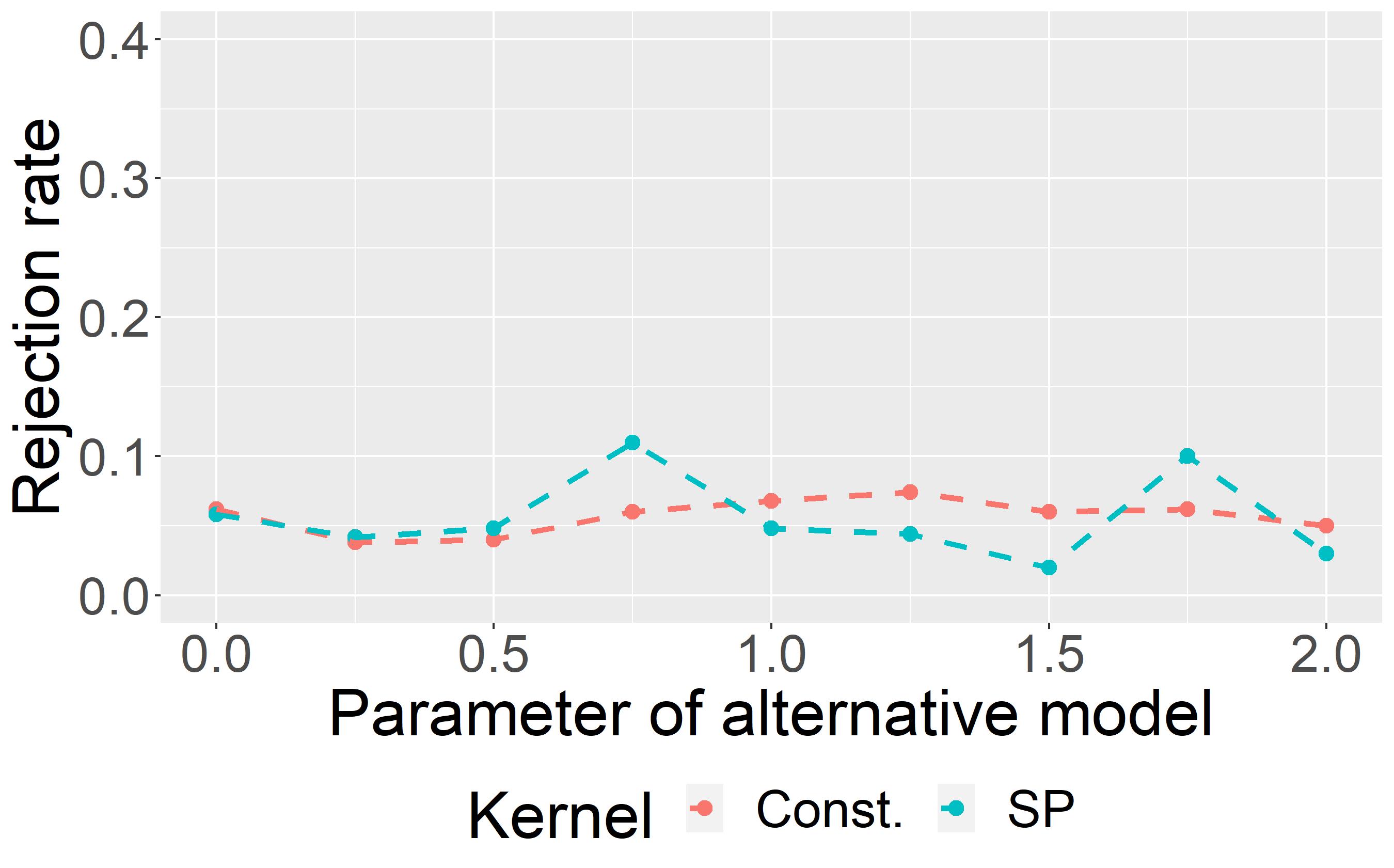}}
\includegraphics[width=0.325\textwidth]{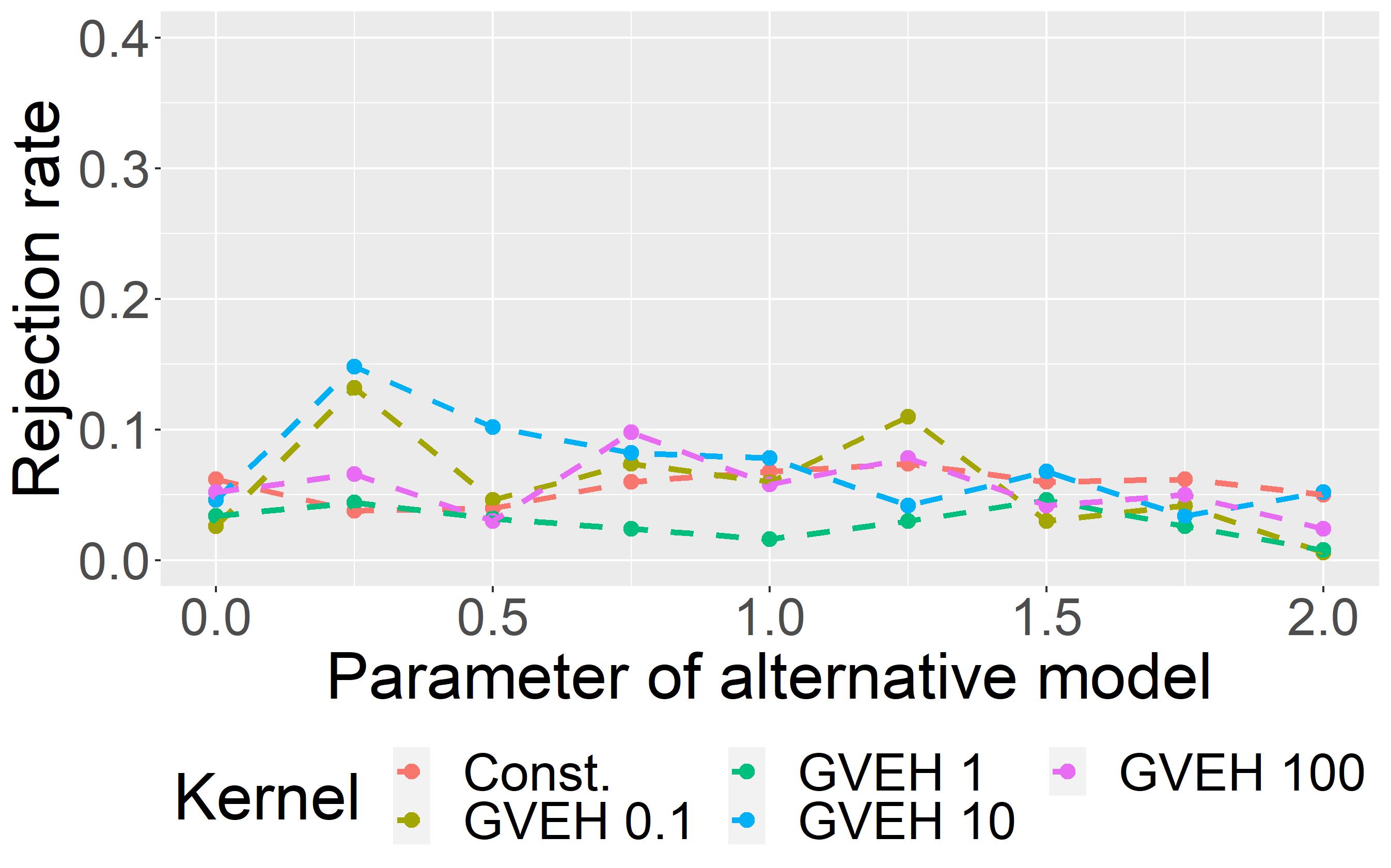}
    {\includegraphics[width=0.33\textwidth]{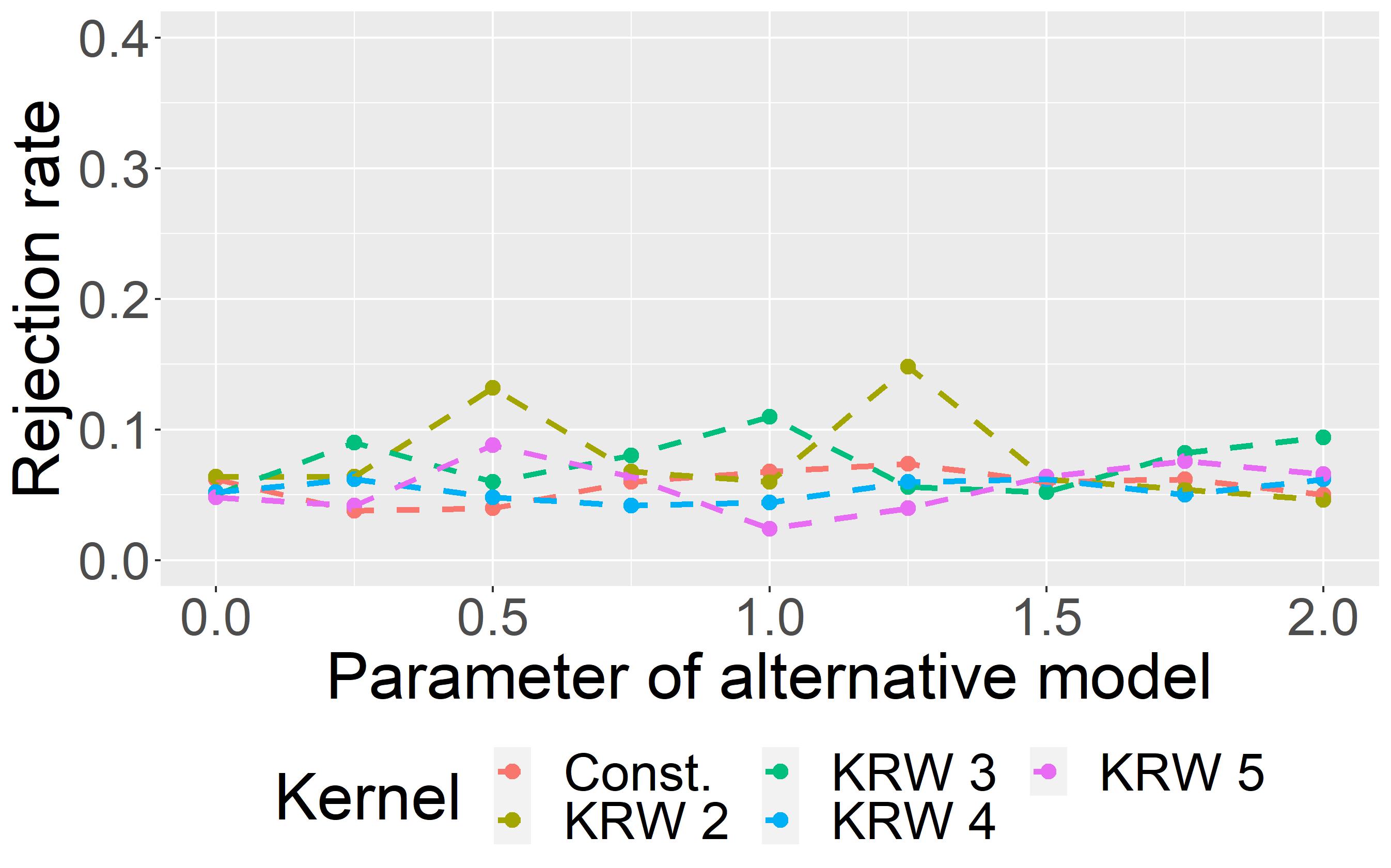}}
    \caption{
AgraSSt for BA model  with $m=2$; $t(x)$ being the edge density. }
    \label{fig:ba-density}
\end{figure*}
\begin{figure*}[t!]
    \centering
    {\includegraphics[width=0.33\textwidth]{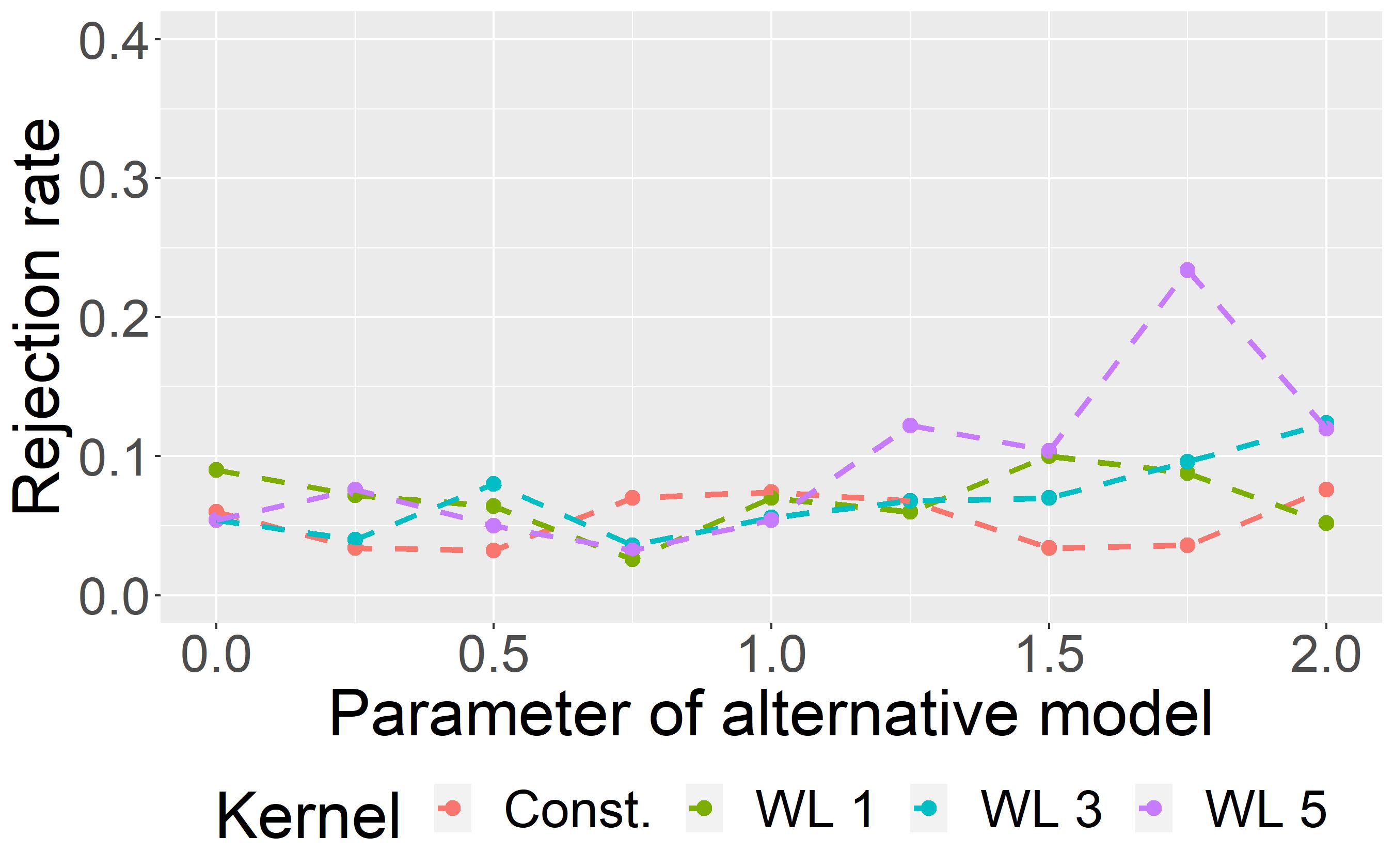}}\includegraphics[width=0.332\textwidth]{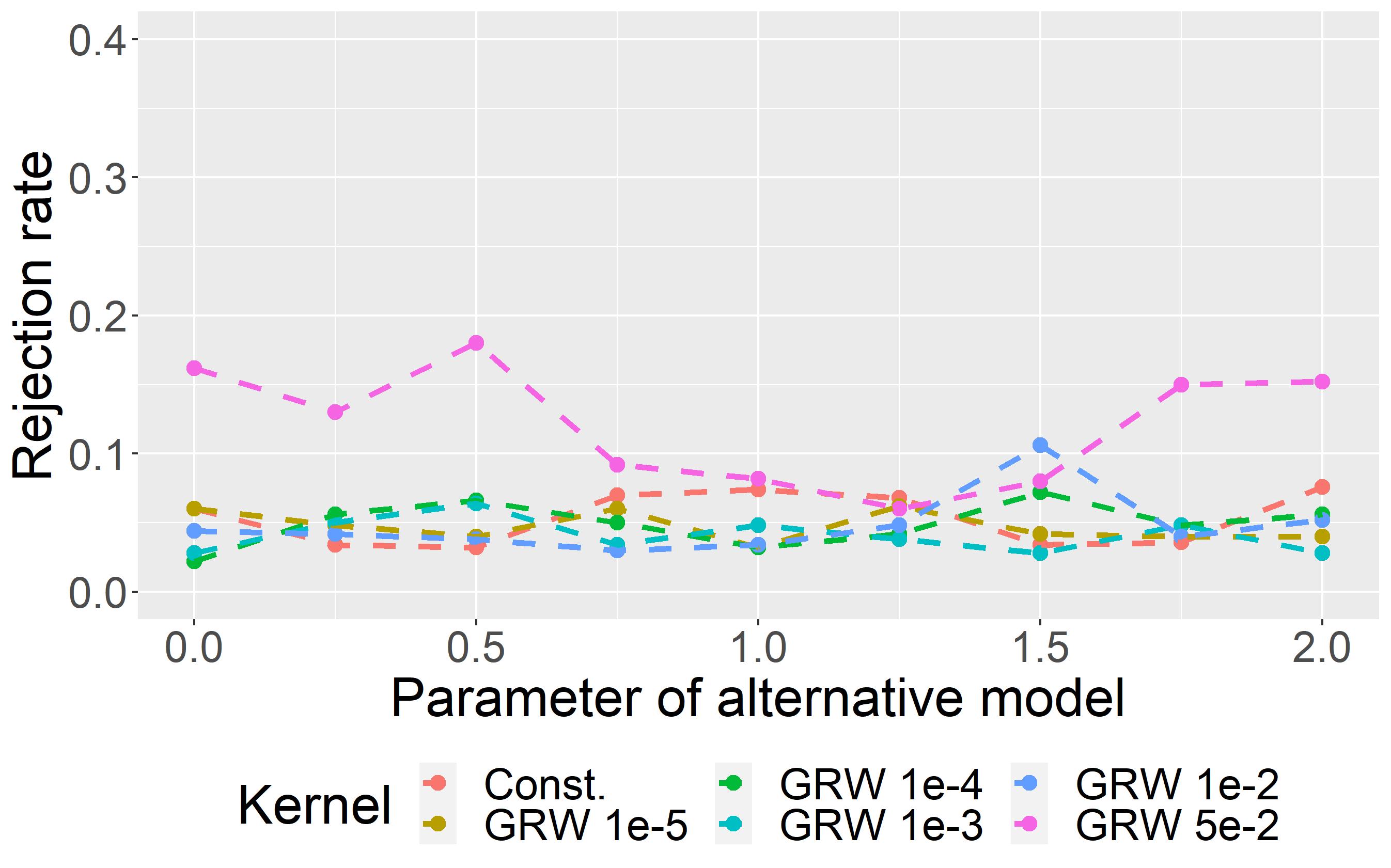}
    {\includegraphics[width=0.328\textwidth]{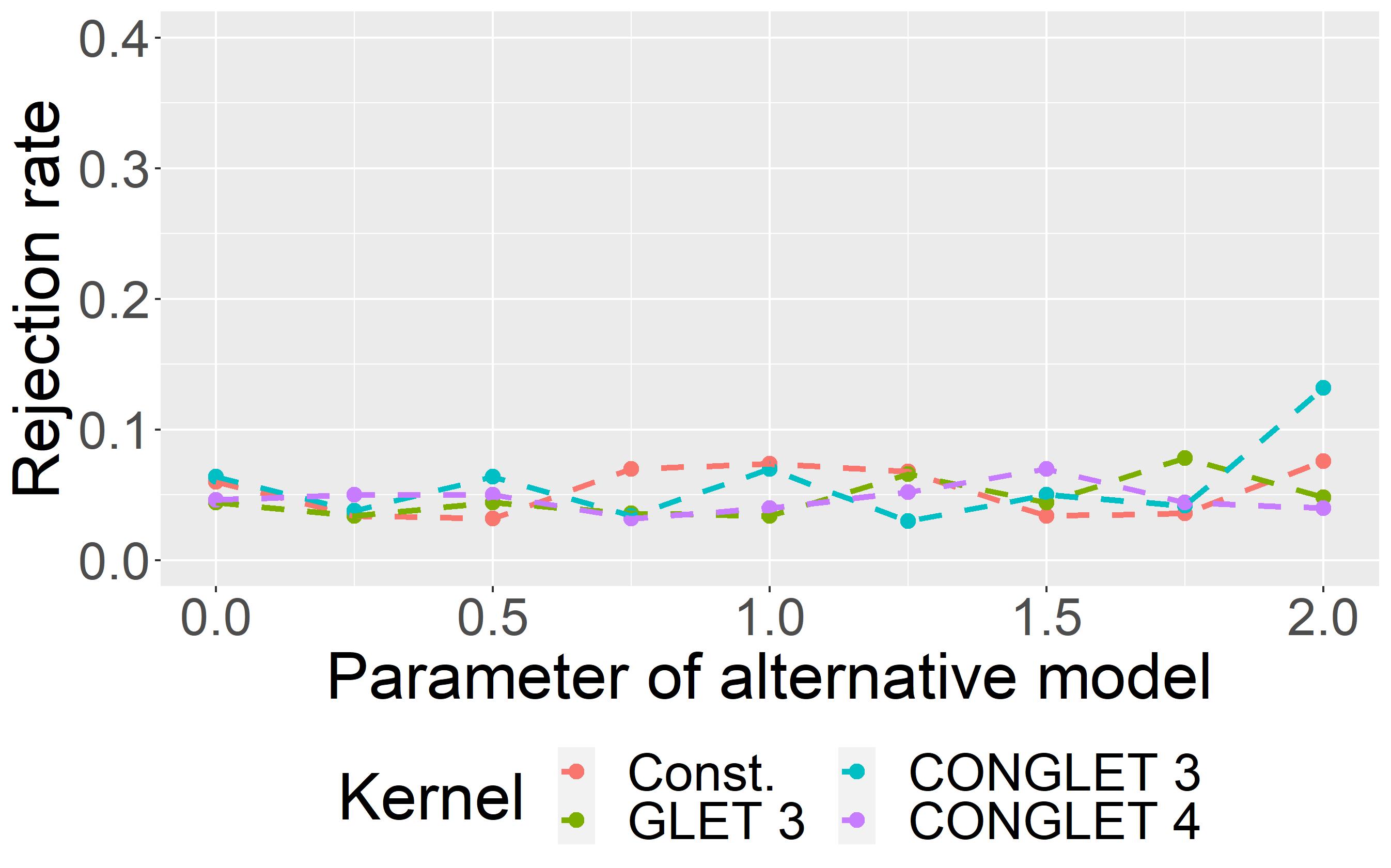}}
        {\includegraphics[width=0.32\textwidth]{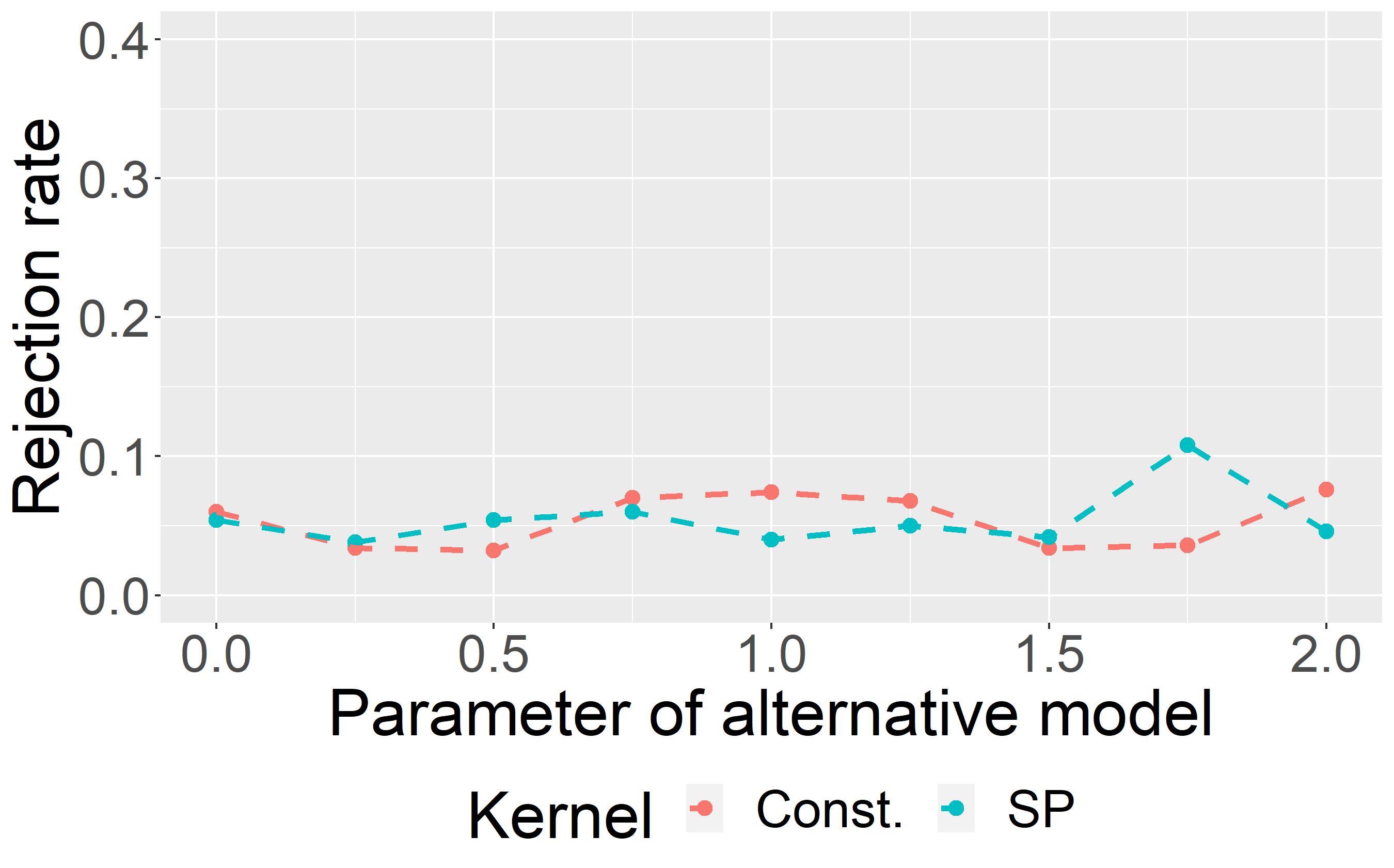}}
\includegraphics[width=0.325\textwidth]{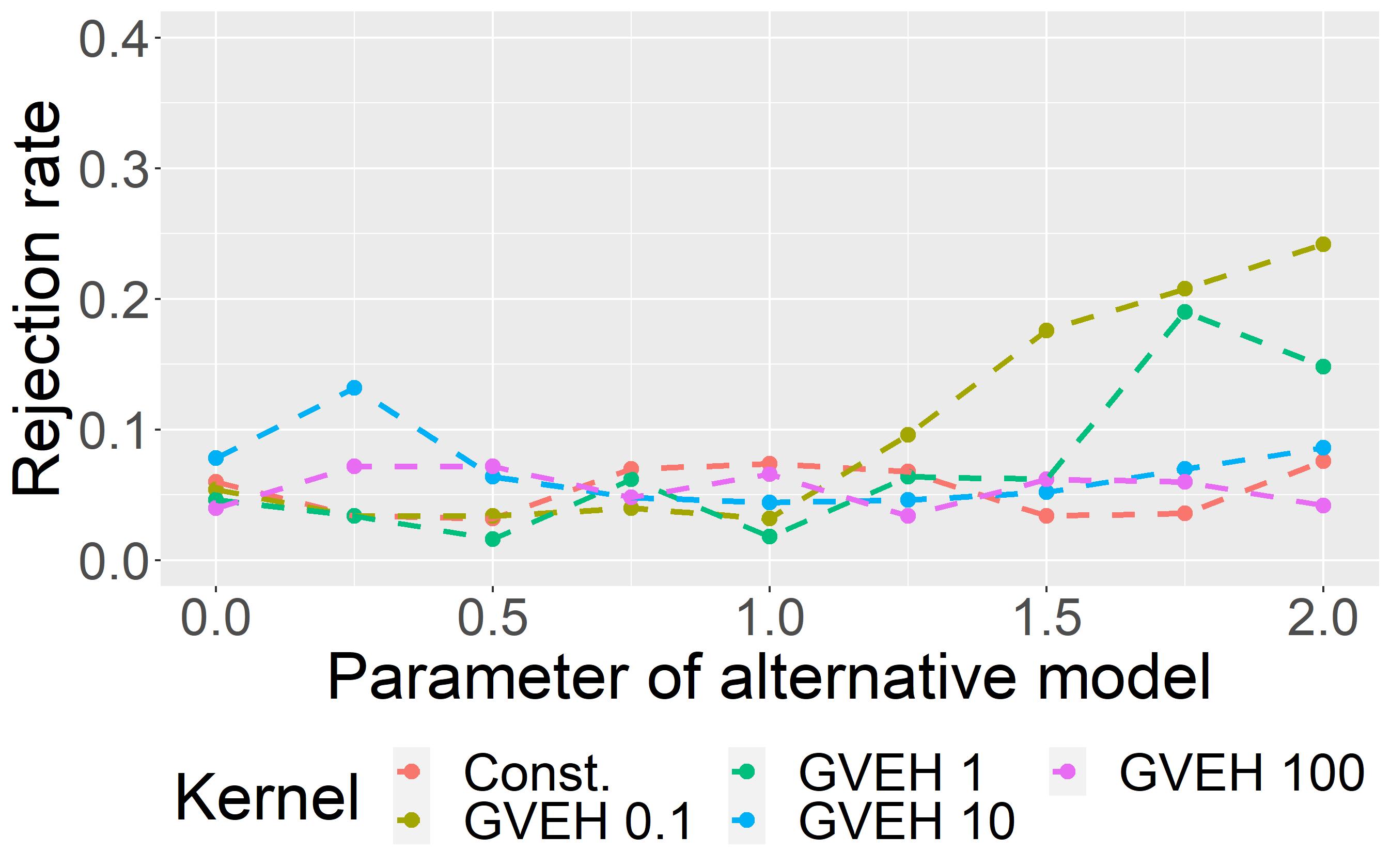}
    {\includegraphics[width=0.328\textwidth]{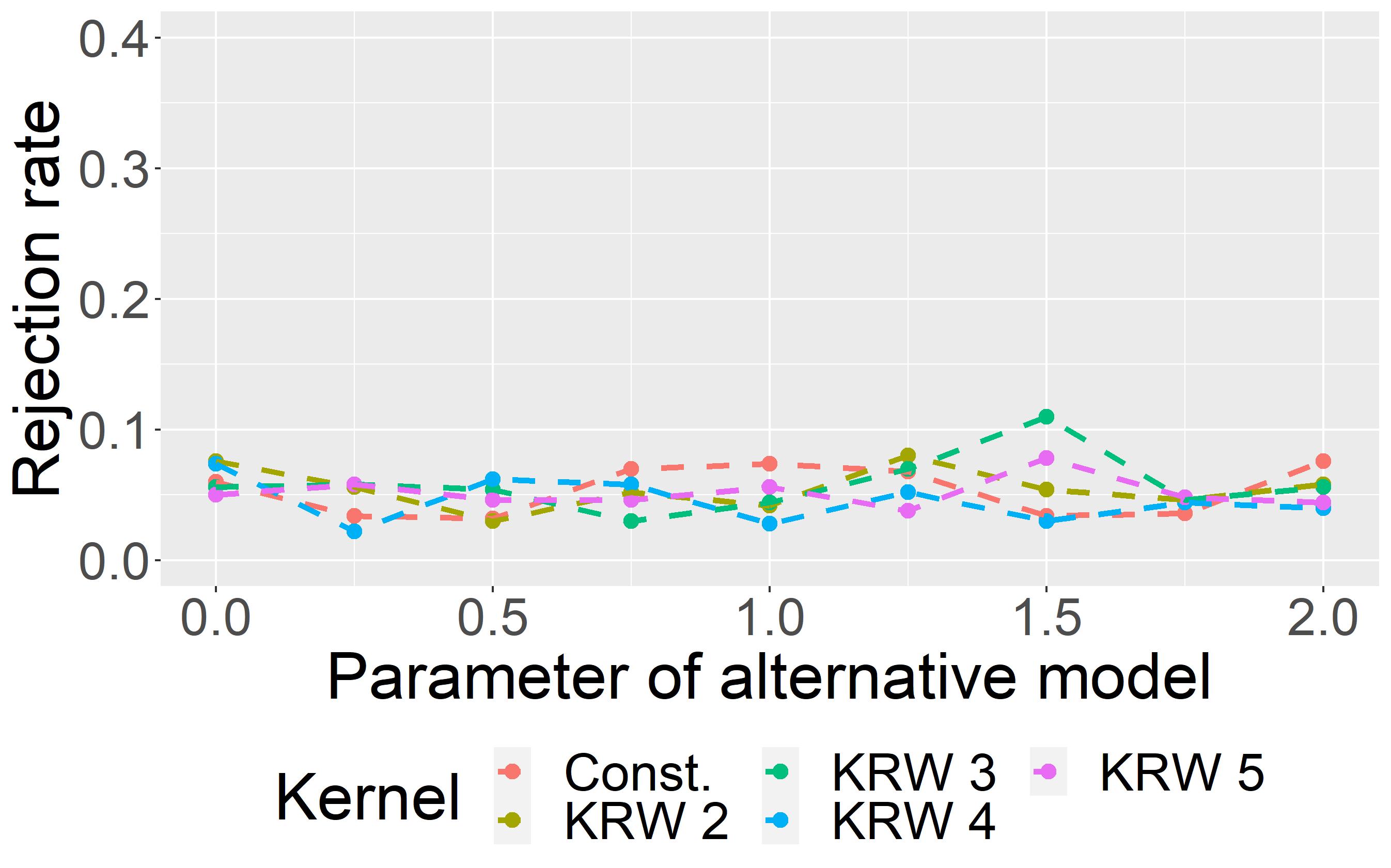}}
    \caption{
    AgraSSt for BA model with $m=2$; $t(x)$ being the bivariate degree vector.}
    \label{fig:ba-bideg}
\end{figure*}

{A Barabasi-Albert model generates scale-free networks using preferential attachment. The algorithm starts with a complete graph of $m$ vertices, where $m \in \mathbb{N}$ is chosen as a parameter of the model. In every step, one vertex is added and connected with $m$ edges to the network. The {version used here is from the R package \texttt{igraph}; the} probability of a vertex $v$ being chosen to connect to the new vertex depends on its current degree via
\begin{equation*}
p_v \propto deg(v)^{\alpha} + 1,
\end{equation*}
where $\alpha$ is a power parameter which governs the intensity of preference for high-degree vertices in the attachment step. {When $m\ge 2$ the degrees are updated after the first edge is added and before the second edge is added, and the second edge is then added according to the updated degrees.}  If $\alpha = 0$, the vertices to attach to are chosen uniformly at random, whereas $\alpha = 2$ leads to graphs which are almost starlike with one central vertex (or for $m>1$ multiple central vertices) and {most of the} remaining vertices only connected to the centre. For $\alpha > 0$ vertices with higher degree are more likely to connect to new vertices, leading to few vertices with unusually high degrees in comparison to other graph generators. Unlike in ERGMs or 
GRG models, a change of the parameter $\alpha$ does not result in a change of edge density. 

{We carry out tests of the form
$M0:\operatorname{BA} \alpha_{M0} = 1$ versus $M1:\operatorname{BA} \alpha_{M1}$, with $\alpha_{M1}$ ranging from $0$ to $2$. 
The rejection rates for $m=1$ are shown in \Cref{fig:ba1-density} and \Cref{fig:ba1-bideg}.
Furthermore,  the results of the  experiment for  $m=2$  are shown in \Cref{fig:ba-density} and \Cref{fig:ba-bideg}. 
We note that here a change in the parameter, $\alpha$, does 
not generally yield high rejection rates. This is not surprising for $t(x)$ the edge density, as edge density is not influenced by $\alpha$, but  for most kernels this also the case for $t(x)$ the bivariate degree vector. For $m=1$ notable exception is the Gaussian vertex-edge histogram kernel with small parameter $\sigma=0.1$. This kernel is tailored to assessing degree pairs and hence it is plausible that it picks up the power law distribution in the degrees. In this example the choice of kernel and of parameter can make a considerable difference. 
For $m=2$ also the GRW kernel with $\lambda=0.5$ and the WL kernel with level $3$ and $5$ pick up some signal. 
}

}

\subsection{Additional CELL experiments} 

\subsubsection{Synthetic data}
For our additional experiments we {first} choose a theoretical graph generator as null model $M0$. 
By the construction of CELL, the generator can only be trained on a single network and by repeating the training process, we reduce the risk of sampling an unrepresentative network from $M0$ and thus making all networks trained on this generator unrepresentative.

For all  sample{s} from ${M1}$, we perform a Monte Carlo test based on sampled AgraSSt with different statistics $t$. {In} the case of the E2S-model {which is a ERGM and hence gKSS is applicable, we} additionally {compare to}  sampled gKSS, to obtain average rejection rates for different graph kernels $k$. As null model, we test the E2S-model with parameters $\beta=(-2,0)$, the Geometric Random Graph model with radius $r = 0.3$ on a unit square without torus structure, and the Barabasi-Albert model with $m=1$ and power parameter $\alpha = 1$.

\paragraph{An E2ST experiment} 
Rejection rates for the E2S-model are displayed in Table \ref{E2S-CELL}. We observe that rejection rates for all statistics and kernels are around 5\%, which is expected if the CELL-simulated samples are not distinguishable from the original graph generator. The two-star coefficient of our null model is $\beta_{2S} = 0$, hence the only criterion of a graph affecting the probability distribution of the model is its edge density. {As the} 
CELL-simulated graphs have   the same edge density as the graph the generator was trained on,  we may expect the alternative model to get rejected in roughly 5\% of cases.  {We note that gKSS, which for this model includes edges and two-stars as sufficient statistics, has only slightly lower rejection rates on average than AgraSSt.} 

\begin{table}[H]
	\centering
	\resizebox{0.95\textwidth}{!}{%
	\begin{tabular}{|l|r|r|r|r|}
		\hline
		Kernel & gKSS & Avg. density & Bidegree statistic & Common neighbour statistic \\ 
		\hline
		Const. & 0.02 & 0.04 & 0.05 & 0.05 \\ 
		GVEH 0.1 & 0.04 & \textcolor{amber}{0.07} & \textcolor{amber}{0.09} & \textcolor{amber}{0.07} \\ 
		GVEH 1 & 0.03 & 0.04 & 0.05 & 0.05 \\ 
		GVEH 10 & 0.01 & 0.04 & 0.02 & \textcolor{amber}{0.06} \\ 
		GVEH 100 & 0.04 & 0.04 & 0.02 & \textcolor{amber}{0.06} \\ 
		SP & \textcolor{amber}{0.06} & 0.05 & 0.05 & 0.05 \\
		KRW 2 & 0.02 & 0.03 & 0.01 & 0.05 \\ 
		KRW 3 & 0.04 & 0.03 & 0.05 & 0.05 \\ 
		KRW 4 & 0.02 & 0.04 & 0.04 & 0.05 \\ 
		KRW 5 & 0.02 & 0.04 & 0.03 & 0.05 \\ 
		GRW 1e-5 & 0.05 & 0.03 & 0.03 & 0.05 \\
		GRW 1e-4 & 0.03 & 0.02 & 0.04 & \textcolor{amber}{0.06} \\
		GRW 1e-3 & 0.03 & 0.02 & 0.04 & \textcolor{amber}{0.06} \\
		GRW 1e-2 & 0.02 & 0.05 & 0.03 & 0.05 \\    
		GRW 5e-2 & 0.02 & 0.04 & 0.03 & 0.05 \\ 
		WL 1 & 0.03 & 0.04 & 0.03 & 0.05 \\ 
		WL 3 & 0.03 & 0.04 & \textcolor{amber}{0.06} & 0.03 \\ 
		WL 5 & 0.04 & 0.04 & 0.05 & 0.04 \\ 
		GLET 3 & 0.02 & 0.02 & 0.03 & \textcolor{amber}{0.06} \\ 
		CONGLET 3 & 0.02 & 0.03 & 0.05 & 0.05 \\ 
		CONGLET 4 & 0.04 & 0.02 & 0.02 & 0.04 \\
		\hline
	\end{tabular}
	}
	\vspace{0.3cm}
	\caption[Rejection rates for CELL-simulated Edge-Two star graph samples]{Rejection rates for CELL-simulated Edge-Two star graph samples with parameters $\beta = (-2,0)$ using the gKSS and AgraSSt testing procedure with different summary statistics. {Rejection rates {over 5\% are marked in amber, and rejection rates} of at least 10\% would have been marked in red}.}
	\label{E2S-CELL}	
\end{table}

\paragraph{The geometric random graph experiment} 
Table \ref{GRG-CELL} shows the rejection rates in {the}  Geometric Random Graph experiment {from \Cref{sec:exp}, now using CELL}. The rates for AgraSSt based on the average density are all roughly 5\%. While this statistic is effective at distinguishing between different radius parameters in the experiment with the Geometric Random Graph model in \Cref{sec:exp}, this {may just reflect that} 
a  change in radius also changes the average edge density. 
When using the bidegree statistics the average rejection rate is slightly higher than 5\% and some kernels achieve a rejection rate of over 10\%. {Only some of the Weisfeiler-Lehman kernels achieve at most 5\% rejection rate, under the bidegree statistic.} 

\begin{figure}[ht!]
	\begin{center}
		\subfigure[]{%
			\label{fig:CELL-GRG1}
			\includegraphics[width=0.24\textwidth]{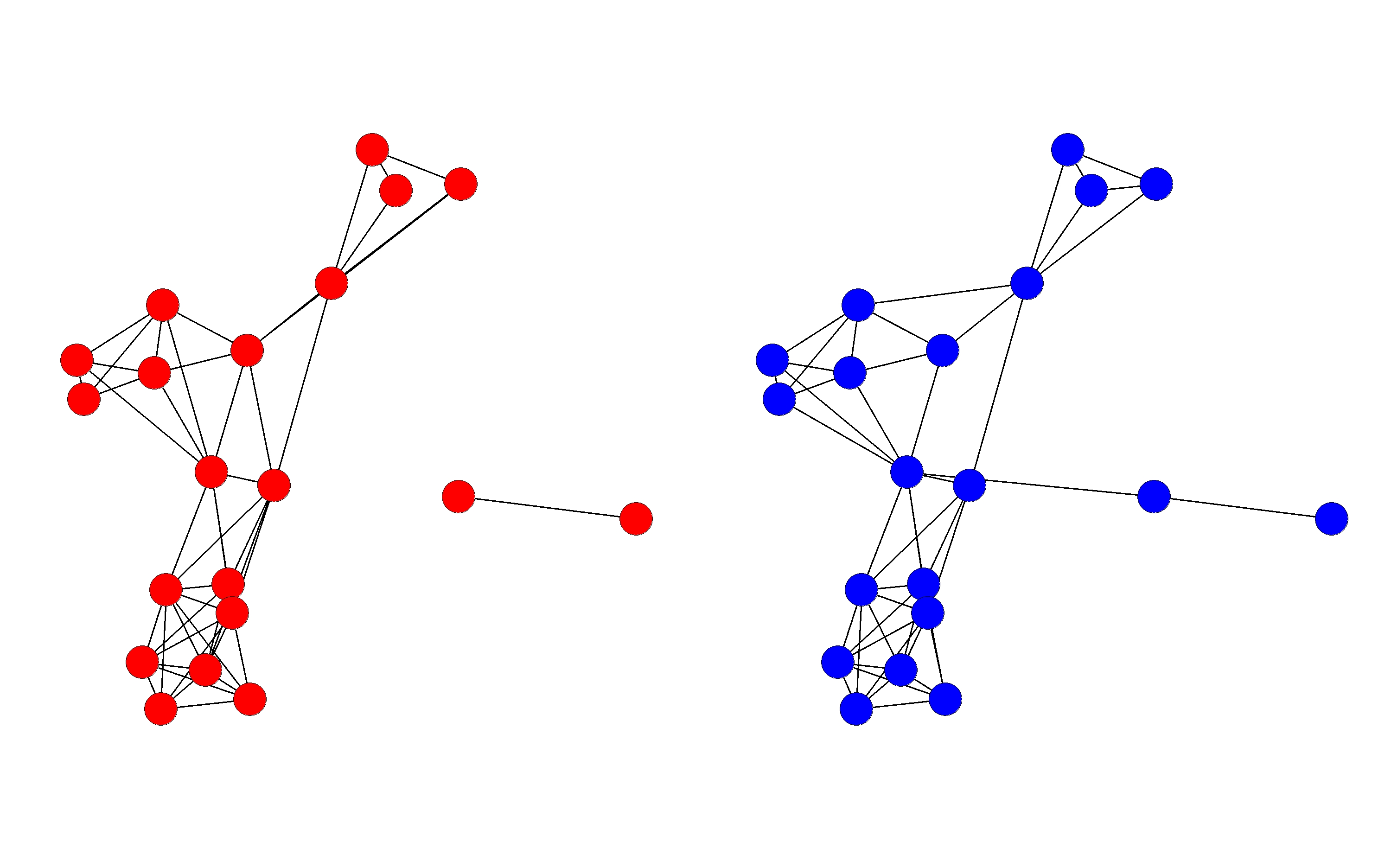}
		}\subfigure[]{%
			\label{fig:CELL-GRG2}
			\includegraphics[width=0.24\textwidth]{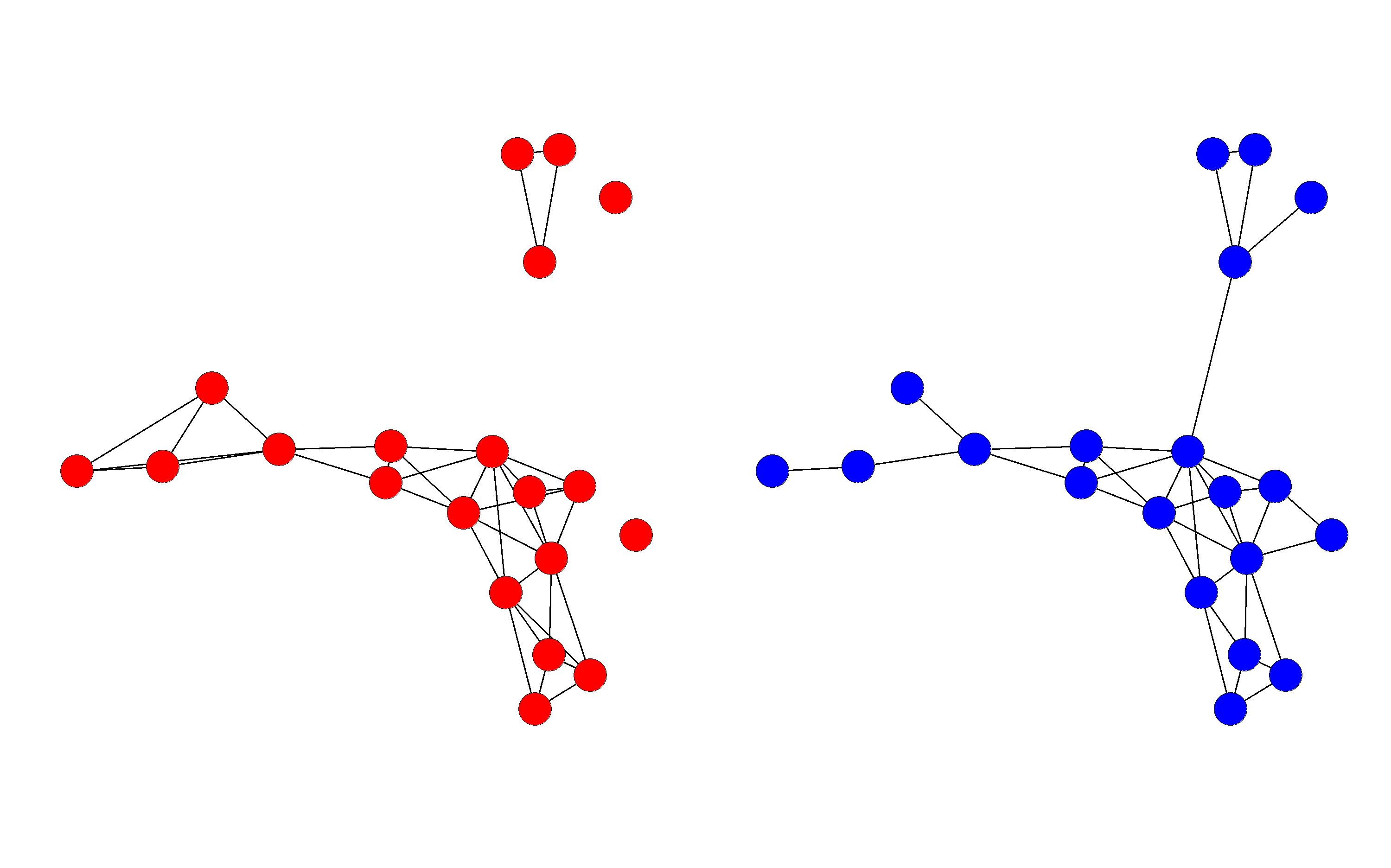}
		}\subfigure[]{%
			\label{fig:CELL-GRG4}
			\includegraphics[width=0.24\textwidth]{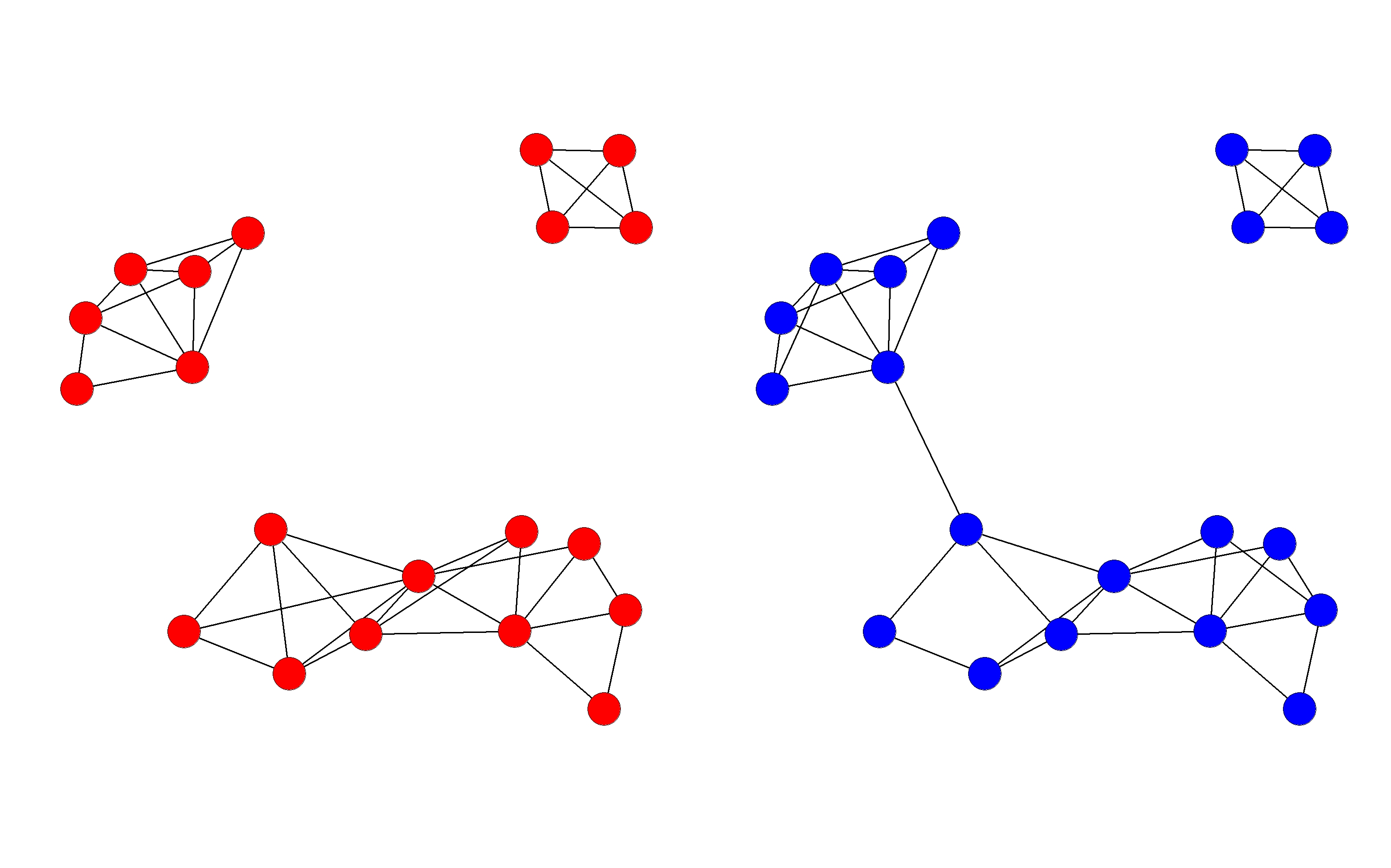}
		}\subfigure[]{%
			\label{fig:CELL-GRG5}
			\includegraphics[width=0.24\textwidth]{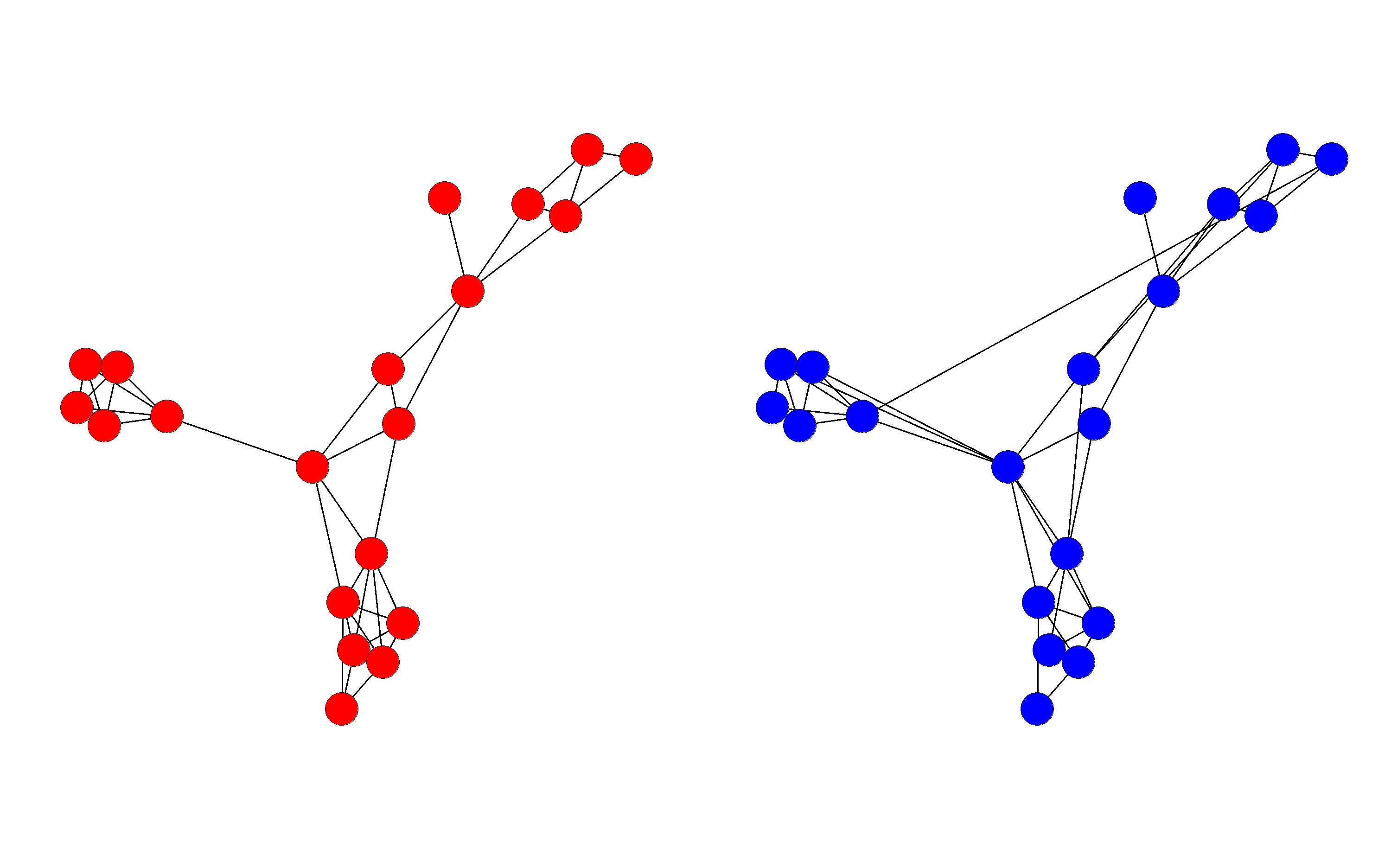}
		}%
	\end{center}
	\caption[Rejected CELL-samples for the Geometric Random Graph model]{%
		CELL-samples for the Geometric Random Graph model: original network is displayed in red, and the CELL sample is displayed in blue.
	}%
	\label{fig:CELL-GRG}
\end{figure}

The common neighbour statistic achieves the highest rejection rates as all but one kernel reject in 10\% or more of cases. The maximal rejection rate of 16\% is achieved by the Gaussian vertex-edge histogram Kernel with bandwidth $\sigma = 0.1$, the Shortest Path kernel and the Geometric Random Walk kernel with weight $\lambda = 0.01$. These results align with our findings in \Cref{sec:exp} where the common neighbour statistic achieved higher rejection rates than the bidegree statistic and the Gaussian vertex-edge Histogram kernel achieved the best results. In analysing the graphs which are rejected by the Shortest Path kernel, we can furthermore see that CELL has a tendency to connect small disconnected components to the rest of the graph and create additional paths between components which are only attached through one edge (see Figure \ref{fig:CELL-GRG}). So it appears that CELL may struggle with generating networks which are constituted by a few disconnected or sparsely connected components. However, the case of many disconnected components, as generated by the sparse E2S-model, seems unproblematic. Altogether however, rejection rates remain fairly low for all kernels, indicating that CELL produces fairly accurate samples despite its flaws.

\begin{table}[t!]
	\centering
	\begin{tabular}{|l|r|r|r|}
		\hline
		Kernel & Avg. density & Bidegree statistic & Common neighbour statistic \\ 
		\hline
		Const. & 0.03 & \textcolor{amber}{0.09} & \textcolor{red}{0.12} \\ 
		GVEH 0.1 & \textcolor{amber}{0.08} & \textcolor{amber}{0.08} & \textcolor{red}{0.16} \\ 
		GVEH 1 & \textcolor{amber}{0.09} & \textcolor{amber}{0.08} & \textcolor{red}{0.11} \\ 
		GVEH 10 & 0.04 & \textcolor{red}{0.13} & \textcolor{red}{0.12} \\ 
		GVEH 100 & 0.03 & \textcolor{amber}{0.09} & \textcolor{amber}{0.07} \\ 
		SP & \textcolor{amber}{0.06} & \textcolor{amber}{0.09} & \textcolor{red}{0.16} \\ 
		KRW 2 & 0.03 & \textcolor{red}{0.10} & \textcolor{red}{0.12} \\ 
		KRW 3 & \textcolor{amber}{0.07} & \textcolor{amber}{0.09} & \textcolor{red}{0.15} \\ 
		KRW 4 & \textcolor{amber}{0.08} & \textcolor{red}{0.10} & \textcolor{red}{0.12} \\ 
		KRW 5 & 0.04 & \textcolor{amber}{0.06} & \textcolor{red}{0.15} \\ 
		GRW 1e-5 & 0.04 & \textcolor{amber}{0.07} & \textcolor{red}{0.10} \\
		GRW 1e-4 & 0.02 & \textcolor{amber}{0.08} & \textcolor{red}{0.15} \\
		GRW 1e-3 & 0.05 & \textcolor{red}{0.10} & \textcolor{red}{0.16} \\
		GRW 1e-2 & \textcolor{amber}{0.06} & \textcolor{amber}{0.07} & \textcolor{red}{0.12} \\ 
		GRW 5e-2 & 0.03 & \textcolor{amber}{0.08} & \textcolor{amber}{0.09} \\ 
		WL 1 & \textcolor{amber}{0.06} & \textcolor{amber}{0.08} & \textcolor{red}{0.14} \\ 
		WL 3 & 0.04 & 0.04 & \textcolor{red}{0.10} \\ 
		WL 5 & 0.04 & 0.05 & \textcolor{red}{0.15} \\ 
		GLET 3 & 0.04 & \textcolor{amber}{0.08} & \textcolor{red}{0.14} \\ 
		CONGLET 3 & 0.04 & \textcolor{amber}{0.08} & \textcolor{red}{0.15} \\ 
		CONGLET 4 & 0.03 & \textcolor{red}{0.10} & \textcolor{red}{0.10} \\ 
		\hline
	\end{tabular}
	\vspace{0.3cm}
	\caption[Rejection rates for CELL-simulated Geometric Random Graph samples]{Rejection rates for CELL-simulated Geometric Random Graph samples with parameters $r = 0.3$ on a unit-square without torus structure using the AgraSSt testing procedure with different summary statistics. {Rejection rates {over 5\% are marked in amber, and rejection rates} of at least 10\% are marked in red}.} 
	\label{GRG-CELL}
\end{table}

\paragraph{The Barabasi-Albert experiment} 

Rejection rates for the Barabasi-Albert model, {now using CELL,} with parameters $m=1$ and $\alpha = 1$ are presented in \Cref{BA-CELL}. We observe that while their average lies slightly above 5\%, samples from CELL would be still accepted {at the 10\% level} in the majority of cases. AgraSSt with the bidegree statistic achieves the largest rejection rates, which agrees with our findings in \Cref{app:experiment-ba}. 
Similarly, the Gaussian vertex-edge histogram kernel achieves the highest rejection rate with a maximum of 16\% with bandwidth $\sigma = 1$. The Weisfeiler-Lehman kernel also performs well, attaining a rejection rate of 13\% for any level parameter and a maximum of 16\% for $h=3$. While the Weisfeiler-Lehman kernel was not effective in distinguishing between different power parameters $\alpha$ for $m=1$, it did boost the rejection rates of sampled AgraSSt with the bidegree statistic for $m=2$ 
Out of the 100 simulated samples, 58 graphs contain multiple disconnected subgraphs or cycles, which should make them clearly distinguishable from graphs created by the Barabasi-Albert model with $m=1$, whereas only 42 are connected and contain no cycle. Most kernels do a good job at accepting connected graphs, above all the connected graphlet kernel of size 4. Out of the 10 graphs it rejects, only one is connected, and the other 41 connected graphs are accepted by the testing procedure. However, all kernels still accept a large number of disconnected networks.

\begin{table}[t!]
	\centering
	\begin{tabular}{|l|r|r|r|}
		\hline
		Kernel & Avg. density & Bidegree statistic & Common neighbour statistic \\ 
		\hline
		Const. & \textcolor{amber}{0.07} & 0.05 & \textcolor{red}{0.10} \\ 
		GVEH 0.1 & \textcolor{amber}{0.08} & \textcolor{red}{0.13} & 0.05 \\ 
		GVEH 1 & \textcolor{amber}{0.06} & \textcolor{red}{0.16} & 0.05 \\ 
		GVEH 10 & 0.04 & \textcolor{amber}{0.09} & \textcolor{amber}{0.09} \\ 
		GVEH 100 & \textcolor{amber}{0.07} & \textcolor{amber}{0.08} & \textcolor{amber}{0.08} \\ 
		SP & 0.05 & \textcolor{amber}{0.07} & \textcolor{amber}{0.08} \\ 
		KRW 2 & \textcolor{amber}{0.07} & \textcolor{amber}{0.08} & \textcolor{amber}{0.07} \\ 
		KRW 3 & \textcolor{red}{0.10} & \textcolor{red}{0.10} & \textcolor{red}{0.12} \\ 
		KRW 4 & \textcolor{amber}{0.08} & \textcolor{amber}{0.06} & \textcolor{amber}{0.09} \\ 
		KRW 5 & \textcolor{amber}{0.07} & \textcolor{amber}{0.08} & \textcolor{red}{0.10} \\
		GRW 1e-5 & \textcolor{amber}{0.09} & \textcolor{amber}{0.09} & \textcolor{red}{0.12} \\  
		GRW 1e-4 & \textcolor{amber}{0.07} & \textcolor{amber}{0.07} & \textcolor{amber}{0.09} \\ 
		GRW 1e-3 & 0.04 & \textcolor{amber}{0.08} & \textcolor{amber}{0.09} \\ 
		GRW 1e-2 & 0.04 & \textcolor{red}{0.12} & \textcolor{amber}{0.09} \\ 
		GRW 5e-2 & \textcolor{amber}{0.08} & \textcolor{amber}{0.08} & \textcolor{amber}{0.08} \\ 
		WL 1 & \textcolor{amber}{0.09} & \textcolor{red}{0.13} & 0.04 \\ 
		WL 3 & \textcolor{amber}{0.08} & \textcolor{red}{0.16} & 0.05 \\ 
		WL 5 & 0.07 & \textcolor{red}{0.13} & 0.08 \\ 
		GLET 3 & \textcolor{red}{0.10} & \textcolor{red}{0.11} & 0.08 \\
		CONGLET 3 & \textcolor{amber}{0.07} & \textcolor{amber}{0.08} & \textcolor{amber}{0.09} \\ 
		CONGLET 4 & \textcolor{red}{0.10} & \textcolor{red}{0.10} & \textcolor{amber}{0.07} \\
		\hline
	\end{tabular}
	\vspace{0.3cm}
	\caption[Rejection rates for CELL-simulated Barabasi-Albert graph samples]{Rejection rates for CELL-simulated Barabasi-Albert graph samples with parameters $m=1$ and $\alpha = 1$ using the AgraSSt testing procedure with different summary statistics. Rejection rates {over 5\% are marked in amber, and} of at least 10\% are marked in red.}
	\label{BA-CELL}
\end{table}

{
\subsubsection{Details on the Karate club network and the CELL results} 

Zachary's Karate Club network \cite{zachary1977information} contains 34 vertices, representing the members of a university sport society before its separation into two new groups due to a conflict between the instructor and the administrator. An edge in the network symbolizes consistent interaction between members outside of karate classes. Using the structural information about friendships in the club, Zachary found a method to cluster the vertices which for all but one member agreed with the side they would end up after the split. The network became a widespread example of community structures after its use by  \cite{girvan2002community}.
} 

{
We perform a Monte Carlo test, in which we compare sampled AgraSSt with sample size $B=200$ of the original network to $n_{M1} = 100$ simulations from CELL and reject at the 5\%-level. This procedure is repeated 100 times to obtain average rejection rates. The complete results  are displayed in Table \ref{zachary-CELL}. Most rejection rates remain at around 5\%, but when using the bidegree statistic both the Gaussian Vertex-Edge Histogram kernel with bandwidth $\sigma = 0.1$ and the Geometric Random Walk kernel with $\lambda = 0.05$\footnote{We may choose $\lambda = 0.05$ as the original and simulated networks have no vertex with degree 20 or above, so the infinite sum in the Geometric Random Walk kernel converges. We could allow for larger $\lambda$, but chose to only consider values up to 0.05 to keep the considered hyperparameters consistent throughout.} achieve rejection rates above 15\%.

{We recall that} in the experiments on the Barabasi-Albert model 
these two kernels were able to detect differences in graph structure in certain cases, so {there is some indication that} these  results {may} extend to real-life applications.
}

\begin{figure}[t!]
	\begin{center}
		\subfigure[Accepted batch]{%
			\label{fig:CELL-zachary-accept}
			\includegraphics[width=0.8\textwidth]{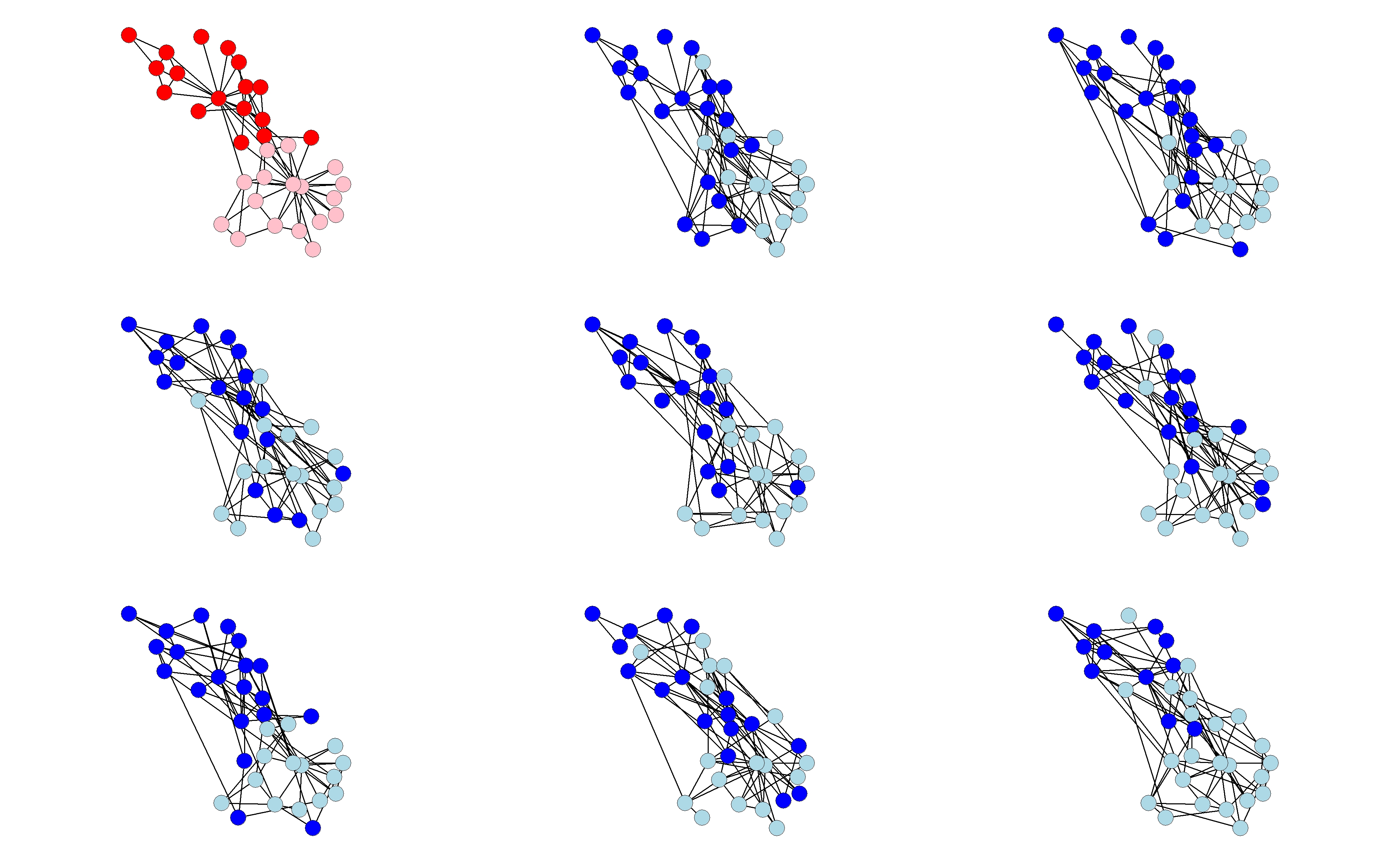}
		}
		
		\subfigure[Rejected batch]{%
			\label{fig:CELL-zachary-reject}
			\includegraphics[width=0.8\textwidth]{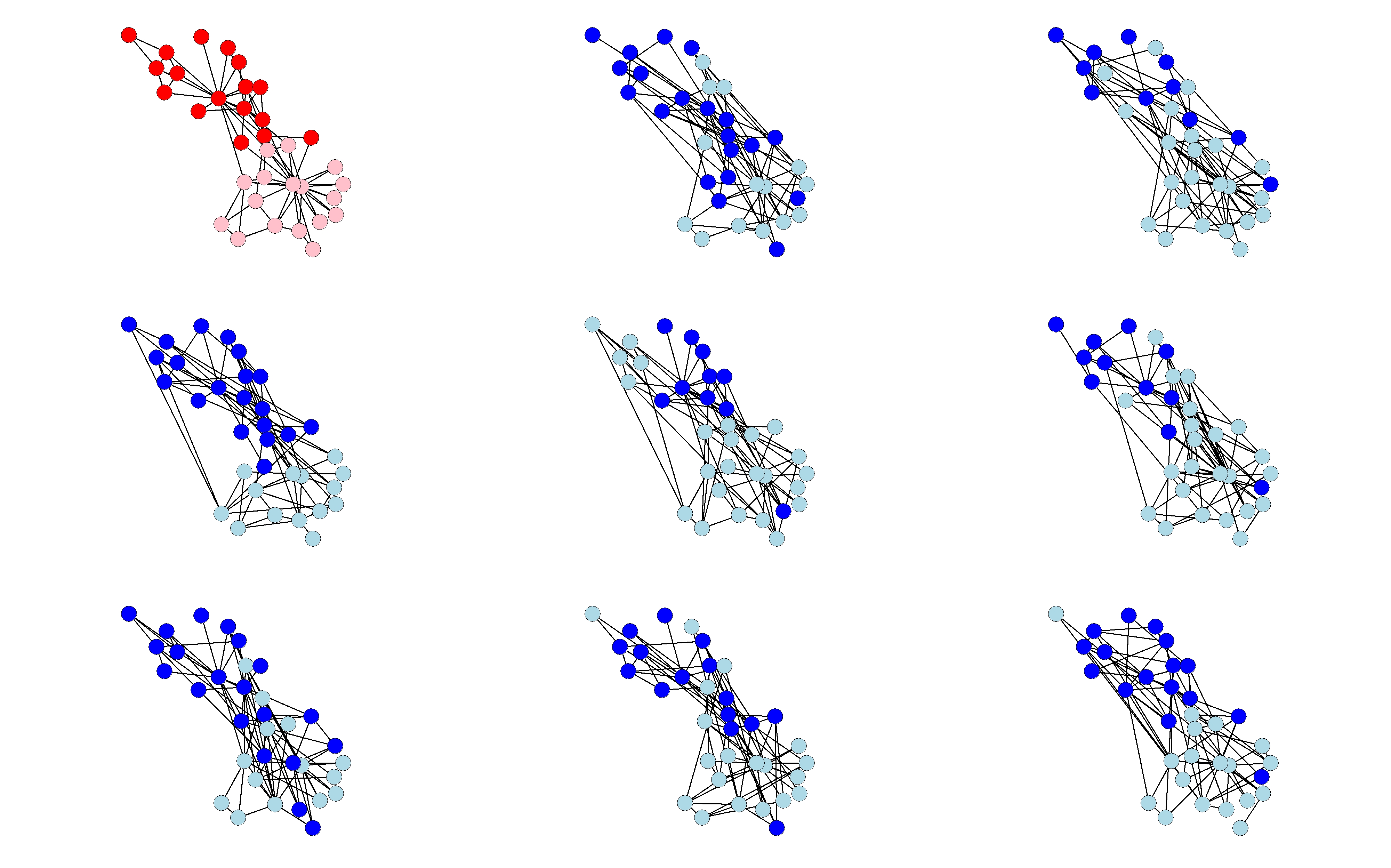}
		}
	\end{center}
	\caption[Rejected and accepted CELL-samples trained on Zarachy's Karate Club network]{%
		Rejected and accepted CELL-samples trained on Zarachy's Karate Club network. The original graph is displayed in red,  simulated graphs are displayed in blue. Different shadings indicate  the cluster membership. The layout is kept fixed for all graphs.
		We  observe no significant difference between the accepted and rejected sample in how well cluster membership in the simulated samples corresponds to cluster membership in the original graph.
	}%
	\label{fig:CELL-zachary}
\end{figure}

However, the {CELL} generator {does not always}  produce graphs which portray the same structures of group membership as the original network and AgraSSt {can fail}  to detect this shortcoming. To illustrate this, we separate the vertices in the training and simulated graphs into two clusters using a greedy algorithm {from} \citet{clauset2004cluster}. The rate of coincidence between the cluster assignment in the original graph and the simulations varies between 64.3\% to 70.8\%. {The AgraSSt test decision however} 
seems to be largely independent of how well the community structure is reproduced. Figure \ref{fig:CELL-zachary} displays two batches of simulated graphs, one accepted and one rejected by the Gaussian Vertex-Edge Histogram kernel. There is no discernible difference in cluster assignment in the two batches;  
{the} group allocation matches the original graph for 66.4\% of vertices in the accepted batch, whereas the rejected batch achieves 70.0\%. Therefore, the current implementation with the chosen graph kernels may have trouble detecting differences in community structures if they are not represented in other statistics such as the degree distribution. One possible solution is assigning each vertex their group membership in the original graph as an attribute, which gives the graph kernels explicit information to detect discrepancies.
On another note, we observe that the Gaussian Vertex-Edge Histogram kernel and the Geometric Random Walk kernel reject almost entirely different batches. 
{As mentioned in \Cref{sec:conclusion}, this finding} opens the possibility for using an ensemble of kernels which may achieve higher power, {as for example in \cite{schrab2022ksd}}.

\begin{table}[t!]
	\centering
	\begin{tabular}{|l|r|r|r|}
		\hline
		Kernel & Avg. density & Bidegree statistic & Common neighbour statistic \\ 
		\hline
		Const. & 0.03 & 0.04 & 0.05 \\
		GVEH 0.1 & 0.04 & \textcolor{red}{0.17} & 0.03 \\ 
		GVEH 1 & \textcolor{amber}{0.07} & \textcolor{red}{0.11} & 0.04 \\ 
		GVEH 10 & 0.05 & \textcolor{amber}{0.06} & 0.04 \\ 
		GVEH 100 & \textcolor{red}{0.12} & \textcolor{amber}{0.06} & \textcolor{red}{0.10} \\
		SP & \textcolor{amber}{0.08} & 0.01 & \textcolor{amber}{0.08} \\ 
		KRW 2 & 0.04 & 0.05 & \textcolor{amber}{0.06} \\ 
		KRW 3 & 0.03 & 0.04 & 0.02 \\ 
		KRW 4 & 0.02 & 0.04 & 0.01 \\ 
		KRW 5 & \textcolor{red}{0.10} & 0.04 & \textcolor{amber}{0.09} \\ 
		GRW 1e-5 & 0.04 & 0.04 & \textcolor{amber}{0.06} \\ 
		GRW 1e-4 & 0.05 & 0.04 & 0.01 \\ 
		GRW 1e-3 & \textcolor{amber}{0.07} & \textcolor{amber}{0.06} & 0.05 \\ 
		GRW 1e-2 & \textcolor{amber}{0.06} & 0.05 & \textcolor{amber}{0.09} \\ 
		GRW 5e-2 & 0.03 & \textcolor{red}{0.16} & 0.03 \\ 
		WL 1 & 0.05 & \textcolor{amber}{0.07} & 0.04 \\ 
		WL 3 & 0.02 & 0.01 & 0.03 \\ 
		WL 5 & 0.03 & 0.05 & 0.03 \\ 
		GLET 3 & \textcolor{amber}{0.08} & 0.04 & 0.04 \\  
		CONGLET 3 & 0.03 & 0.05 & 0.04 \\ 
		CONGLET 4 & 0.02 & 0.02 & \textcolor{amber}{0.07} \\ 
		\hline
	\end{tabular}   
    \vspace{0.3cm}
	\caption[Rejection rates for CELL-simulated Zachary's Karate Club samples]{Rejection rates for CELL-simulated Zachary's Karate Club samples using the AgraSSt testing procedure with different summary statistics. Rejection rates over 5\% are marked in amber and those {of at least}  10\% are marked in red.}
	\label{zachary-CELL}
\end{table}

\section{Runtime {considerations}}

\subsection{Runtime experiments} 
{Here w}e present the runtimes for calculating sampled {gKSS} with sample size $B=200$ using the kernel implementations by the R package \texttt{graphkernel}. We use an Edge-2Star model with parameters $\beta = (-2,0)$ (sparse regime, average edge density 11.8\%) and $\beta = (1,0)$ (dense regime, average edge density 73.2\%) for $n=20$ and $n=40$ vertices. {For a given graph, we run each kernel ten times on the graph and pick the median runtime to largely remove randomness in the runtime due to the momentary performance of the machine from the runtime analysis. For each of the four set-ups, we repeat this procedure for 100 different graphs, obtaining 100 median runtimes per set-up. We report their minimum, average and maximum for every kernel. We consider different hyperparameters for the kernels, {to assess whether} the hyperparameter affects the computational complexity of the algorithm.}

\begin{table}[t!]
	\centering
	\resizebox{\textwidth}{!}{%
		\begin{tabular}{|l|rrr|rrr|}
			\hline
			& \multicolumn{3}{c|}{Runtime (ms) for $n=20$, sparse} & \multicolumn{3}{c|}{Runtime (ms) for $n=40$, sparse}\\
			\hline
			Kernel & Minimum & Average & Maximum & Minimum & Average & Maximum \\ 
			\hline
			Const. & 0.47 & 0.50 & 1.03 & 0.51 & 0.57 & 1.10 \\ 
			WL 1 & 199.51 & 206.42 & 214.23 & 219.38 & 224.10 & 233.86 \\ 
			WL 3 & 201.54 & 210.59 & 218.99 & 229.06 & 235.09 & 242.58 \\ 
			WL 5 & 208.74 & 218.26 & 225.49 & 245.64 & 255.18 & 267.64 \\ 
			GLET 3 & 192.08 & 202.50 & 214.43 & 311.96 & 326.72 & 345.26 \\ 
			GVEH & 318.32 & 331.63 & 342.52 & 428.01 & 448.00 & 470.37 \\ 
			CONGLET 3 & 385.88 & 400.35 & 417.83 & 512.63 & 539.49 & 560.63 \\ 
			CONGLET 4 & 379.00 & 407.84 & 438.10 & 620.05 & 754.20 & 968.26 \\ 
			KRW 3 & 432.39 & 682.93 & 1,010.08 & 3,638.44 & 5,460.50 & 7,713.08 \\ 
			KRW 5 & 432.76 & 682.45 & 1,008.46 & 3,616.95 & 5,463.80 & 7,715.39 \\ 
			GRW & 427.29 & 678.56 & 1,002.74 & 3,590.22 & 5,473.98 & 7,725.11 \\ 
			SP & 655.11 & 940.04 & 1,278.72 & 4,071.83 & 5,984.19 & 8,280.77 \\ 
			\hline
		\end{tabular}
	}
    \vspace{0.2pt}
	\caption[Runtimes of graph kernels in sparse graphs]{\label{tab:runtime-sparse}Runtime of graph kernels with different hyperparameters on sparse graphs of size 20 and 40 in milliseconds. The sparse graphs are simulated from an E2S-model with parameters $\beta = (-2,0)$.}
	\label{runtime-sparse}
\end{table}

\begin{table}[t!]
	\centering
	\resizebox{\textwidth}{!}{%
		\begin{tabular}{|l|rrr|rrr|}
			\hline
			& \multicolumn{3}{c|}{Runtime (ms) for $n=20$, dense} & \multicolumn{3}{c|}{Runtime (ms) for $n=40$, dense}\\
			\hline
			Kernel & Minimum  & Average& Maximum & Minimum & Average & Maximum \\ 
			\hline
			Const. & 0.47 & 0.53 & 1.04 & 0.51 & 0.54 & 1.12 \\ 
			WL 1 & 210.82 & 217.08 & 224.71 & 245.06 & 253.82 & 415.35 \\ 
			WL 3 & 215.61 & 227.18 & 235.57 & 268.51 & 290.70 & 460.15 \\ 
			WL 5 & 235.62 & 243.09 & 249.93 & 319.81 & 351.17 & 376.40 \\ 
			GLET 3 & 212.13 & 219.29 & 230.56 & 406.23 & 447.68 & 482.55 \\ 
			GVEH & 366.71 & 384.54 & 405.73 & 581.44 & 640.82 & 890.56 \\ 
			CONGLET 3 & 425.63 & 444.61 & 466.28 & 644.86 & 796.21 & 928.90 \\ 
			CONGLET 4 & 2,145.85 & 2,439.85 & 2,838.56 & 16,114.99 & 41,525.31 & 54,248.50 \\ 
			KRW 3 & 9,241.91 & 10,184.59 & 11,846.73 & 72,359.72 & 135,137.37 & 165,997.28 \\ 
			KRW 5 & 9,225.76 & 10,188.56 & 11,801.09 & 72,578.23 & 135,177.07 & 165,510.63 \\ 
			GRW & 9,206.17 & 10,222.37 & 11,985.30 & 72,276.88 & 135,096.51 & 165,456.58 \\ 
			SP & 9,767.99 & 10,706.33 & 12,311.66 & 74,118.71 & 136,571.69 & 166,649.77 \\ 
			\hline
		\end{tabular}
	}
    \vspace{0.2pt}
	\caption[Runtimes of graph kernels in dense graphs]{\label{tab:runtime-dense}Runtime of graph kernels with different hyperparameters on dense graphs of size 20 and 40 in milliseconds. The dense graphs are simulated from an E2S-model with parameters $\beta = (1,0)$.}
	\label{runtime-dense}
\end{table}

{The results are shown in \Cref{runtime-sparse}, for sparse networks, and \Cref{runtime-dense}, for dense networks.} 
{The constant kernel is the quickest to evaluate by {two orders of magnitude}
compared to the runner-up. The fastest non-trivial choice is the Weisfeiler-Lehman kernel, with which {gKSS} on average takes about 0.2 to 0.35 seconds to calculate. The runtime is very consistent in every set-up with only little deviation and only slightly increases with increased level or density.
Similarly, {gKSS} using the Gaussian vertex-edge histogram kernel takes between around 0.33 seconds to compute on the network of 20 vertices and 0.45 seconds on the network of 40 vertices, with no large increase in runtime on the denser networks.

For the graphlet kernel on three vertices, it takes roughly 0.2 seconds to calculate {gKSS} on the network on 20 vertices and 0.33 seconds on the network of 40 vertices. As the number of triplets of vertices which need to be checked for calculating the kernel is independent of the edge structure of the graph, there is little difference in runtime for the sparse or dense network. This is very different for the connected graphlet {kernel} as an action such as increasing the count of a certain graphlet is only needed if the vertex set that is currently examined by the algorithm is connected. Using the connected graphlet kernel takes longer than the graphlet kernel as additional checks for {connectivity}  
are needed. In the sparse regime, runtimes of the kernel on graphlets of size 3 or 4 are comparable, with a runtime of 0.4 seconds for both on 20 vertices and a runtime of 0.54 and 0.75 seconds on 40 vertices. However, a large disparity emerges in the dense regime: Whereas, for graphlets of size 3, the kernel takes 0.44 seconds on the smaller and 0.8 seconds on the bigger network, for graphlets of size 4, runtimes increase to 2.4 seconds on 20 vertices and even 41.5 seconds on 40 vertices.\\
Both the $k$-Random Walk kernels and the Geometric Random Walk kernel have a similar runtime, irrespective of the chosen hyperparameters. While their runtime is still competitive  on the smaller and sparser graphs, their runtime increases by an order of magnitude when doubling the number of vertices or moving from the sparse to the dense regime.
The slowest of the kernels is the shortest path kernel, though its runtime is largely comparable to the random walk kernels. In the sparse case it runs on average for 0.94 seconds on the smaller and 5.9 seconds on the larger network. In the dense case, however, runtime increases to 10.7 seconds for the smaller and to more than 2 minutes in the larger network. This renders the kernel in its current implementation unserviceable in practice. The reasons for this are twofold: Firstly, unlike the other kernels, its code is written in Python and not C++, thus making the implementation slower irrespective of the used algorithm. Secondly, the authors use a basic approach for calculation of the shortest path graph by running Dijkstra's algorithm for every vertex. This means that even with optimal data storage implementation, the runtime complexity of the algorithm for a graph on $n$ vertices and $m$ edge is $O(n^2\log(n) + nm)$ steps \cite{cormen2022introduction}. Generally, kernels considering paths in the graph have the longest runtimes and their comparative disadvantage becomes worse the larger and denser the network becomes. Furthermore, unlike the other kernels, their runtime varies greatly even in the same regime, so calculation times may strongly deviate from the average.}

 {As kernel implementation may have a considerable effect on the runtime, next we detail a computationally efficient implementation of GRW kernels}. 

\subsection{Efficient computation for GRW kernels}\label{app:runtimeidea}
{The} GRW kernel  {with parameter $\lambda$ for networks $x,x'$} {is}  
$k_{GRW}^{\lambda}(x, x' ) =  \sum_{i,j=1}^{ |V_{\otimes}|} \left[(I - \lambda A_{\otimes})^{-1}\right]_{i, j} = \mathbf{1}_n^T (I_n - \lambda A_{\otimes})^{-1} \mathbf{1}_n.
$
\footnote{details can be found in \Cref{app:graph_kernel}}. {This expression} 
involves inverting a $n\times n$ matrix, 
at 
cost $O(n^3)$. {Due to the special form of KSS the following theorem shows that $O(n)$ computation cost suffices}. 
%

\begin{Theorem}\label{th:efficient}
Let $B$ be a symmetric invertible matrix and $C = B^{-1}$. {Let} $C_i$ {denote} the $i$-th column and $c_{ij}$ 
the $(i,j)$-th entry of $C$; $e_i$ is the $i$-th coordinate vector. {Let} 
$\mu \in \R$ {satisfy}  $1+\mu c_{ij}\neq 0$ and $(1+\mu c_{ij})^2 - \mu^2 c_{ii} c_{jj}\neq 0$. {T}hen $M=(B + \mu(e_i e_j^T + e_j e_i^T ))$ is invertible {and} 
\begin{equation}\label{lemma-inverse-equation}
M^{-1} =  C - \frac{\mu(1 + \mu c_{ij})}{
(1 + \mu c_{ij} )^2 - \mu^2 c_{ii}c_{jj}}
(
C_i C^T_j + C_j C^T_i - \frac{\mu c_{jj}}{
1 + \mu c_{ij}}C_iC^T_i - \frac{\mu c_{ii}}{
1 + \mu c_{ij}}C_jC^T_j ).
\end{equation}
\end{Theorem}
{{For} $A_{\otimes} = A - (e_i e_j^T + e_j e_i^T)$ and 
$B=I_n - \lambda A$, taking  $\mu=\lambda$  yields a 
fast rank 1 computation of $B_{\otimes}=I_n - \lambda A_{\otimes}$ for GRW kernels.} 

To prove 
\Cref{th:efficient},
we apply the 
Sherman-Morrison formula  \citep{sherman1950adjustment} (as a special case of Woodbury matrix identity); we repeat it here for convenience. 

\begin{Proposition}[Sherman-Morrison]\label{sherman-morrison} Let $A\in\R^{n \times n}$ be an invertible matrix and let $u, v \in \R^n$
be column vectors. Then the matrix $A + uv^T $ is invertible if and only if $\mathbb{1} + v^T A ^{-1} u \neq 0$, and in
this case
\begin{equation}\label{eq:sherman-morrison}
    (A + uv^T)^{-1} = A^{-1} - \frac{1}{1+v^T A^{-1}u} A^{-1} uv^T A^{-1}.
\end{equation}
\end{Proposition}

\paragraph{Proof of \Cref{th:efficient}}
\begin{proof}
	The statement follows from applying \Cref{sherman-morrison} twice. First, we use the formula with $A=B$, $u= \mu e_i$, $v=e_j$ and note that as $1 + \mu c_{ij} \neq 0$, we may apply the proposition. Then by the symmetry of the inverse matrix $C$
	\begin{align}
		\nonumber (B+\mu e_i e_j^T)^{-1} &= B^{-1} - \frac{1}{1+ e_j^TB^{-1}\mu e_i} B^{-1} \mu e_i e_j^T B^{-1}\\& \label{first-eq-inversion} = C - \frac{\mu}{1+\mu c_{ij}} (C e_i) (e_j^TC)= C - \frac{\mu}{1+\mu c_{ij}} C_i C_j^T.
	\end{align}
	Applying the theorem again with $A=B+\mu e_i e_j^T$, $u=\mu e_j$, $v=e_i$ and assuming that $(1+\mu c_{ij})^2-\mu^2 c_{ii} c_{jj} \neq 0$, we can calculate the inverse as
	\begin{eqnarray}
	\label{second-eq-inversion}
		\big((B+\mu e_i e_j^T)+\mu e_j e_i^T\big)^{-1} &=& (B+\mu e_i e_j^T)^{-1} - \frac{\mu}{1 + e_i^T (B+\mu e_i e_j^T)^{-1} \mu e_j} \nonumber \\ 
		&& \times \big((B+\mu e_i e_j^T)^{-1} e_j\big) \big(e_i^T(B+\mu e_i e_j^T)^{-1}\big).
	\end{eqnarray}
	We use the expression in Eq.\eqref{first-eq-inversion} to calculate the terms of Eq.\eqref{second-eq-inversion}. We first calculate the $(ij)$-th entry
	\begin{align*}
		e_i^T (B+\mu e_i e_j^T)^{-1} e_j &= e_i^T C e_j - \frac{\mu}{1+\mu c_{i,j}}(e_i^TC_i)(C_j^Te_j) = c_{ij} - \frac{\mu c_{ii}c_{jj}}{1+\mu c_{ij}} \\ &= \frac{(1+\mu c_{ij})c_{ij} - \mu c_{ii}c_{jj}}{1+\mu c_{ij}},
	\end{align*}
	with which the fraction in Eq.\eqref{second-eq-inversion} calculates as
	\begin{align*}
		&\frac{\mu}{1 + e_i^T (B+\mu e_i e_j^T)^{-1} \mu e_j} = \frac{\mu}{1+ \mu (1+\mu c_{ij})^{-1} \{(1+\mu c_{ij})c_{ij} - \mu c_{ii}c_{jj}\}}\\&= \frac{\mu (1 + \mu c_{ij})}{(1 + \mu c_{ij}) + \mu (1 + \mu c_{ij}) c_{ij} - \mu^2 c_{i,i}c_{jj}}=\frac{\mu (1 + \mu c_{ij})}{(1 + \mu c_{ij})^2 - \mu^2 c_{ii}c_{jj}}.
	\end{align*}
	Calculating the column vectors
	\begin{align*}
		(B+\mu e_i e_j^T)^{-1} e_j = C e_j - \frac{\mu}{1+ \mu c_{ij}} C_i C_j^T e_j = C_j - \frac{\mu c_{jj}}{1 + \mu c_{ij}} C_i
	\end{align*}
	and 
	\begin{align*}
		e_i^T (B+\mu e_i e_j^T)^{-1} = C_i^T - \frac{\mu c_{ii}}{1+\mu c_{ij}}C_j^T.
	\end{align*}
	Putting these expressions and Eq.\eqref{first-eq-inversion} into Equation \eqref{second-eq-inversion} yields the identity
	\begin{align*}
	(B + \mu (e_ie_j^T + e_j e_i^T))^{-1} = C - \frac{\mu}{1 + \mu c_{ij}} C_iC_j^{T}-\frac{\mu (1 + \mu c_{ij})}{(1+\mu c_{ij})^2-\mu^2 c_{ii} c_{jj}}\\
	\times \left( C_j C_i^{T} - \frac{\mu c_{jj}}{1 + \mu c_{ij}} C_iC_i^{T} - \frac{\mu c_{ii}}{1 + \mu c_{ij}} C_jC_j^{T} +
	\frac{\mu^2 c_{ii} c_{jj}}{(1 + \mu c_{ij})^2}  C_iC^{T}_j\right).
	\end{align*}
	The final form in Eq.\eqref{lemma-inverse-equation} follows from the algebraic identity
	\begin{equation*}
		\frac{\mu}{1 + \mu c_{ij}} + \frac{\mu (1 + \mu c_{ij})}{(1+\mu c_{ij})^2-\mu^2 c_{ii} c_{jj}} \times \frac{\mu^2 c_{ii} c_{jj}}{(1 + \mu c_{ij})^2} = \frac{\mu (1 + \mu c_{ij})}{(1+\mu c_{ij})^2-\mu^2 c_{ii} c_{jj}}.
	\end{equation*}
\end{proof}

	The required criteria
	$1+\mu c_{ij}\neq 0$ and $(1+\mu c_{ij})^2 - \mu^2 c_{ii} c_{jj}\neq 0$
	are sufficient but not necessary. See for example
	\begin{equation*}
		B = \begin{bmatrix}
		1 & 1\\
		1 & 2
		\end{bmatrix},\
		C = B^{-1} = \begin{bmatrix}
		2 & -1\\
		-1 & 1
		\end{bmatrix},\
		\mu = 1.
	\end{equation*}
	Then $1+ \mu c_{12} =  1 - 1 = 0$ and while neither $B+ \mu e_1 e_2^T$ nor $B + \mu e_2 e_1^T$ are invertible, we have
	\begin{equation*}
		B + \mu (e_1e_2^T + e_2 e_1^T) = \begin{bmatrix}
		1 & 2\\
		2 & 2
		\end{bmatrix},\
		\big(B + \mu (e_1e_2^T + e_2 e_1^T)\big)^{-1} = \begin{bmatrix}
		-1 & 1\\
		1 & -\frac12
		\end{bmatrix}.
	\end{equation*}
	However, if criterion $1+\mu c_{ij}\neq 0$ is fulfilled, then the criterion $(1+\mu c_{ij})^2 - \mu^2 c_{ii} c_{jj}\neq 0$  is necessary and sufficient for $B+\mu(e_ie_j^T + e_j e_i^T)$ to be invertible. This follows directly from the Sherman-Morrison in \Cref{sherman-morrison}. Note further that if any of the two expressions $1+\mu c_{ij}$ and $(1+\mu c_{ij})^2 - \mu^2 c_{ii} c_{jj}$  are close to zero, then the formula may become numerically unstable.

\end{document}